\documentclass[acmsmall,screen,natbib=false]{acmart}
\usepackage{hhline}
\usepackage[backend=bibtex,sorting=none]{biblatex}
\usepackage{multirow}
\usepackage{adjustbox}

\AtBeginDocument{%
  \providecommand\BibTeX{{%
    \normalfont B\kern-0.5em{\scshape i\kern-0.25em b}\kern-0.8em\TeX}}}

\setcopyright{acmcopyright}
\copyrightyear{2023}
\acmYear{2023}
\acmDOI{XXXXXXX.XXXXXXX}

\begin{document}
\title{A Systematic Literature Review on Hardware Reliability Assessment Methods for Deep Neural Networks}

\author{Mohammad Hasan Ahmadilivani}
\email{mohammad.ahmadilivani@taltech.ee}
\affiliation{%
  \institution{Tallinn University of Technology}
  \city{Tallinn}
  \country{Estonia}}
  
\author{Mahdi Taheri}
\email{mahdi.taheri@taltech.ee}
\affiliation{%
  \institution{Tallinn University of Technology}
  \city{Tallinn}
  \country{Estonia}}

\author{Jaan Raik}
\email{jaan.raik@taltech.ee}
\affiliation{%
  \institution{Tallinn University of Technology}
  \city{Tallinn}
  \country{Estonia}}
  
\author{Masoud Daneshtalab}
\email{masoud.daneshtalab@mdu.se}
\affiliation{%
  \institution{Mälardalen University}
  \city{Västerås}
  \country{Sweden}}
  
\author{Maksim Jenihhin}
\email{maksim.jenihhin@taltech.ee}
\affiliation{%
  \institution{Tallinn University of Technology}
  \city{Tallinn}
  \country{Estonia}}

\renewcommand{\shortauthors}{M. H. Ahmadilivani et al.}

\begin{abstract}

Artificial Intelligence (AI) and, in particular, Machine Learning (ML) have emerged to be utilized in various applications due to their capability to learn how to solve complex problems. Over the last decade, rapid advances in ML have presented Deep Neural Networks (DNNs) consisting of a large number of neurons and layers. DNN Hardware Accelerators (DHAs) are leveraged to deploy DNNs in the target applications. Safety-critical applications, where hardware faults/errors would result in catastrophic consequences, also benefit from DHAs. Therefore, the reliability of DNNs is an essential subject of research.

In recent years, several studies have been published accordingly to assess the reliability of DNNs. In this regard, various reliability assessment methods have been proposed on a variety of platforms and applications. Hence, there is a need to summarize the state of the art to identify the gaps in the study of the reliability of DNNs. In this work, we conduct a Systematic Literature Review (SLR) on the reliability assessment methods of DNNs to collect relevant research works as much as possible, present a categorization of them, and address the open challenges.

Through this SLR, three kinds of methods for reliability assessment of DNNs are identified including Fault Injection (FI), Analytical, and Hybrid methods. Since the majority of works assess the DNN reliability by FI, we characterize different approaches and platforms of the FI method comprehensively. Moreover, Analytical and Hybrid methods are propounded. Thus, different reliability assessment methods for DNNs have been elaborated on their conducted DNN platforms and reliability evaluation metrics. Finally, we highlight the advantages and disadvantages of the identified methods and address the open challenges in the research area. We have concluded that Analytical and Hybrid methods are light-weight yet sufficiently accurate and have the potential to be extended in future research and to be utilized in establishing novel DNN reliability assessment frameworks. 
\end{abstract}



\maketitle

\section{Introduction} \label{intro}

Deep Neural Networks (DNNs) are nowadays extensively applied to a wide variety of applications due to their impressive ability to approximate complex functions (e.g. classification and regression tasks) via learning. Since powerful processing systems have evolved in the recent decade, DNNs have emerged to be deeper and more efficient as well as employed in an ever broader extent of domains. Moreover, using DNN Hardware Accelerators (DHAs) in safety-critical applications, including autonomous driving, raises reliability concerns \cite{bosio2021emerging}\cite{forsberg2020challenges}\cite{perez2022gpu}. In compliance to ISO 26262 functional safety standard for road vehicles, the evaluated FIT (Failures In Time) rates of hardware components must be less than 10 (meaning 10 failures in 1 billion hours) to pass the highest reliability level \cite{nardi2017functional} which requires diligent design.

\begin{figure}[h]
    \includegraphics[width=0.5\textwidth]{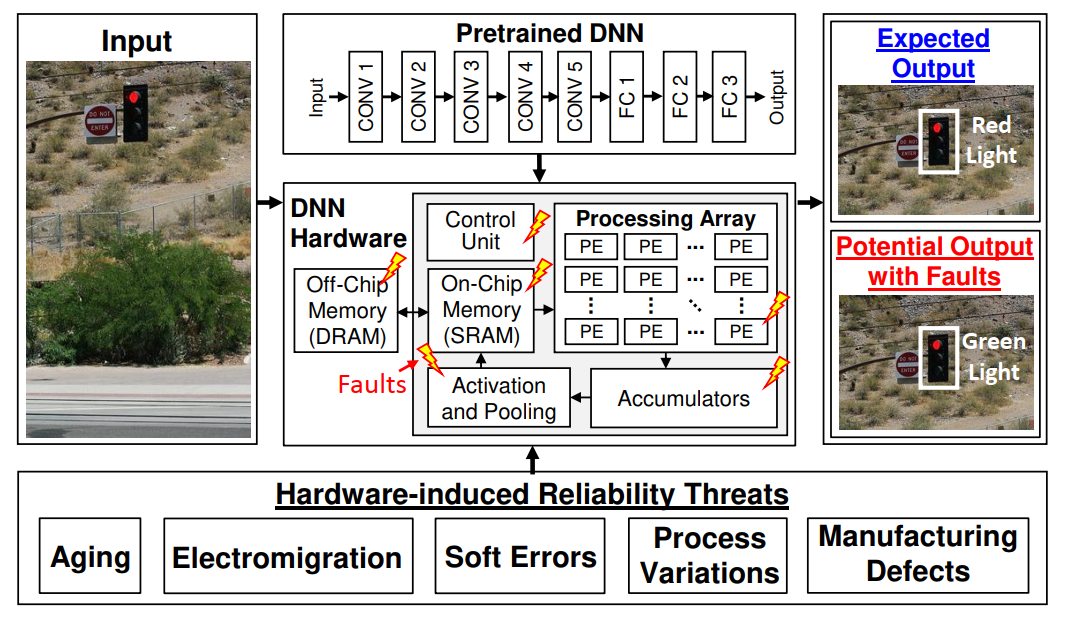}
    \centering
    \caption {Hardware-induced reliability threats in an example DHA and their possible impact on the output \cite{bosio2021emerging}.}
    \label{fig:rel_thr}
\end{figure}

DNNs are deployed in target applications using different DHA platforms, including Field-Programmable Gate Arrays (FPGAs), Application-Specific Integrated Circuits (ASICs), Graphics Processing Units (GPUs), and multi-core processors \cite{ibrahim2020soft}. Depending on the DHA platform and application environment, different fault types can pose a threat to the reliability of the component \cite{shafique2020robust}. Fig.~\ref{fig:rel_thr} illustrates the reliability threats (described in Section \ref{pre-definitions}) in an example DHA. In this figure, different fault types originating from various sources may occur in any of the DHA's components, leading to a disastrous misclassification such as detecting a red light as a green light. Although faults are hardware-induced, they can also be modeled in software platforms for ease of study. Therefore, the reliability of DNNs is tightly coupled with the reliability of DHAs. It is worth highlighting that the reliability in this paper does not encompass software engineering or security issues, such as adversarial attacks.

It has been shown in several studies that the functionality of DNNs in terms of accuracy is remarkably degraded in the presence of faults \cite{burel2021mozart}\cite{chitty2020model}\cite{khoshavi2020shieldenn}\cite{ozen2020low}\cite{dos2018analyzing}. Recently, numerous research works have been published on the assessment and enhancement of DNNs' reliability. However, due to the extent of the DNNs domain, these works approach the problem of the reliability of DNNs from various perspectives. We are faced with several applications of DNNs as well as a variety of DNN algorithms for different tasks. Therefore, it will lead to distinct platforms and reliability threats which hinders unifying and generalizing the methods of reliability assessment and enhancement of DNNs.

Throughout the literature, various methods of DNN reliability assessment and enhancement are presented. Some review papers have been published on the topic of DNNs reliability enhancement methods \cite{ibrahim2020soft}\cite{shafique2020robust}\cite{mittal2020survey}\cite{ruospo2023survey}\cite{su2023testability}\cite{torres2017fault}. These works aim to formulate the reliability problem in DNNs, categorize available reliability improvement methods in this domain, and overview the fault injection methods for reliability assessment. The analysis in \cite{torres2017fault} is the first review on the subject of fault tolerance in DNNs and describes different fault models and reliability improvement methods in DNNs. However, the topic was still not as mature as it is today, and numerous works have been published afterwards. Subsequent works such as \cite{shafique2020robust}\cite{ibrahim2020soft}\cite{mittal2020survey} provide extensive reviews on the reliability improvement methods for DNNs and characterize taxonomies of different methods. Nevertheless, they do not consider the assessment and evaluation methods of the reliability for DNNs. Other surveys  \cite{ruospo2023survey}\cite{su2023testability} have reviewed fault injection methods for DNNs reliability assessment, with the former work has focused merely on fault criticality assessment and the latter have included only a few papers in the survey. In this paper, we present the first Systematic Literature Review (SLR) dedicated to all methods of reliability assessment of DNNs.

Reliability assessment of DNNs is a process for evaluating the reliability of a DNN that is being executed either as a software model or by a hardware platform.
However, the assessment method for reliability may vary depending on the platform. In this regard, it is necessary to comprehend and distinguish the different methods used to assess the reliability of DNNs across platforms. This paper establishes a thorough picture of the reliability assessment methods for DNNs and systematically reviews the relevant literature. To achieve this, we carry out the SLR methodology \cite{cicchetti2019multi}\cite{lavallee2013performing} to present this survey. The primary focus of this review is to investigate the methods of reliability assessment for DNNs, generalize and characterize the methods, and identify the open challenges in the domain. 

To the best of our knowledge, this survey represents  the first comprehensive literature review on reliability assessment methods for DNNs. We cover all published papers from 2017 to 2022 that could be found through a systematic search. The main contributions of this paper are:

\begin{itemize}
    \item Reviewing the literature of the reliability assessment methods of DNNs, systematically;
    \item Analyzing the trends of published papers over different years and methods;
    \item Characterizing and categorizing the reliability assessment methods for DNNs;
    \item Identifying fault injection methods based on the DNN platforms;
    \item Introducing analytical and hybrid reliability assessment methods along with fault injection;
    \item Addressing the open challenges in the research area and recommendations for future research directions.
\end{itemize}

The structure of the paper is as follows. Section \ref{preliminaries} presents the background on DNNs and reliability concepts, Section \ref{method} explains the methodology of this survey and addresses the research questions, Section \ref{overview} reviews the study briefly, presents the statistics of the publications, and depicts the top-level taxonomy of reliability assessment methods for DNNs, in Section \ref{charaterization} the details of the reliability assessment methods are explained, Section \ref{discussion} includes pros and cons of methods and open challenges of the study domain, and Section \ref{conclusion} provides the conclusions of this survey.

\section{Preliminaries} \label{preliminaries}

\subsection{Deep Neural Networks} \label{pre-dnns}

Deep Learning (DL) is a sub-domain of Machine Learning (ML) which is the study of making computers learn to solve problems without being directly programmed \cite{sze2017efficient}. Regarding the impressive ability of DNNs in learning, they are applicable in a vast variety of domains like image and video processing, data mining, robotics, autonomous cars, gaming, etc. 

DNNs are inspired by the human brain, and they have two major phases: training and inference. In the training phase which is an iterative process and performed once, the hyper-parameters (e.g., weights, and biases) of the neural network are updated on a determined dataset. A loss function is adopted in the training phase that measures the difference between the expected and the estimated output of DNN to achieve higher accuracy. Accuracy expresses the proportion of the DNN outputs coinciding with the expected output. On the other hand, in the inference phase, representing the DNN deployment, the network is being run several times with the parameters obtained during the training phase \cite{sze2017efficient}.

DNNs are constructed of the units of neurons. Each neuron receives some activation inputs and multiplies them by the corresponding weights. Then, it conveys the summation of the weighted activations to its output. A set of neurons build up a layer that may have other additional functions, e.g., activation function (ReLu, sigmoid etc.), batch normalization, (max or average) pooling, etc. \cite{sze2017efficient}. Equation \eqref{eq:DNN} represents the function of the \textit{i}-th neuron in layer \textit{l} (denoted as \(N_i^l\)) with input activations from the previous layer \textit{l-1} with \textit{n} outputs (denoted as \(X^{l-1}\)), where \textit{W} and \textit{b} represent weights and bias, respectively.

\begin{equation}
N_i^l = \phi(\sum_{j=0}^{n} X_j^{l-1} \times  W_{ij}^l + b^l)
\label{eq:DNN}
\end{equation}

\begin{figure}[ht]
    \includegraphics[width=0.5\textwidth]{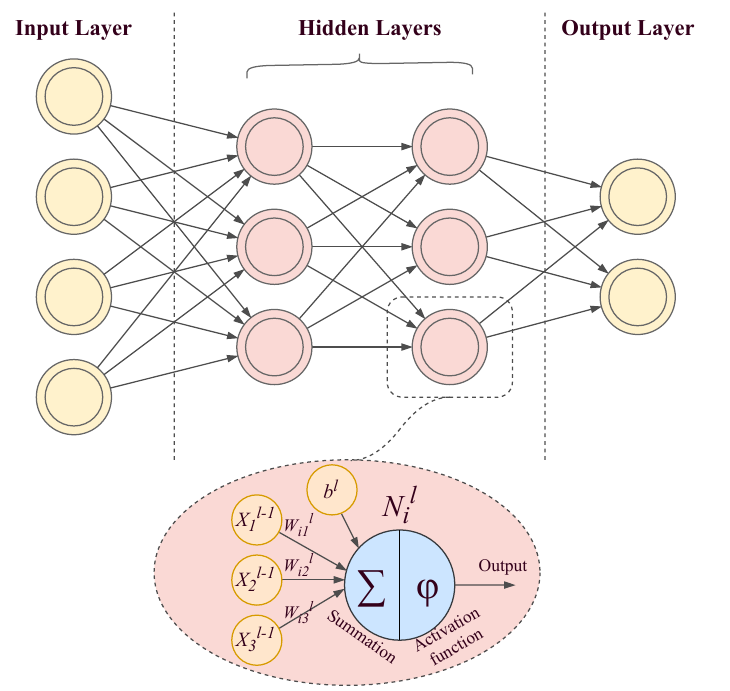}
    \centering
    \caption {Abstract view of a simple neural network with the detail of a neuron}
    \label{fig:DNN}
\end{figure}

An abstract view of a neuron and a neural network is depicted in Fig.~\ref{fig:DNN}. As shown, inputs are fed into the network through the input layer. The middle layers, called hidden layers, determine the depth of the network and conduct the function of the DNN. The output layer is where the network decides. It produces some probabilities of the possible outputs, i.e., output confidence score, and the class with the highest value is the top-ranked output.

DNNs have various architectures each suitable for specific applications. Nevertheless, it is worth mentioning some terms which are used in this paper. Convolutional Neural Networks (CNNs) are extensively used in classification, object detection and semantic segmentation tasks and consist of multiple convolutional (CONV) and fully-connected (FC) layers. CONV layers have a set of two-dimensional (2D) weights, called filters, that extract a specific feature from the input of the layer. A channel is a set of input feature maps (ifmap) that is convolved with filters resulting in the output feature maps (ofmap) \cite{sze2017efficient}.

In the research area of CNNs, there are some models of networks that are most frequently used. For instance, LeNet-5 \cite{lecun1998gradient}, AlexNet \cite{krizhevsky2012imagenet}, GoogLeNet \cite{szegedy2015going}, VGG \cite{simonyan2014very}, and ResNet \cite{he2016deep} are introduced for image classification, and YOLO \cite{redmon2016you} is designed for object detection. In addition, prominent datasets that are mostly used for training networks on image classification tasks are MNIST \cite{MNIST_DS}, CIFAR \cite{CIFAR_DS}, and ImageNet \cite{deng2009imagenet}, and on object detection are KITTI \cite{menze2015object}, and PASCAL VOC \cite{everingham2010pascal}.

In addition, regarding the large number of parameters and calculations of DNNs, Quantized Neural Networks (QNNs) \cite{hubara2017quantized} and Binarized Neural Networks (BNNs) \cite{courbariaux2016binarized} are introduced to reduce the complexity, memory usage, and energy consumption of DNNs. These DNNs are the quantized versions of existing DNNs that reduce the bit-width of DNNs parameters and calculations with an acceptable accuracy loss.

\subsection{DNN Platforms} \label{pre-platforms}

\subsubsection{Software Frameworks}

DNN software frameworks and libraries in high-level programming languages have been developed to ease the process of designing, training, and testing DNNs. These frameworks are widely used due to their high abstraction level of modeling and short design time. Some of well-known software frameworks that are being used for training the DNNs are: TensorFlow \cite{abadi2016tensorflow}, Keras \cite{keras_tool}, PyTorch \cite{paszke2019pytorch}, DarkNet \cite{darknet13}, and Tiny-DNN \cite{tinydnn}. All these frameworks are capable of using both CPU and GPU to accelerate the training process. 

\subsubsection{DNN Hardware Accelerators (DHAs)}

DHAs are used for the training as well as the inference phase of DNNs. They are called accelerators due to their dedicated design employing parallelism for reducing the execution time of the DNN, either in training or inference. DHAs can be generally categorized into four classes: FPGAs, ASICs, GPUs, and multi-core processors \cite{abich2020soft}\cite{talib2021systematic}. 

\begin{figure}[h]
\includegraphics[width=0.45\textwidth]{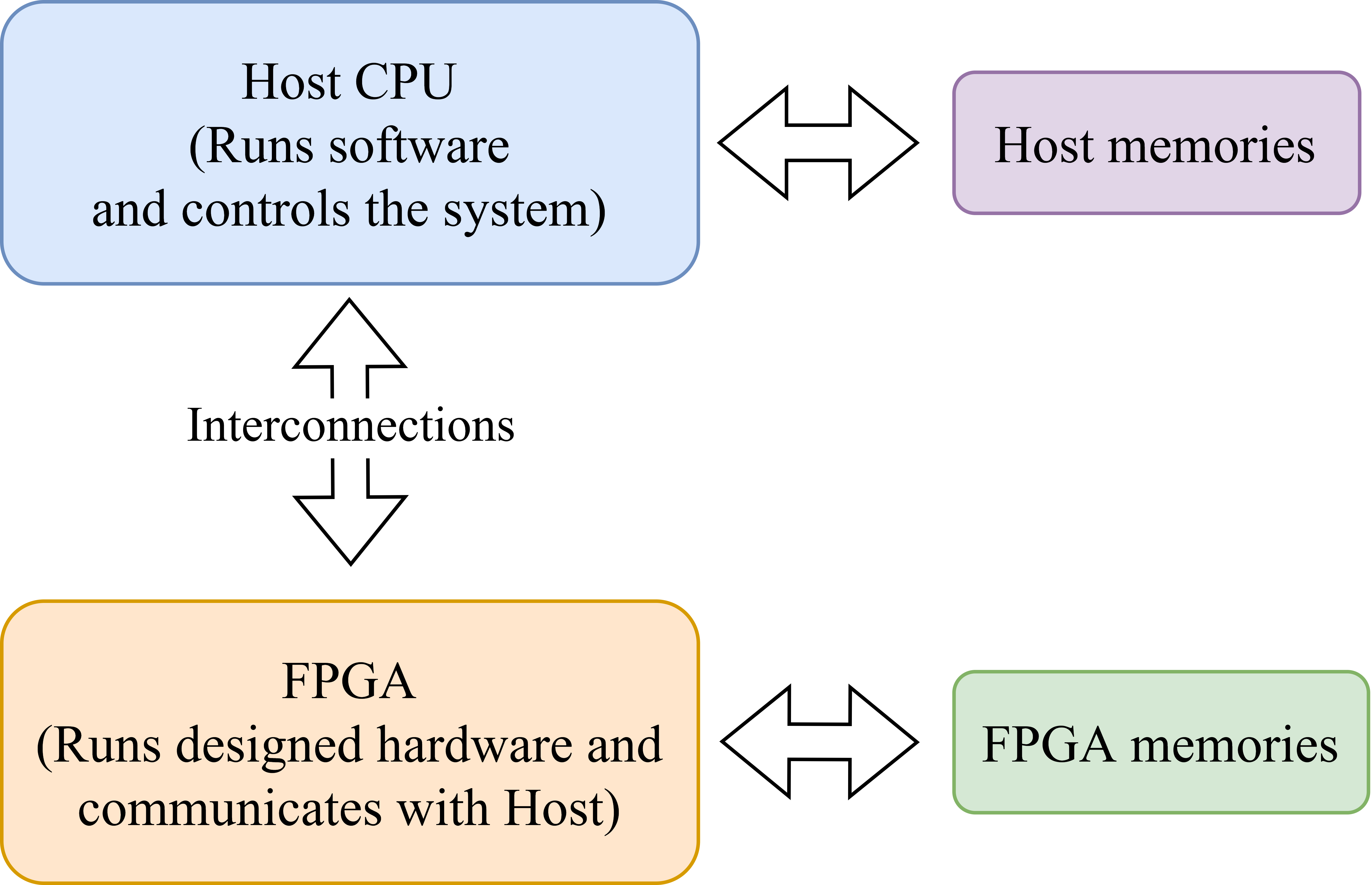}
\centering
\caption {Typical structure of an FPGA-based DNN accelerator \cite{guo2019dl}}
\label{fig:fpga_dha}
\end{figure}

According to the literature review of DHAs in \cite{talib2021systematic}, FPGAs are used more frequently than other DHA platforms in terms of implementing DNNs, due to its availability and design flexibility for different applications \cite{guo2019dl}. FPGAs are programmed via their configuration bits that determine the functionality of the FPGA. The system of FPGA-based DNN accelerators usually consists of a host CPU and an FPGA part with corresponding interconnections between them. In this design model, the DNN is implemented on the FPGA part and the CPU controls the accelerator with software, while each part is integrated with memories \cite{guo2019dl}. A typical structure of FPGA-based DNN accelerator is depicted in Fig.~\ref{fig:fpga_dha} which is based on HW/SW co-design, that means separating the implementation of DNNs on the integrated CPU (the software) and FPGA (the hardware) that are communicating with one another \cite{hou2019survey}. High-Level Synthesis (HLS) tools which can synthesize high-level programming languages to RTL are also used for developing FPGA-based DNN accelerators \cite{guo2019dl}. 

\begin{figure}[h]
\includegraphics[width=0.5\textwidth]{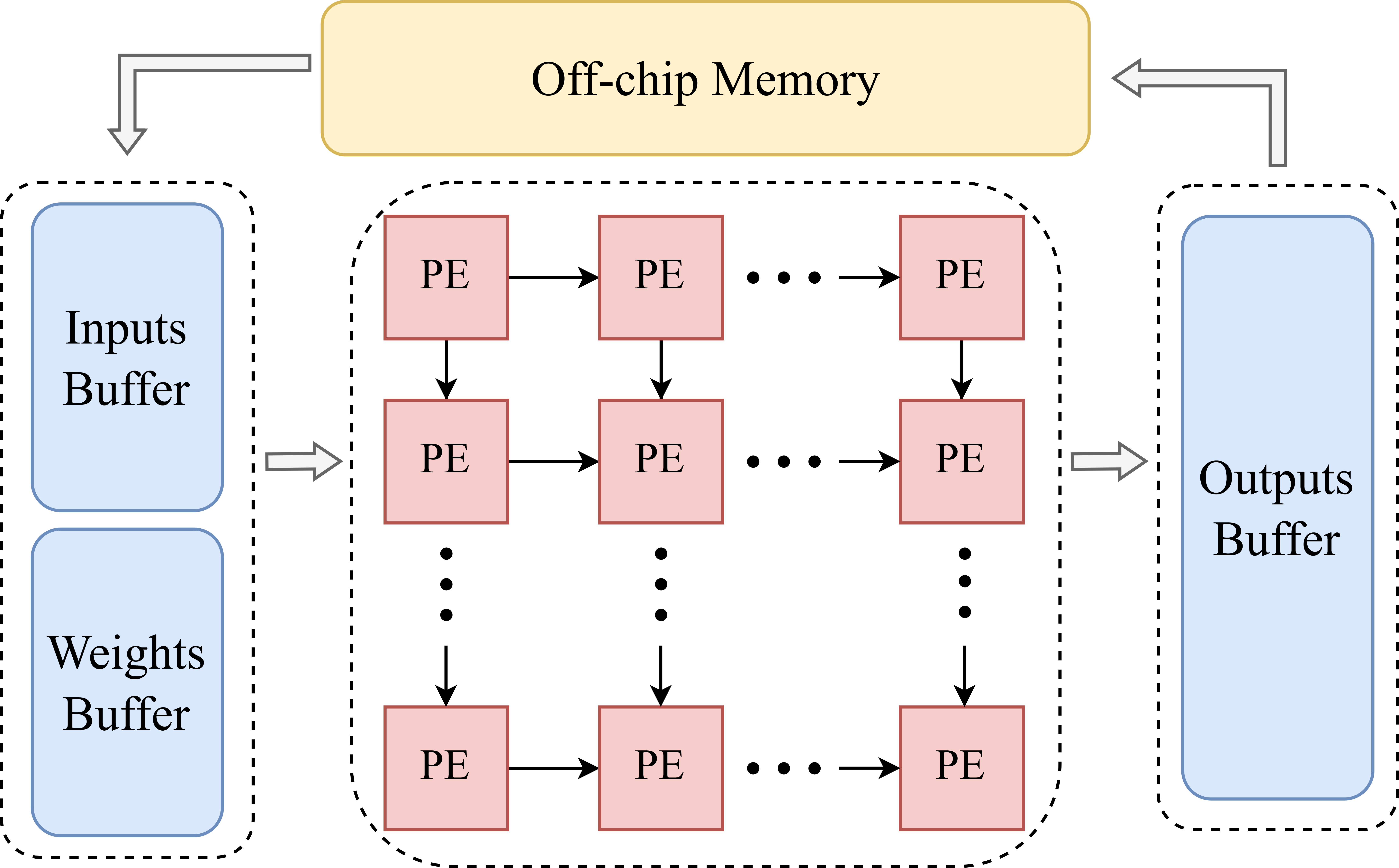}
\centering
\caption {An example of spatial architecture for ASIC-based DNN accelerators \cite{moolchandani2021accelerating}}
\label{fig:asic_dha}
\end{figure}

ASIC-based DNN accelerators are more efficient than FPGAs in terms of performance and power consumption but less flexible in terms of applications and require a long design time \cite{dhouibi2021accelerating}. There are two general types of architectures for ASIC-based DHA platforms: spatial and temporal \cite{sze2017efficient}. Fig.~\ref{fig:asic_dha} depicts an example of a spatial architecture model that is constructed of 2D arrays of Processing Elements (PEs) flowing data horizontally and vertically from input/weight buffers to output buffers. PEs perform Multiply-Accumulate (MAC) operations on inputs and weights representing a neuron operation in the DNN. Off-chip memories are required to store the parameters of DNNs and save the intermediate results from PEs. Tensor Processing Unit (TPU) produced by Google, one of the most applicable ASIC-based DNN accelerators, is based on this type of architecture \cite{jouppi2017datacenter}. 

\begin{figure}[h]
\includegraphics[width=0.5\textwidth]{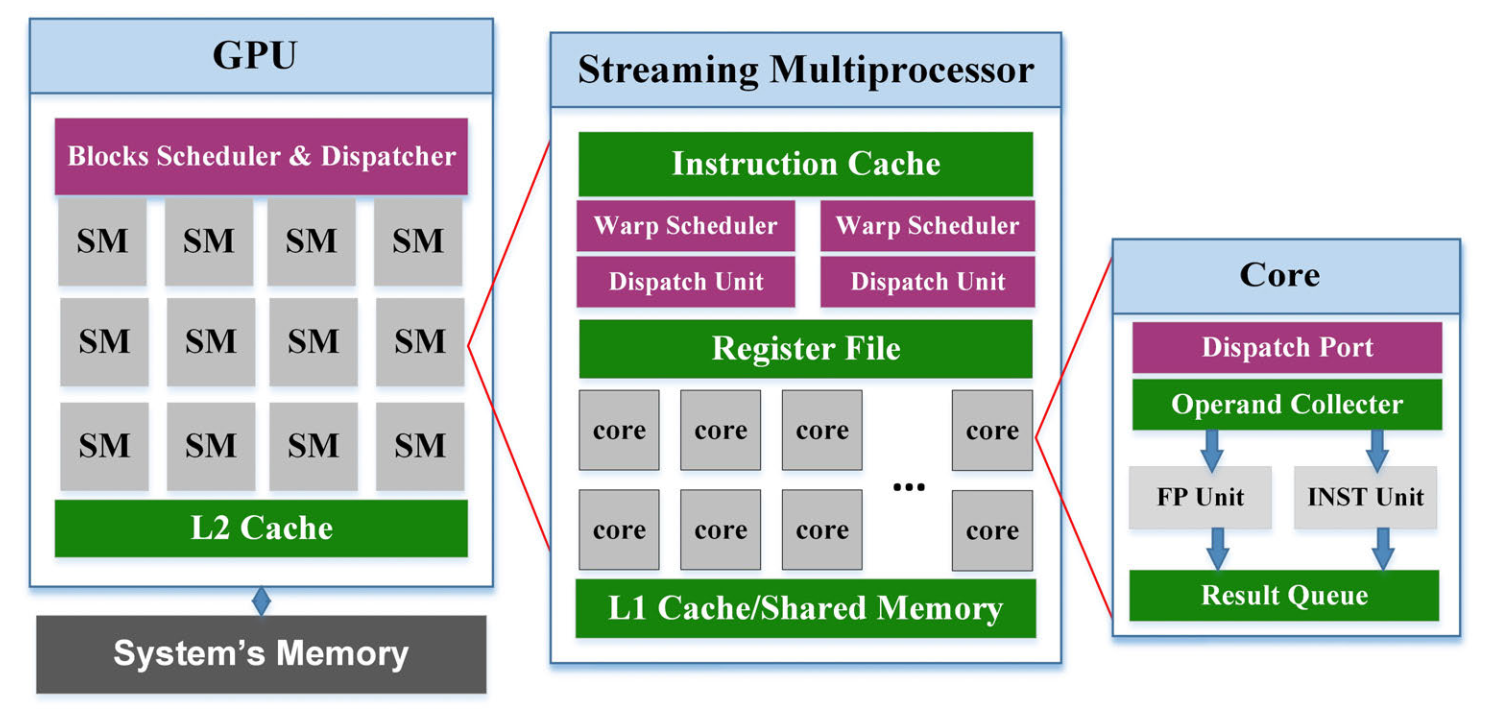}
\centering
\caption {General architecture of CUDA-based GPUs \cite{ibrahim2020softerror}}
\label{fig:gpu_arch}
\end{figure}

GPUs are a powerful platform for training and inferring deep networks and are vastly used in safety-critical applications \cite{perez2022gpu}. GPUs include up to thousands of parallel cores, which make them efficient for DNN algorithms, especially in the training phase \cite{dhouibi2021accelerating}. GPUs are designed to run several threads of a program and are also exploited to accelerate running DNNs \cite{talib2021systematic}. The general architecture of GPUs is depicted in Fig.~\ref{fig:gpu_arch}. There are numerous Streaming Multiprocessors (SMs) in the GPU, each having several cores with a shared register file and caches, while a scheduler and dispatchers control the tasks among and within SMs and cores \cite{ibrahim2020softerror}. 

Multi-core processors, e.g., ARM processors, deploy DNNs mostly for edge processing and Internet of Things (IoT) applications \cite{lai2018cmsis}\cite{mahdavinejad2018machine}\cite{sanchez2020tinyml}. They facilitate DNNs with parallel computing and low power consumption and provide wider range of applications for DNNs. 

\subsection{Reliability, Threats, Fault Models, and Evaluation} \label{pre-definitions}
Terms of robustness, reliability, and resilience are mostly used in the research pertaining to the reliability of DNNs. These terms are often used interchangeably and ambiguously. In the following, we present the definitions of these three terms as applied in the current literature review:

\begin{itemize}
\item \textbf{Reliability} concerns DNN accelerators' ability to perform correctly in the presence of faults, which may occur during the deployment caused by physical effects either from the environment (e.g. soft errors, electromagnetic effects) or from within the device (e.g. manufacturing defects, aging effects, process variations).

\item \textbf{Robustness} refers to the property of DNNs expressing that the network is able to continue functioning with high integrity despite the alteration of inputs or parameters due to noise or malicious intent.

\item \textbf{Resilience} is the feature of DNN to tolerate faults in terms of output accuracy.
\end{itemize}

In this work, we are concerned about the reliability of DNNs, which refers to the ability of accelerators to continue functioning correctly in a specified period of time with the presence of faults. Reliability in this paper does not relate to the reliability and test in software engineering or security issues e.g., adversarial attacks in which an attacker perturbs the inputs or parameters.

\textit{Faults} are the sources of threatening the reliability of DNN accelerators (See Fig.~\ref{fig:rel_thr}) that can be caused by several reasons, e.g., soft errors, aging, process variation, etc. \cite{bosio2021emerging}. Soft errors are transient faults induced by radiation that are caused by striking charged particles to transistors \cite{baumann2005radiation}. Aging is the time-dependent effect of the increasing threshold voltage of transistors due to physical phenomena that will lead to timing errors and permanent faults \cite{chen2003dynamic}. Process variations are alteration of transistor's attributes in the process of chip fabrication that may cause occurring faults by voltage scaling \cite{borkar2005designing}.

Faults as reliability threats are generally modeled as \textit{permanent} and \textit{transient} faults \cite{mittal2020survey}\cite{shafique2020robust}\cite{torres2017fault}. Permanent faults result from process variations, manufacturing defects, aging, etc., and they stay constant and stable during the run-time. On the other hand, transient faults are caused by soft errors, electromagnetic effects, voltage and temperature variations, etc., and they show up for a short period of time. Nevertheless, once a faulty value from a component is read by another component and the propagated value does not coincide with the expected one, an \textit{error} happens. Therefore, a fault is an erroneous state of hardware or software, and an error is a manifestation of it at the output. \textit{Failure} or system malfunction is the corruption or abnormal operation of the system which is caused by errors \cite{torres2017fault}\cite{koren2007fault}\cite{johnson1984fault}.

\begin{figure}[h]
\includegraphics[width=0.55\textwidth]{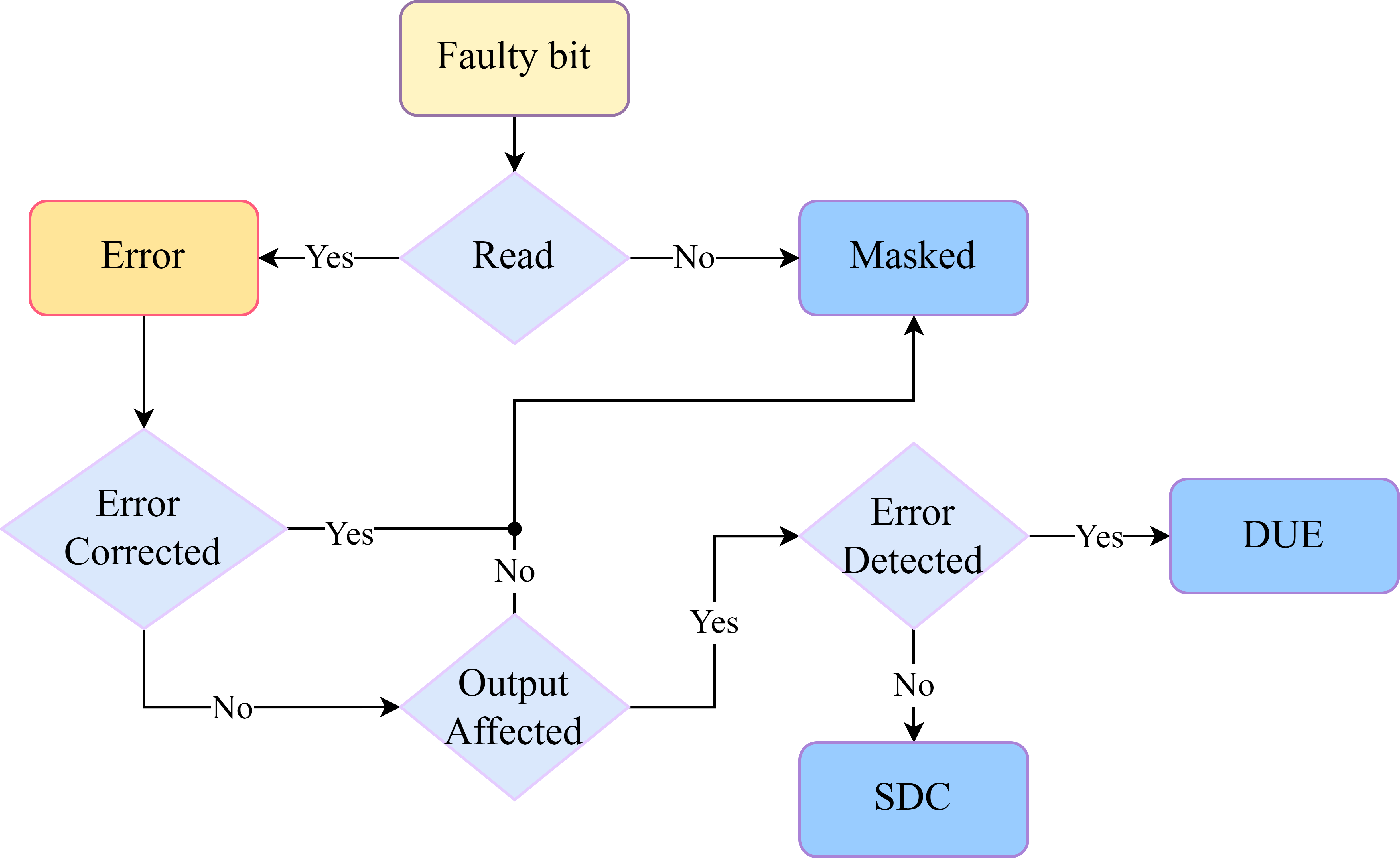}
\centering
\caption {The adopted fault classification based on the output point of view, as in \cite{koren2007fault}}
\label{fig:faults}
\end{figure}

Faults may have different impacts on the output of DNNs and can be classified based on their effects. A fault may be masked or corrected if detected or result in different outputs compared to the fault-free execution (golden model), in which case the fault is propagated and observed at the output. Faults observed at the output of the system can be classified in two categories: Silent Data Corruption (SDC) and Detected Unrecoverable Errors (DUE), depending on whether a fault is undetected (SDC) or detected (DUE) \cite{mittal2020survey}\cite{biswas2005computing}. Fig.~\ref{fig:faults} illustrates this general fault classification scheme regarding the output of systems adopted from \cite{koren2007fault}.

\textit{Reliability assessment} is the process in which the target system or platform is modeled or presented, and by means of simulations, experiments, or analysis, the reliability is measured and evaluated. Reliability assessment is a challenging process and several methods can be adopted for modeling and evaluating reliability. In general, evaluating the reliability of a system can be performed by three approaches: Fault Injection (FI) methods, analytical methods, and hybrid methods \cite{eslami2020survey}. FI methods are exploited to inject a model of faults into the system implemented either in software or hardware, while the system is in simulation or being executed. Analytical methods attempt to model the function of the system and its reliability with mathematical equations depending on the target architecture. In hybrid methods, an analytical model is adopted alongside an FI to evaluate the reliability. Generally, FI methods are more realistic than analytical and hybrid methods; however, FI is a time-consuming process with a high computational complexity \cite{ruospo2021pros}. 

In the reliability assessment using FI, it is necessary to determine the target platform, potential fault locations (logic or memory), and  the fault type (transient or permanent). Transient faults in logic show up in one clock cycle, while in the memory, they flip a bit that will remain until the end of the execution. Permanent faults are modeled as stuck-at-0 (sa-0), or stuck-at-1 (sa-1), and they exist during the whole execution. According to the selected fault model, perturbation of the model is performed, the system is run, and the outputs are gathered. The output of faulty execution should be compared with the one of the golden-model to measure the impact of faults on the system. 

FI allows calculating reliability metrics, e.g., Failures-In-Time (FIT),  Architectural Vulnerability Factor (AVF), SDC rate, Soft Error Rate (SER), cross-section, etc. FIT is the number of failures in $10^9$ hours, AVF is the probability of fault propagation from a component to other components in a design, SDC rate refers to the ratio of the outputs affected by faults, SER refers to the ratio of soft error occurrence and cross-section is the proportion of observed errors over all collided particles. These quantitative evaluation metrics are usually tightly coupled to each other, yet follow a different purpose to express the reliability of a system. 

Exhaustive fault injection into all bits of a platform at every clock cycle requires an extensive simulation. Therefore, to determine how many faults could be injected into the system in order to be representative statistically, a confidence level with an error margin is presented \cite{leveugle2009statistical}. It provides a fault rate or Bit Error Rate (BER) for an FI experiment. The number of FI experiments' repetitions regarding the number of possible bit and clock cycle combinations to support the number of injected faults determines the execution space for the FI task.

\section{Review Methodology} \label{method}
Systematic Literature Review (SLR) is a standard methodology for reviewing the literature in a recursive process and minimizing bias in the study \cite{cicchetti2019multi}\cite{lavallee2013performing}\cite{talib2021systematic}. Hence, the SLR methodology is adopted in this survey. The methodology determines:

\begin{itemize}
\item Specifying the Research Questions (RQs), 
\item Specifying the search method for finding and filtering the related papers, 
\item Extracting corresponding data from the found papers based on the RQs, 
\item Synthesizing and analyzing the extracted data.
\end{itemize}

Therefore, based on the aforementioned steps of SLR, the RQs which we attempt to answer are:

\begin{itemize}
\item \textbf{RQ1:} What is the distribution of the research works in the domain of reliability assessment? (To obtain the trend of publications in this domain).
\item \textbf{RQ2:} What are the existing methods of reliability assessment for DNNs? (To comprehend the entire variety of methods in this domain).
\item \textbf{RQ3:} How could the existing methods be characterized and categorized in terms of reliability assessment methods? (To categorize existing works and provide the taxonomy, a systematic instruction for finding the suitable method for potential applications in this domain).
\item \textbf{RQ4:} What are the open challenges in the domain of reliability assessment methods for DNNs? (To specify the remaining areas for future research).
\end{itemize}

The motivation for this survey is the numerous recent  papers published on the reliability of DNNs emphasizing the need for such a literature review. We have searched for the papers systematically through scientific search servers. The main databases and publishers we have used are: Google Scholar, IEEE Explore, ACM Digital Library, Science Direct, and Elsevier. The initial set of papers are provided by searching some keywords in the mentioned servers, including "reliability of DNNs", "hardware reliability of DNN accelerators", "resilient DNNs", "robust DNNs", "the vulnerability of DNNs", "soft errors in DNNs", "fault injection in DNNs" ("DNN" also replaced with "CNN"). 

Subsequently, based on the title and abstract of each paper, we select them. This selection is based on the criterion of whether the paper may concern the reliability of DNNs or not. In addition, the references and citations of the papers have been checked for the chosen papers to find more related papers. In this process, we selected 242 papers based on their title and abstract.

In the next step, we study the introduction, conclusion, and methodology sections of each paper to decide whether we include the paper in the review or not. The inclusion criteria of the papers are:

\begin{itemize}
    \item The paper is published by one of the scientific publishers and has passed through a peer-review process,
    \item The focus of the work is DNN, neither generic reliability assessment methods using DNNs as one of the examples nor employing DNNs for assessing the reliability of a platform.
    \item The work includes a reliability assessment method for DNNs,
    \item The method of reliability assessment is clear and well-defined,
    \item Terms including reliability, robustness, resilience, or vulnerability \textbf{must} refer clearly to reliability issues, as defined in subsection \ref{pre-definitions}.
\end{itemize}

Papers that have included similar keywords but have not matched the above conditions are excluded. As a result, we have included 139 papers published from 2017 to the end of 2022 in this literature review to build up the taxonomy of the literature review and methods' categorization.

In the following, we have designed a Data Extraction Form (DEF) based on the RQs. In this form, we have taken note of reviewing the papers to find some specific data such as:
\begin{itemize}
    \item General method of reliability modeling (FI, analytical, or hybrid),
    \item The platform where DNNs are implemented,
    \item The fault model and fault locations in case of FI,
    \item Details of reliability assessment method,
    \item Reliability evaluation metrics.
\end{itemize}

In the final step, after reviewing all the selected papers and filling in the DEF, we synthesized and analyzed the obtained data from the papers. Thereafter, we have provided the categorization taxonomy of the reliability assessment methods for DNNs, have characterized them in this paper, and analyzed them to find the open challenges.

\section{Study Overview} \label{overview}

This section presents an overview of the study and the analyzed statistics of the included works in different categories. As mentioned, we have included 139 papers from 2017 to 2022 for categorizing the reliability assessment methods for DNNs.

\subsection{Taxonomy}
\begin{figure*}[h]
\includegraphics[width=\textwidth]{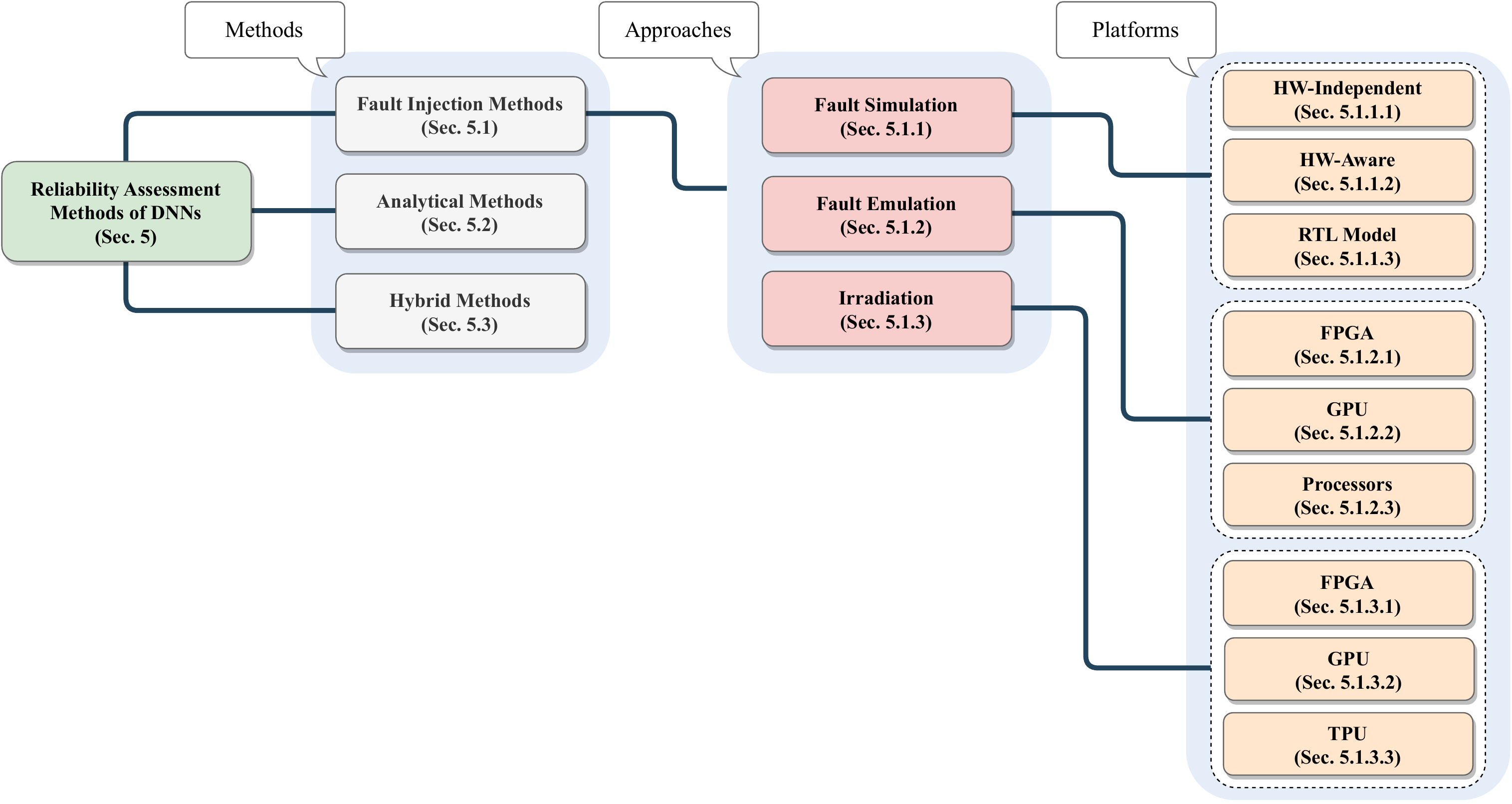}
\centering
\caption {Top-level overview of the reliability assessment methods in this work.}
\label{fig:top-level}
\end{figure*}

Fig.~\ref{fig:top-level} represents the top-level categorization overview of the study to address RQ2 and RQ3. Reliability assessment of DNNs, are categorized into three main methods: Fault Injection, Analytical, and Hybrid.

\subsubsection{\textbf{Fault Injection (FI) Methods}}

The works based on this method evaluate the reliability of DNNs by fault injection campaign. There exist several taxonomies for the fault injection approaches in the hardware reliability domain \cite{ruospo2023survey}\cite{eslami2020survey}\cite{ruospo2021pros}\cite{benso2011art}\cite{ruospo2020pipelined}. Therefore, we adapt them for categorizing the related works on DNNs into three approaches addressed in Fig.~\ref{fig:top-level} and Table \ref{tab:FI-cat}. FI methods are categorized into three approaches of fault injection as follows:

\begin{itemize}
    \item \textbf{Fault Simulation:} DNNs are implemented either in software by high-level programming languages or Hardware Description Languages (HDL) and faults are injected into the model of the DNN. In the former case, some works consider a DHA model in their software implementations while others do not. We divide works on this approach into hardware-independent, hardware-aware, and RTL model platforms. RTL models represent ASIC-based DHAs.
    \item \textbf{Emulation in Hardware:} Research works on this approach implement and run DNNs on a DHA (i.e., FPGA, GPU, or processor) and inject the faults into the components of the accelerator by a software function, FI framework, etc.
    \item \textbf{Irradiation:} DNN is implemented on a DHA (i.e., FPGA, GPU, or TPU) placed under an irradiating facility to inject beams onto it.
\end{itemize}

Most of the works on DNNs' reliability assessment use FI methods. Therefore, we characterize three approaches of FI methods in Table \ref{tab:FI-cat}. In each approach of FI methods, the works are distinguished based on DNN platforms. Furthermore, in each category, we elaborate on how the works determine the fault types and locations and evaluate the reliability by metrics. The details will be discussed in subsection \ref{char-FI}.


\begin{table*}[h!]
\caption{Fault injection categorization with the corresponding references.}
\begin{adjustbox}{angle=90,width=0.7\textwidth}
\begin{tabular}{|cccccccccc|}
\hline
  \multicolumn{1}{|c|}{\textbf{FI Method}} &
  \multicolumn{3}{c||}{\textbf{Fault Simulation}} &
  \multicolumn{3}{c||}{\textbf{Fault Emulation}} &
  \multicolumn{3}{c|}{\textbf{Irradiation}} \\ 
  \hline
  \multicolumn{1}{|c|}{\textbf{DNN Platform}} &
  \multicolumn{1}{c|}{\textbf{HW-Independent}} &
  \multicolumn{1}{c|}{\textbf{HW-Aware}} &
  \multicolumn{1}{c||}{\textbf{RTL Model}} &
  \multicolumn{1}{c|}{\textbf{FPGA}} &
  \multicolumn{1}{c|}{\textbf{GPU}} &
  \multicolumn{1}{c||}{\textbf{Processors}} &
  \multicolumn{1}{c|}{\textbf{FPGA}} &
  \multicolumn{1}{c|}{\textbf{GPU}} &
  \textbf{TPU} 
  \\ \hline
\multicolumn{1}{|c|}{\multirow{2}{*}{\begin{tabular}[c]{@{}c@{}} \\ \textbf{Fault Type}\end{tabular}}} &
  \multicolumn{1}{c|}{\begin{tabular}[c]{@{}c@{}}Transient \cite{ali2020erdnn,amarnath2022soft,arechiga2018effect} \\ \cite{arechiga2018robustness,burel2021zero,burelt2022improving,cantoro2020evaluating,chen2021low,deligiannis2021towards} \\ \cite{gao2020reliability,ghavami2022fitact,goldstein2020reliability,guan2019place,hoang2020ft,jang2021mate} \\ 
  \cite{lee2022bipolar,malekzadeh2021impact,neggaz2018reliability,neggaz2019cnns,ozen2021snr} \\
  \cite{ponader2021milr,sabbagh2019evaluating,syed2021fault,zhan2021improving}
  \end{tabular}} &
  
  \multicolumn{1}{c|}{\begin{tabular}[c]{@{}c@{}}Transient \cite{ozen2020low,azizimazreah2018tolerating} \\ \cite{goldstein2021lightweight,jasemi2020enhancing,kim2019dris,li2017understanding} \\ 
  \cite{li2020soft,ozen2020boosting,ozen2020just,ozen2019sanity}  \end{tabular}} &
  
  \multicolumn{1}{c||}{\begin{tabular}[c]{@{}c@{}}Transient \\ \cite{abich2020soft,bandeira2019non,corneliou2021fine,hosseinkhani2021improving}  \end{tabular}} &
  
  \multicolumn{1}{c|}{\begin{tabular}[c]{@{}c@{}}Transient \cite{khoshavi2020shieldenn} \\ 
  \cite{benevenuti2018comparative,benevenuti2021neutron,de2020emulation,de2022firenn,du2019reliability} \\ 
  \cite{gao2022reliability,khoshavi2020compression,khoshavi2020fiji} \\ 
  \cite{libano2018selective,libano2020understanding,luza2021model} \\
  \cite{matanaluza2021emulating,souvatzoglou2021analyzing,wang2021impact} \end{tabular}} &
  
  \multicolumn{1}{c|}{\begin{tabular}[c]{@{}c@{}}Transient \cite{dos2018analyzing} \\ 
  \cite{ibrahim2020softerror,adam2021selective,adam2021analyzing,adam2021impact} \\
  \cite{bolchini2022selective,cavagnero2022transient,condia2021combining,dos2022characterizing} \\ 
  \cite{garrett2021improving,hari2021making,ibrahim2020analyzing,,ibrahim2020analyzinginstruction} \\ 
  \cite{rech2020impact,dos2017evaluation,dos2019impact} \end{tabular}} &
  
  \multicolumn{1}{c||}{\multirow{2}{*}{\begin{tabular}[c]{@{}c@{}}Transient \cite{abich2020soft} \\
  \cite{bandeira2019non,abich2021impactprecision,abich2021applying} \\
  \cite{abich2022impactthread,abich2022impact,gava2022soft}  \\
  \cite{liu2022using,liu2022efficient,liu2021analyzing}  \end{tabular}}}  &
  
  \multicolumn{1}{c|}{\multirow{2}{*}{\begin{tabular}[c]{@{}c@{}}Transient \\
  \cite{benevenuti2018comparative,benevenuti2021neutron,libano2018selective,matanaluza2021emulating} \\  
  \cite{wang2021impact,agiakatsikas2021evaluation,gambardella2022accelerated} \\
  \cite{libano2021reduced,luza2020investigating,maillard2022radiation}  \end{tabular}}} &
  
  \multicolumn{1}{c|}{\multirow{2}{*}{\begin{tabular}[c]{@{}c@{}}Transient \\ \cite{dos2018analyzing,dos2022characterizing,hari2021making} \\ 
  \cite{dos2017evaluation,dos2019impact} \\
  \cite{basso2020impact,lotfi2019resiliency}   \end{tabular}}} &
  
  \multirow{2}{*}{\begin{tabular}[c]{@{}c@{}}Transient \\ 
  \cite{junior2022high,rech2022reliability} \end{tabular}}
  \\
  \cline{2-6}
  
\multicolumn{1}{|c|}{} &
  \multicolumn{1}{c|}{\begin{tabular}[c]{@{}c@{}}Permanent \cite{hoang2020ft} \\ \cite{bosio2019reliability,lee2022value,ruospo2020evaluating,ruospo2021investigating}  \end{tabular}} &
  
  \multicolumn{1}{c|}{\begin{tabular}[c]{@{}c@{}}Permanent \cite{burel2022mozart+} \\ 
  \cite{burel2021mozart,hoang2021tre,nguyen2021low,siddique2021exploring} \\
  \cite{tsai2021evaluating,zahid2020fat,zhao2022fsa}  \end{tabular}} &
  
  \multicolumn{1}{c||}{\begin{tabular}[c]{@{}c@{}}Permanent \\ \cite{chitty2020model,ruospo2020pipelined,liu2021hyca} \\ \cite{xu2020hybrid,zhang2019fault,zhang2018analyzing} \end{tabular}} &
  
  \multicolumn{1}{c|}{\begin{tabular}[c]{@{}c@{}}Permanent \\ \cite{luza2021model,matanaluza2021emulating} \\ 
  \cite{gambardella2019efficient,salami2018resilience} \end{tabular}} &
  
  \multicolumn{1}{c|}{\begin{tabular}[c]{@{}c@{}}Permanent \\ \cite{lotfi2019resiliency,condia2022multi,guerrero2022neural} \\
  \cite{guerrero2022effective,guerrero2022evaluating} \end{tabular}} &
  \multicolumn{1}{c||}{} &
  \multicolumn{1}{c|}{} &
  \multicolumn{1}{c|}{} &
   \\   \hline
        \hline

\multicolumn{1}{|c|}{\multirow{2}{*}{\begin{tabular}[c]{@{}c@{}} \\ \\ \textbf{Fault Location}\end{tabular}}} &
  \multicolumn{1}{c|}{\begin{tabular}[c]{@{}c@{}}Weights \cite{arechiga2018effect,arechiga2018robustness} \\ \cite{burel2021zero,burelt2022improving,deligiannis2021towards,cantoro2020evaluating,ghavami2022fitact} \\ 
  \cite{gao2020reliability,goldstein2020reliability,guan2019place,jang2021mate,lee2022bipolar} \\
  \cite{malekzadeh2021impact,neggaz2018reliability,neggaz2019cnns,ozen2021snr,ponader2021milr} \\ 
  \cite{sabbagh2019evaluating,syed2021fault,zhan2021improving,bosio2019reliability} \\
  \cite{lee2022value,ruospo2020evaluating,ruospo2021investigating,yan2020single,}
  \end{tabular}} &
  
  \multicolumn{1}{c|}{\begin{tabular}[c]{@{}c@{}}Weights \\ \cite{ozen2020low,jasemi2020enhancing,kim2019dris} \\ \cite{li2020soft,ozen2020boosting,ozen2020just} \\ 
  \cite{ozen2019sanity,nguyen2021low,tsai2021evaluating} \end{tabular}} &
  
  \multicolumn{1}{c||}{\begin{tabular}[c]{@{}c@{}}PEs, MACs \\ \cite{chitty2020model,liu2021hyca,xu2020hybrid} \\ \cite{zhang2019fault,zhang2018analyzing} \end{tabular}} &
  
  \multicolumn{1}{c|}{\begin{tabular}[c]{@{}c@{}}Configuration \\ Bits \cite{khoshavi2020shieldenn,benevenuti2018comparative} \\ 
  \cite{benevenuti2021neutron,de2020emulation,de2022firenn,du2019reliability} \\ 
  \cite{gao2022reliability,khoshavi2020compression,khoshavi2020fiji} \\ 
  \cite{libano2018selective,libano2020understanding,souvatzoglou2021analyzing} \\
  \cite{wang2021impact,xu2020persistent,xu2021reliability}  \end{tabular}} &
  
  \multicolumn{1}{c|}{\begin{tabular}[c]{@{}c@{}}Registers, \\ Instructions \\ \cite{dos2018analyzing,ibrahim2020softerror,adam2021selective,adam2021analyzing} \\
  \cite{adam2021impact,bolchini2022selective,cavagnero2022transient,condia2021combining} \\
  \cite{dos2022characterizing,garrett2021improving,ibrahim2020analyzing} \\
  \cite{ibrahim2020analyzinginstruction,rech2020impact,dos2017evaluation,dos2019impact} \\
  \cite{condia2022multi,guerrero2022neural,guerrero2022effective,guerrero2022evaluating} \end{tabular}} &
  
  \multicolumn{1}{c||}{\begin{tabular}[c]{@{}c@{}}Register File \\ 
   \cite{abich2020soft,bandeira2019non,abich2021impactprecision,abich2021applying} \\
  \cite{abich2022impactthread,abich2022impact,gava2022soft}  \\
  \cite{liu2022using,liu2022efficient,liu2021analyzing}  \end{tabular}} &
  
  \multicolumn{1}{c|}{\begin{tabular}[c]{@{}c@{}}Entire FPGA \\ Package \cite{benevenuti2018comparative} \\ \cite{benevenuti2021neutron,gambardella2022accelerated,libano2018selective} \\
  \cite{wang2021impact,libano2021reduced,agiakatsikas2021evaluation} \\
  \cite{maillard2022radiation}  \end{tabular}} &
  
  \multicolumn{1}{c|}{\multirow{2}{*}{\begin{tabular}[c]{@{}c@{}}Entire GPU \\  \cite{dos2018analyzing,dos2022characterizing,hari2021making} \\ 
  \cite{dos2017evaluation,dos2019impact} \\
  \cite{basso2020impact,lotfi2019resiliency,} \end{tabular}}} &
  
  \multirow{2}{*}{\begin{tabular}[c]{@{}c@{}}Entire Chip \\  refs \cite{junior2022high,rech2022reliability} \end{tabular}} \\ 
  \cline{2-8} &
  

  \multicolumn{1}{|c|}{\begin{tabular}[c]{@{}c@{}}Activations \cite{ali2020erdnn} \\ \cite{amarnath2022soft,burelt2022improving,chen2021low,hoang2020ft} \\
  \cite{neggaz2018reliability,ozen2021snr} \end{tabular}} &
  
  \multicolumn{1}{c|}{\begin{tabular}[c]{@{}c@{}}Activations \cite{burel2021mozart} \\ \cite{ozen2020low,azizimazreah2018tolerating,goldstein2021lightweight,} \\ 
  \cite{li2017understanding,li2020soft,ozen2020boosting,ozen2020just}  \\
  \cite{ozen2019sanity,burel2022mozart+,hoang2021tre} \\ 
  \cite{siddique2021exploring,tsai2021evaluating,zahid2020fat} \end{tabular}} &
  
  \multicolumn{1}{c||}{\begin{tabular}[c]{@{}c@{}}Registers,\\ Buffers, LUTs \\ \cite{abich2020soft,bandeira2019non,corneliou2021fine,hosseinkhani2021improving}\end{tabular}} &
  
  \multicolumn{1}{c|}{\begin{tabular}[c]{@{}c@{}}On-Chip \\ Memories \cite{khoshavi2020shieldenn} \\ 
  \cite{de2022firenn,gao2022reliability,khoshavi2020compression} \\ 
  \cite{khoshavi2020fiji,luza2021model,matanaluza2021emulating} \\
  \cite{gambardella2019efficient,xu2020persistent,xu2021reliability} \end{tabular}} &
  
  \multicolumn{1}{c|}{\begin{tabular}[c]{@{}c@{}}Weights,\\ Activations \\ \cite{hari2021making,lotfi2019resiliency}\end{tabular}} &
  
  \multicolumn{1}{c||}{\begin{tabular}[c]{@{}c@{}}Instructions \\ 
  \cite{liu2021analyzing} \\ \end{tabular}} &
  
  \multicolumn{1}{c|}{\begin{tabular}[c]{@{}c@{}}HyperRAM \\ \cite{matanaluza2021emulating,luza2020investigating} \end{tabular}} &
  \multicolumn{1}{c|}{} &
   \\   \hline
        \hline
        
\multicolumn{1}{|c|}{\multirow{3}{*}{\begin{tabular}[c]{@{}c@{}} \\ \\ \\ \\ \\  \textbf{Evaluation}\end{tabular}}} &
  \multicolumn{1}{c|}{\begin{tabular}[c]{@{}c@{}}Accuracy Loss \cite{ali2020erdnn} \\ \cite{amarnath2022soft,arechiga2018effect,arechiga2018robustness,burel2021zero,gao2020reliability} \\ 
  \cite{ghavami2022fitact,guan2019place,hoang2020ft,jang2021mate,lee2022bipolar} \\ 
  \cite{malekzadeh2021impact,neggaz2018reliability,neggaz2019cnns,ozen2021snr,ponader2021milr} \\
  \cite{sabbagh2019evaluating,syed2021fault,zhan2021improving,lee2022value,ruospo2021investigating}
  \end{tabular}} &
  
  \multicolumn{1}{c|}{\begin{tabular}[c]{@{}c@{}}Accuracy Loss \\ \cite{burel2021mozart,ozen2020low,goldstein2021lightweight,li2017understanding,li2020soft} \\ 
  \cite{ozen2020boosting,ozen2020just,ozen2019sanity,burel2022mozart+} \\ 
  \cite{hoang2021tre,nguyen2021low,siddique2021exploring} \\
  \cite{tsai2021evaluating,zahid2020fat,zhao2022fsa}  \end{tabular}} &
  
  \multicolumn{1}{c||}{\begin{tabular}[c]{@{}c@{}}Accuracy Loss \\ \cite{chitty2020model,corneliou2021fine,liu2021hyca} \\ \cite{xu2020hybrid,zhang2019fault,zhang2018analyzing} \end{tabular}} &
  
  \multicolumn{1}{c|}{\begin{tabular}[c]{@{}c@{}}Accuracy Loss \\ \cite{khoshavi2020shieldenn,de2020emulation,du2019reliability,,gao2022reliability} \\ 
  \cite{khoshavi2020compression,khoshavi2020fiji,luza2021model} \\ 
  \cite{matanaluza2021emulating,souvatzoglou2021analyzing,salami2018resilience} \\
  \cite{wang2021impact,gambardella2019efficient,xu2020persistent,xu2021reliability} \end{tabular}} &
  
  \multicolumn{1}{c|}{\begin{tabular}[c]{@{}c@{}}Accuracy Loss \\ \cite{cavagnero2022transient,rech2020impact,guerrero2022neural} \\
  \cite{guerrero2022effective,guerrero2022evaluating} \end{tabular}} &
  
  \multicolumn{1}{c||}{\begin{tabular}[c]{@{}c@{}}Fault \\ Classification \\ 
  \cite{abich2020soft,bandeira2019non,abich2021impactprecision,abich2021applying} \\
  \cite{abich2022impactthread,abich2022impact,gava2022soft}  \\
  \cite{liu2022using,liu2022efficient,liu2021analyzing} \end{tabular}} &
  
  \multicolumn{1}{c|}{\begin{tabular}[c]{@{}c@{}}Fault \\ Classification \\ \cite{libano2018selective,libano2021reduced,maillard2022radiation} \end{tabular}} &
  
  \multicolumn{1}{c|}{\begin{tabular}[c]{@{}c@{}}Fault \\ Classification \\ \cite{dos2018analyzing,dos2022characterizing,dos2017evaluation} \\ \cite{dos2019impact,lotfi2019resiliency} \end{tabular}} &
  
  \begin{tabular}[c]{@{}c@{}}Fault \\ Classification \\ 
  \cite{junior2022high,rech2022reliability}  \end{tabular} \\ 
  \cline{2-10} &
  
  \multicolumn{1}{|c|}{\begin{tabular}[c]{@{}c@{}}Fault \\ Classification \\ \cite{burelt2022improving,cantoro2020evaluating,deligiannis2021towards} \\ 
  \cite{ruospo2020evaluating,ruospo2021investigating,bosio2019reliability} \end{tabular}} &
  
  \multicolumn{1}{c|}{\begin{tabular}[c]{@{}c@{}}Fault \\ Classification \\ 
  \cite{li2017understanding}\end{tabular}} &
  
  \multicolumn{1}{c||}{\multirow{2}{*}{\begin{tabular}[c]{@{}c@{}}Fault \\ Classification \\ 
  \cite{abich2020soft,ruospo2020pipelined,bandeira2019non} \\
  \cite{corneliou2021fine,hosseinkhani2021improving}
  \end{tabular}}} &
  
  \multicolumn{1}{c|}{\begin{tabular}[c]{@{}c@{}}Fault \\ Classification \\ \cite{khoshavi2020shieldenn,de2020emulation,de2022firenn,du2019reliability} \\
  \cite{khoshavi2020compression,khoshavi2020fiji,libano2018selective} \\ \cite{libano2020understanding,souvatzoglou2021analyzing,xu2021reliability} \end{tabular}} &
  
  \multicolumn{1}{c|}{\begin{tabular}[c]{@{}c@{}}Fault \\ Classification \\ \cite{dos2018analyzing,ibrahim2020softerror,adam2021selective,adam2021analyzing,adam2021impact} \\ 
  \cite{condia2021combining,dos2022characterizing,garrett2021improving,ibrahim2020analyzing} \\ 
  \cite{ibrahim2020analyzinginstruction,dos2017evaluation,dos2019impact,lotfi2019resiliency} \\
  \cite{condia2022multi,guerrero2022neural,guerrero2022effective,guerrero2022evaluating} 
   \end{tabular}} &
  
  \multicolumn{1}{c||}{\begin{tabular}[c]{@{}c@{}}Reliability \\ Equations  \\
  \cite{abich2020soft,abich2021applying,abich2022impact} \\
  \cite{liu2022efficient,liu2021analyzing}\end{tabular}} &
  
  \multicolumn{1}{c|}{\multirow{2}{*}{\begin{tabular}[c]{@{}c@{}} Reliability \\ Equations  \\ 
  \cite{benevenuti2018comparative,benevenuti2021neutron,libano2018selective} \\ 
  \cite{matanaluza2021emulating,wang2021impact} \\ 
  \cite{agiakatsikas2021evaluation,libano2021reduced,luza2020investigating} \\ \end{tabular}}} &
  
   \multicolumn{1}{c|}{\multirow{2}{*}{\begin{tabular}[c]{@{}c@{}}FIT Rate \\ 
   \cite{dos2018analyzing,dos2022characterizing} \\
   \cite{dos2017evaluation,dos2019impact} \end{tabular}}} &
   
  \multirow{2}{*}{\begin{tabular}[c]{@{}c@{}}FIT Rate \\
  \cite{junior2022high,rech2022reliability}  \end{tabular}} \\ 
  \cline{2-3} \cline{5-7} &
  
  \multicolumn{1}{|c|}{\begin{tabular}[c]{@{}c@{}}SDC Rate \\ \cite{chen2021low,goldstein2020reliability}\end{tabular}} &
  
  \multicolumn{1}{c|}{\begin{tabular}[c]{@{}c@{}}SDC Rate \\ \cite{azizimazreah2018tolerating,jasemi2020enhancing}\end{tabular}} &
  
  \multicolumn{1}{c||}{} &
  \multicolumn{1}{c|}{\begin{tabular}[c]{@{}c@{}}Reliability \\ Equations \\ \cite{benevenuti2018comparative,benevenuti2021neutron,libano2018selective} \\ \cite{libano2020understanding,luza2021model}\end{tabular}} &
  
  \multicolumn{1}{c|}{\begin{tabular}[c]{@{}c@{}}Vulnerability \\ Factors \cite{dos2018analyzing,ibrahim2020softerror} \\ \cite{adam2021selective,adam2021analyzing,bolchini2022selective,cavagnero2022transient} \\
  \cite{condia2021combining,garrett2021improving,ibrahim2020analyzing} \\ 
  \cite{ibrahim2020analyzinginstruction,dos2017evaluation,dos2019impact} \end{tabular}} &
  
  \multicolumn{1}{c||}{\begin{tabular}[c]{@{}c@{}}Vulnerability \\ Factors \\
  \cite{liu2022efficient,liu2021analyzing}\end{tabular}} &
  
  \multicolumn{1}{c|}{} &
  \multicolumn{1}{c|}{} &
   \\ \hline
\end{tabular}%
\centering
\label{tab:FI-cat}
\end{adjustbox}
\end{table*}

\subsubsection{\textbf{Analytical Methods}}
Works relying on an analytical method for estimating DNNs' reliability attempt to determine how parameters and neurons of a DNN affect the output based on the connections of neurons and layers. Therefore, they analyze the structure of DNNs and provide a model for the impact of faults on the outputs to find more critical and sensitive components in the DNN. Hence, they can evaluate the reliability of DNNs by means of vulnerability analysis derived by analyses, and eliminate the complexity of simulating/emulating the faults in reliability assessment.

\subsubsection{\textbf{Hybrid Methods}}
Both, fault injection and analytical methods are used in this category of works to take advantage of both. In this regard, analytical methods can provide some mathematical models in addition to a straight-forward fault injection into the system for reliability evaluation, so that metrics of reliability evaluation can be obtained with less complexity than extensive FI experiments and more realistic than analytical methods.

\subsection{Research Trends}
\begin{figure}[h]
\includegraphics[width=0.5\textwidth]{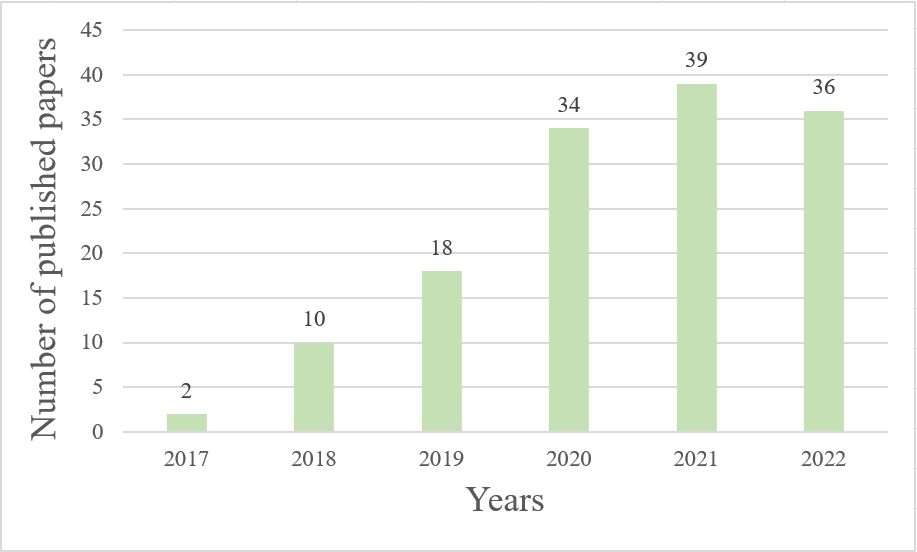}
\centering
\caption {Number of included papers over years}
\label{fig:published-years}
\end{figure}

To address RQ1, we present the main statistics on the papers included in this study. Fig.~\ref{fig:published-years} shows the distribution of the 139 included papers published over  years 2017-2022. Regarding the chart of Fig.~\ref{fig:published-years}, it can be seen that research on the topic of DNNs' reliability started in 2017 and in the following years it drew increasingly more attention and turned into an active topic of study. 

\begin{figure}[h]
\includegraphics[width=0.35\textwidth]{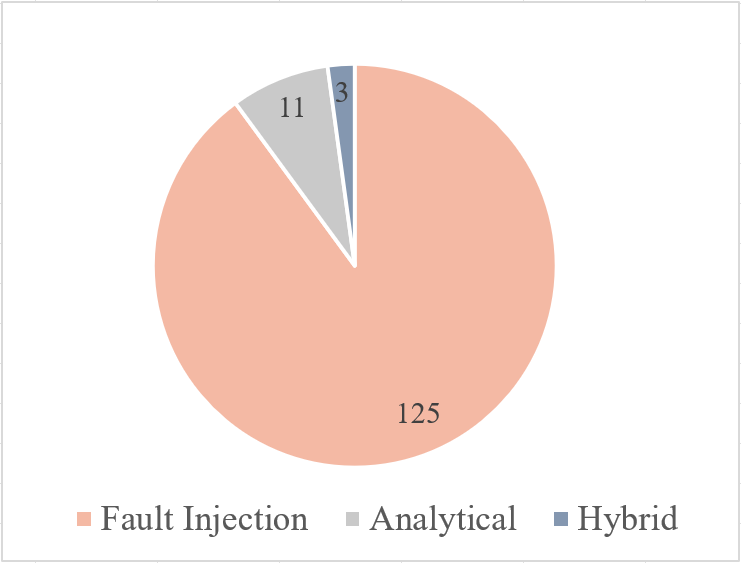}
\centering
\caption {Proportion of each method in the reliability assessment of DNNs among included works}
\label{fig:methods-dist}
\end{figure}

\begin{figure}[ht]
\includegraphics[width=0.6\textwidth]{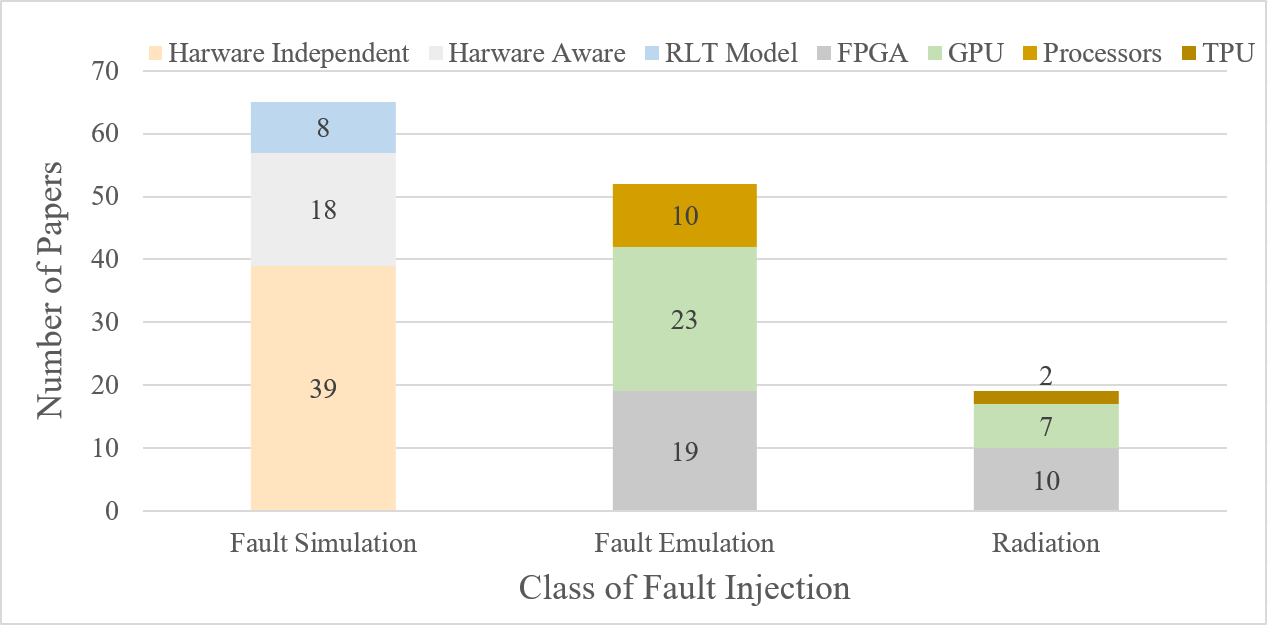}
\centering
\caption {Distribution of included papers over different FI approaches and platforms}
\label{fig:FI-dist}
\end{figure}

Fig.~\ref{fig:methods-dist} illustrates the number of papers based on different reliability assessment methods among all identified works in this literature review. It can be observed that the majority of works use fault injection to assess the reliability of DNNs while only 10\% of the works consider analytical (11 works) and hybrid analytical/FI (3 works) methods. In this regard, we present Fig.~\ref{fig:FI-dist} to illustrate the distribution of works using FI over different approaches and DNN platforms. It shows that most of the works belong to the hardware-independent platform of simulation in the software approach. Moreover, in the emulation in hardware approach, most of the works are done on the GPU platform. Hence, the figures present the trend of research domain, and distribution of works over different methods and approaches leading to areas where there is still room for future research.

\section{Characterization} \label{charaterization}
In this Section, details of reliability assessment methods for DNNs are presented based on the categorizations in Fig.~\ref{fig:top-level}, and  
Table \ref{tab:FI-cat}. We start from FI methods which include the majority of works. Then, analytical and hybrid methods will be discussed.

\subsection{\large{Fault Injection Methods}} \label{char-FI}
In FI methods of reliability assessment, once the DNN platform and fault model are determined, perturbation and system execution are performed, and the reliability is evaluated. Regarding the categorization in  
Table \ref{tab:FI-cat}, the identified approaches of FI methods on DNN reliability assessment are presented in this subsection, separately. Since FI is the most frequently used method in the reliability assessment of DNNs, there are various presented evaluation metrics. To elaborate and distinguish different evaluation metrics, we have presented them for different approaches and platforms, separately.

\subsubsection{\large{\textbf{Fault Simulation}}}  

In this subsection, the works assessing the reliability of DNNs by FI with a fault simulation approach are described. There are three platforms in this approach i.e., hardware-independent, hardware-aware, and RTL models that are explained in the corresponding subsections.

\paragraph{\textbf{5.1.1.1 \hspace{1mm} Hardware-Independent Platform}}

In this platform, DNNs are implemented in software DNN frameworks. Therefore, fault injection is performed on top of the frameworks, i.e., PyTorch (used in \cite{ali2020erdnn}\cite{gao2020reliability}\cite{goldstein2020reliability}\cite{guan2019place}\cite{hoang2020ft}\cite{neggaz2018reliability}\cite{ozen2021snr}), Keras (used in \cite{arechiga2018effect}\cite{arechiga2018robustness}\cite{sabbagh2019evaluating}\cite{syed2021fault}), TensorFlow (used in \cite{chen2021low}\cite{ponader2021milr}), Caffe (used in \cite{neggaz2019cnns}), DarkNet (used in \cite{jang2021mate}\cite{bosio2019reliability}\cite{ruospo2020evaluating}). Implementing the DNN in software provides a flexible environment for studying the effect of various fault models. As shown in the corresponding branch of Table \ref{tab:FI-cat}, both transient and permanent faults are studied in this platform. However, most of the works studied transient faults (soft errors, SEU, MBU, etc.). 

To model faults at the software level, the fault model is determined differently regarding the fault type and general aspect of DHAs. In this regard, modeling and injecting permanent faults are straight-forward. They are active throughout the entire execution and set the value of a bit or variable (in weights, or activations) to 0 or 1, as experimented in \cite{hoang2020ft}\cite{bosio2019reliability}\cite{ruospo2020evaluating}. To model transient faults, the following assumptions are considered for injecting faults into parameters, i.e.:

\begin{itemize}
    \item DNN's parameters (e.g., weights) are stored in the memory of accelerator. Hence, random transient faults are injected into random bits of weights as a bitflip, and the faulty value remains until it gets overwritten, are experimented in \cite{arechiga2018effect}\cite{arechiga2018robustness}\cite{burel2021zero}\cite{burelt2022improving}\cite{cantoro2020evaluating}\cite{deligiannis2021towards}\cite{gao2020reliability}\cite{ghavami2022fitact}\cite{goldstein2020reliability}\cite{guan2019place} \cite{jang2021mate}\cite{lee2022bipolar}\cite{malekzadeh2021impact}\cite{neggaz2018reliability}\cite{neggaz2019cnns}\cite{ozen2021snr}\cite{ponader2021milr}\cite{sabbagh2019evaluating}\cite{syed2021fault}\cite{zhan2021improving}\cite{lee2022value}\cite{ruospo2021investigating}\cite{yan2020single}.
    
    \item Faults in inputs/outputs of DNN's layers (i.e, activations) lead to study their impacts on both memory and logic. Activation memory faults are studied in \cite{hoang2020ft}\cite{neggaz2018reliability}, and faults in logic or datapath are investigated in \cite{ali2020erdnn}\cite{amarnath2022soft}\cite{burelt2022improving}\cite{chen2021low}\cite{ozen2021snr}.
\end{itemize}

Therefore, to experiment the impact of faults on memory elements of DHAs at software level, faults are injected into random weights and activations, and to model fault effects on logic, faults are injected into random activations. Most of relevant works on Hardware-Independent platform inject transient faults into the bits of randomly selected weights. Nearly all works in this class, inject faults based on BER which determines how large portion of all the bits are faulty. In addition, to reach the 95\% confidence level with 1\% error margin, they repeat the tests several times with different random faults as in \cite{bosio2019reliability}\cite{neggaz2019cnns}\cite{ruospo2020evaluating}\cite{sabbagh2019evaluating}.

\textbf{\textit{Evaluation}:} For evaluating the reliability, different metrics are considered. References \cite{ali2020erdnn}\cite{amarnath2022soft}\cite{arechiga2018effect} \cite{arechiga2018robustness}\cite{burel2021zero}\cite{gao2020reliability}\cite{ghavami2022fitact}\cite{guan2019place}\cite{hoang2020ft}\cite{jang2021mate}\cite{lee2022bipolar}\cite{malekzadeh2021impact}\cite{neggaz2018reliability}\cite{neggaz2019cnns}\cite{ozen2021snr}\cite{ponader2021milr}\cite{sabbagh2019evaluating}\cite{syed2021fault}\cite{zhan2021improving}\cite{lee2022value}\cite{ruospo2021investigating} report accuracy loss under fault campaign experiments. They compare the accuracy of the faulty network with the accuracy of the fault-free network on the same test set. Some works classify the injected faults regarding the outputs of the faulty network compared with the golden model output. References \cite{bosio2019reliability}\cite{ruospo2020evaluating}\cite{ruospo2021investigating} inject one permanent fault per experiment and classify them into three classes:

\begin{itemize}
    \item \textbf{Masked:} No difference between the outputs of the faulty network and the golden model.
    \item \textbf{Observed-Safe:} Different output of the faulty network with the golden model, while the confidence score of the top-ranked element is reduced by less than 5\% with respect to the one of the golden one.
    \item \textbf{Observed-Unsafe:} Different output of faulty network with the golden model, while the confidence score of the top-ranked element is reduced by more than 5\% with respect to the one of the golden one.
\end{itemize}

Moreover, in \cite{cantoro2020evaluating}\cite{deligiannis2021towards} transient faults are injected into the encrypted weights of a network and they are classified based on the effect of faults on execution of the program and results, as:

\begin{itemize}
    \item \textbf{Silent or safe:} Similar to "masked" mentioned above in \cite{bosio2019reliability}\cite{ruospo2020evaluating}.
    \item \textbf{SDC:} Only affects the output results of the network.
    \item \textbf{Detected as a software exception:} Affects the execution of the program and stops it.
    \item \textbf{Detected by padding check action:} Corrupts the ciphertext.
\end{itemize}

Burel et al. \cite{burelt2022improving} have adopted the fault classification scheme for semantic segmentation applications in which DNNs label each pixel of an input image according to a set of known classes. The corresponding classes are:

\begin{itemize}
    \item \textbf{Masked:} Similar to "masked" mentioned above.
    \item \textbf{No Impact SDC:} No labels of pixels are modified.
    \item \textbf{Tolerable SDC:} Labels of less than 1\% of pixels are modified and no class is removed/added due to the fault.
    \item \textbf{Critical SDC:} Labels of more than 1\% of pixels are modified or any class is removed/added due to the fault.
\end{itemize}

A specific way of fault evaluation based on fault classification is only considering the faults which affect the output as SDC, since they are critical. References \cite{chen2021low}\cite{goldstein2020reliability} evaluate the network based on the proportion of faults that affect the output classification results as SDC rate. Therefore, the reliability of a network can be evaluated by fault classification based on their effect on the outputs, whether by changing the output results, or by a threshold of accuracy loss, or system exceptions. This way of evaluation assists in understanding how faults would be propagated and affect the network.

\textbf{\textit{Software FI Tools}:} Some fault injectors are presented as tools that are able to support the reliability study of DNNs with different fault models in software frameworks of DNNs. PyTorchFI \cite{mahmoud2020pytorchfi}, TensorFI \cite{chen2020tensorfi}\cite{li2018tensorfi}\cite{narayanan2022fault} and its extension TensorFI+ \cite{laskar2022tensorfi+}\cite{laskar2022characterizing}, and Ares \cite{reagen2018ares} inject faults into DNNs which are implemented in PyTorch, Tensorflow, and Keras, respectively. All of these open-source frameworks can inject, both, permanent and transient faults into weights as well as activations with specified error rates, hence, the accuracy loss can be evaluated. TensorFI also benefits from providing the SDC rate. These frameworks are used in the reliability studies of DNNs, e.g., PyTorchFI in \cite{amarnath2022soft}\cite{goldstein2020reliability}, TensorFI in \cite{chen2021low}, and Ares in \cite{sabbagh2019evaluating}.

Moreover, to enhance the efficiency of the aforementioned tools, additional fault injectors have been introduced. One such injector, known as BinFI \cite{chen2019binfi}, is an extension of TensorFI that aims to identify critical bits in DNNs. Another fault injector, namely LLTFI \cite{agarwal2022lltfi}, is proposed to inject transient faults into specific instructions of DNN models in either PyTorch or TensorFlow and has been found to be faster than TensorFI. Additionally, a check-point based fault injector is proposed in \cite{rojas2021understanding} that enables studying the impact of SDCs independently of the DNN implementation framework.

\paragraph{\textbf{5.1.1.2 \hspace{1mm} Hardware-Aware Platform}} \label{char-fi-hap}

This platform includes works that consider an abstract model of the accelerator in their implementation of DNNs in software. They implement the network in DNN software frameworks as well as high-level programming languages. Therefore, they take advantages of simulation in software fault injection while they also apply the reliability assessment to the abstract model of the accelerator.

References \cite{azizimazreah2018tolerating}\cite{li2017understanding} implement a DNN in Tiny-DNN, and map it to the RTL implementation of the accelerator. They study the effect of transient faults in memory and datapath accurately. In these studies, FI is performed in software while all of its parameters are integrated with the corresponding hardware components. Authors in \cite{li2020soft} implement the DNN and the fault injector in software inspired by an FPGA-based DNN accelerator. Moreover, in \cite{ozen2020low}\cite{ozen2019sanity} DNN and FI are implemented in Keras, and the architecture of a systolic array accelerator is considered for a fault-tolerant design. Similarly, authors in \cite{jasemi2020enhancing} and \cite{kim2019dris}  evaluate their proposed reliability improvement technique on memories in TensorFlow while injecting transient faults into the weights. PyTorch is used in \cite{ozen2020boosting}\cite{ozen2020just} to implement the DNN, and transient faults are injected into activations (datapath or MAC units) and weights (memory) regarding the systolic array accelerator model. Reference \cite{goldstein2021lightweight} also uses PyTorch and injects faults by a custom framework called TorchFI to inject faults into the outputs of CONV and FC layers of the network.

The effect of permanent faults at PEs' outputs is studied in \cite{burel2021mozart}\cite{burel2022mozart+} where the model of the accelerator is adopted from implementing the DNN in an N2D2 framework \cite{N2D2_fr}. Furthermore, authors in \cite{hoang2021tre}\cite{zahid2020fat} use PyTorch and study permanent faults in MAC units of an accelerator while training to improve the reliability at inference. Authors in \cite{tsai2021evaluating} have developed a Keras-based accelerator simulator to study the effect of permanent faults on the on-chip memory of accelerators by injecting permanent faults into fmaps and weights. Weight remapping strategy in memory to decrease the effect of permanent faults is evaluated in \cite{nguyen2021low} using Ares. SCALE-Sim \cite{samajdar2018scale}, a systolic CNN accelerator simulator, is adopted in \cite{zhao2022fsa} to study permanent faults in PEs and computing arrays in systolic array-based accelerators.

Similar to the Hardware-Independent platform, faults are injected based on BER, or fault rate, and experiments are repeated to reach 95\% confidence level and 1\% error margin \cite{ozen2020low}\cite{li2017understanding}\cite{ozen2019sanity}. 

\textbf{\textit{Evaluation}:} Nearly all works in this class, evaluate the DNN by accuracy loss after fault injection \cite{burel2021mozart}\cite{ozen2020low}\cite{goldstein2021lightweight}\cite{kim2019dris} \cite{li2020soft}\cite{ozen2020boosting}\cite{ozen2020just}\cite{ozen2019sanity}\cite{burel2022mozart+}\cite{nguyen2021low}\cite{siddique2021exploring}\cite{tsai2021evaluating}\cite{zahid2020fat}\cite{zhao2022fsa}. References \cite{azizimazreah2018tolerating} and \cite{jasemi2020enhancing} evaluate the reliability by SDC rate as the proportion of faults that caused misclassification in comparison with the golden model. In addition, authors in \cite{li2017understanding} differentiate SDCs of injected transient faults into defined classes and calculate FIT for the accelerator (\textit{accel}) by its components (\textit{comp}) with \eqref{eq:FIT1} in which \(FIT_{raw}\) is provided by the manufacturer, \(Size_{comp}\) is the total number of the component bits, and \(SDC_{comp}\) is obtained by FI.
\begin{equation}
FIT_{accel} = \sum_{comp} FIT_{raw} \times Size_{comp} \times SDC_{comp} 
\label{eq:FIT1}
\end{equation}

In addition, in this work SDCs are classified by comparing the faulty and golden model outputs as:

\begin{itemize}
    \item \textbf{SDC-1:} Fault caused a misclassification in the top-ranked output class.
    \item \textbf{SDC-5:} Fault caused the top-ranked element not to exist in the top-5 predicted output classes.
    \item \textbf{SDC-10\%:} Fault caused a variation in the output confidence score of the top-ranked output class more than 10\% compared to the golden model.
    \item \textbf{SDC-20\%:} Fault caused a variation in the output confidence score of the top-ranked output class more than 20\% compared to the golden model.
\end{itemize}

\paragraph{\textbf{5.1.1.3 \hspace{1mm} RTL Model Platform}} 

Research works that leverage the RTL model of ASIC-based DHAs and simulate fault injections are described in the following. We identify three groups of FI experiments in this platform, divided based on the architecture of DHAs: 

\begin{itemize}
    \item 2D systolic array accelerators \cite{chitty2020model}\cite{corneliou2021fine}\cite{liu2021hyca}\cite{xu2020hybrid}\cite{zhang2019fault}\cite{zhang2018analyzing},
    \item RTL implementation of DNNs \cite{hosseinkhani2021improving},
    \item Multi-Processor System-on-Chips (MPSoCs) for DNNs, 
    \cite{ruospo2020pipelined}.
\end{itemize}

In the first group, a configuration of TPU is utilized in \cite{chitty2020model}\cite{corneliou2021fine}\cite{zhang2019fault}\cite{zhang2018analyzing}, and a model of a 2D systolic array is implemented in \cite{liu2021hyca}\cite{xu2020hybrid}. Reference \cite{chitty2020model} also uses Eyeriss \cite{chen2016eyeriss} architecture for the accelerator. In this group, FI is performed at RTL, and all works inject random permanent faults into PEs/MACs of the arrays, except \cite{corneliou2021fine} which injects random transient faults into buffers, control and data registers. 

The second group which includes \cite{hosseinkhani2021improving} implements DNNs in RTL to enable a fault simulation study in approximated DNNs. In this work, SEU injected into Look-Up Tables are simulated and studied.

In the third group which exploits MPSoCs, faults are emulated in the components of the target multicore processor. 
Authors in \cite{ruospo2020pipelined} propose a three-level pipeline FI framework that simulates permanent faults in the hardware model of an MPSoC and evaluate the reliability at the software level. In their framework, the RTL model of the platform is provided as well as the fault injector unit at the lowest level. The software implementation of the DNN exists in the middle level of the framework that performs a pipelined inference and runs each layer of the network on a separate core. In the top-level of the framework, synchronization of layers and reliability evaluation is fulfilled.

\textbf{\textit{Evaluation}:} Most works in this class evaluate the reliability by accuracy loss. Nonetheless, fault classification is performed in 
\cite{corneliou2021fine}\cite{hosseinkhani2021improving}\cite{ruospo2020pipelined}. Authors in \cite{ruospo2020pipelined} adopted the classification of \cite{li2017understanding} which was discussed in Hardware-Aware platform (subsection \ref{char-fi-hap}) previously. Furthermore, they added two more classes for the faults that cause Hang (the HDL simulation never finishes) and Crash (the HDL simulation immediately stops). Authors in \cite{hosseinkhani2021improving} classify the faults similar to the general fault classification scheme (Masked, SDC, crash) with different terminology.



In addition, \cite{corneliou2021fine} classifies SDCs on how they impact classification outputs compared with the golden model:

\begin{itemize}
    \item \textbf{Tolerable Misclassification:} The input is misclassified the same as the golden model with different output confidence scores,  
    \item \textbf{No Impact Misclassification:} The input is misclassified in both golden and faulty models but into different classes, 
    \item \textbf{Critical Misclassification:} The input is correctly classified in the golden model but misclassified in the faulty model,
    \item \textbf{Tolerable Correct Classification:} The input is correctly classified in both golden and faulty models with different output confidence scores,
    \item \textbf{Beneficial Correct Classification:} The input is misclassified in golden model but correctly classified in the faulty model.
\end{itemize}

\subsubsection{\large{Fault Emulation}} 

In this subsection, research works that assess the reliability of DNNs by emulating FI in hardware accelerators are explored. FPGA and GPU platforms are described, respectively.

\paragraph{\textbf{5.1.2.1 \hspace{1mm} FPGA Platform}}

DNNs are implemented fully or partially (e.g., one layer) on FPGAs to perform the inference phase as described in subsection \ref{pre-platforms}, and faults are being emulated on different locations of the accelerator. In most of the works on the FPGA platform, the fault injector unit is implemented in software that is run on a processor and faults are injected into the FPGA running the DNN under analysis. This HW/SW co-design process benefits from the high-performance execution of DNNs and fast fault injection. It is worth mentioning that some works implement only a part of the DNN (e.g., one specific layer) on the FPGA \cite{de2020emulation}\cite{de2022firenn}\cite{wang2021impact}. 

In this group of works, Zynq-based architecture System-on-Chips (SoCs) \cite{zynq_xilinx} which take advantage of an ARM processor co-existing with the FPGA are deployed. We categorize this group of studies into three classes:

\begin{itemize}
    \item A host computer (e.g., a PC) initializes the faults \cite{de2020emulation}\cite{de2022firenn}\cite{du2019reliability}\cite{souvatzoglou2021analyzing}\cite{wang2021impact},
    \item The on-board embedded processor initializes the faults \cite{khoshavi2020shieldenn}\cite{benevenuti2018comparative}\cite{gao2022reliability}\cite{khoshavi2020compression}\cite{khoshavi2020fiji}\cite{libano2018selective}\cite{libano2020understanding} \cite{luza2021model}\cite{matanaluza2021emulating}\cite{xu2020persistent}\cite{xu2021reliability},
    \item Fault injection module resides inside the hardware design implementation \cite{benevenuti2021neutron}\cite{gambardella2019efficient}\cite{salami2018resilience}.
\end{itemize}

In the first class, faults are generated by a host computer of the accelerator design. Then, the faults, network parameters, and FPGA configuration bits will be sent to the board. The FPGA starts running, and the on-board processor would collect the results. The on-board processor is playing the role of a controller between FPGA and the host computer. At the end, the results would be passed back to the host computer for further processing and reliability evaluation. All works of this class emulate transient faults (SEU) in configuration bits of the FPGA and exploit the accuracy loss of the DNN for the reliability evaluation. Nevertheless, authors in \cite{souvatzoglou2021analyzing} explore transient faults in flipflops exhaustively beside random transient faults in configuration memory, and classify them as tolerable, critical, and crashes.

\begin{figure*}[h]
\includegraphics[width=0.5\textwidth]{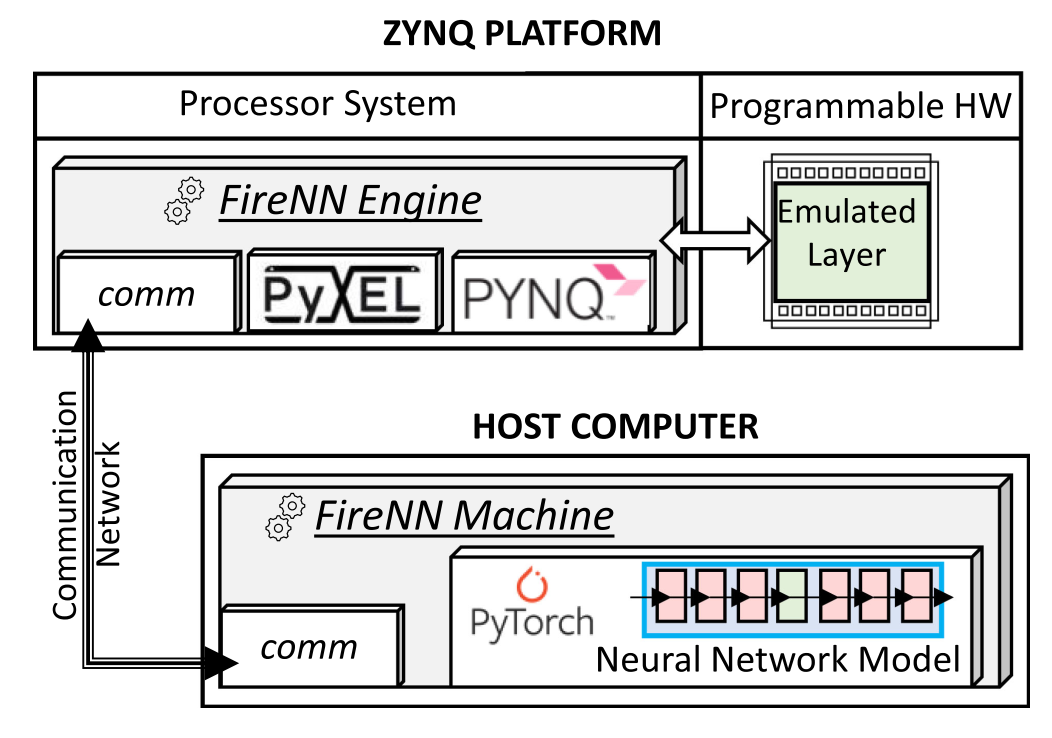}
\centering
\caption {An overview of the architecture of the FireNN platform \cite{de2020emulation}\cite{de2022firenn}.}
\label{fig:fireNN}
\end{figure*}

FireNN is proposed in \cite{de2020emulation}\cite{de2022firenn} as a platform for deploying DNNs on Zynq-based architecture SoCs along with a host computer in a way that DNN is run partially on the FPGA to perform a reliability evaluation. As shown in Fig.~\ref{fig:fireNN} \textit{FireNN machine} runs the neural network and communicates with the \textit{FireNN engine} for reliability evaluation of the  layer under analysis running on the FPGA. Faults are generated by the host computer and are injected to the FPGA through the engine. This platform injects SEUs in weights, layer inputs, and configuration bits. 

In the second class, faults are generated and injected into the FPGA's configuration bits or on-chip memories by the embedded processor. The embedded processor or a host computer is responsible for the reliability evaluation. The proposed method in \cite{xu2020persistent}\cite{xu2021reliability} provides an injection of permanent faults into the configuration bits of the FPGA as well as into the on-chip memory blocks through the interfaces between the embedded processor and FPGA on Zynq SoC. References \cite{benevenuti2018comparative}\cite{libano2018selective}\cite{libano2020understanding} provide a similar design to inject transient faults into configuration bits of the FPGA. The effects of transient faults into both, on-chip memories and configuration bits of an FPGA running pruned DNNs are studied in \cite{gao2022reliability}. Authors in \cite{benevenuti2018comparative} provide random-accumulated FI and exhaustive FI approach on the configuration bits to emulate neutron and ionizing radiation. Moreover, permanent and transient faults in on-chip memory (HyperRAM) are studied in \cite{luza2021model}\cite{matanaluza2021emulating} with a software emulator and are validated by radiation results. 

It is worth mentioning that injecting faults into the configuration memory is a repetitive process, where in each experiment of FI, the faulty configuration bits are loaded to the configuration memory. Then, the system is run and the results are collected. Thereafter, the next fault(s) are injected into the fault-free configuration bits loaded to the corresponding memory to analyze the newly injected fault(s).

\begin{figure*}[h]
\includegraphics[width=\textwidth]{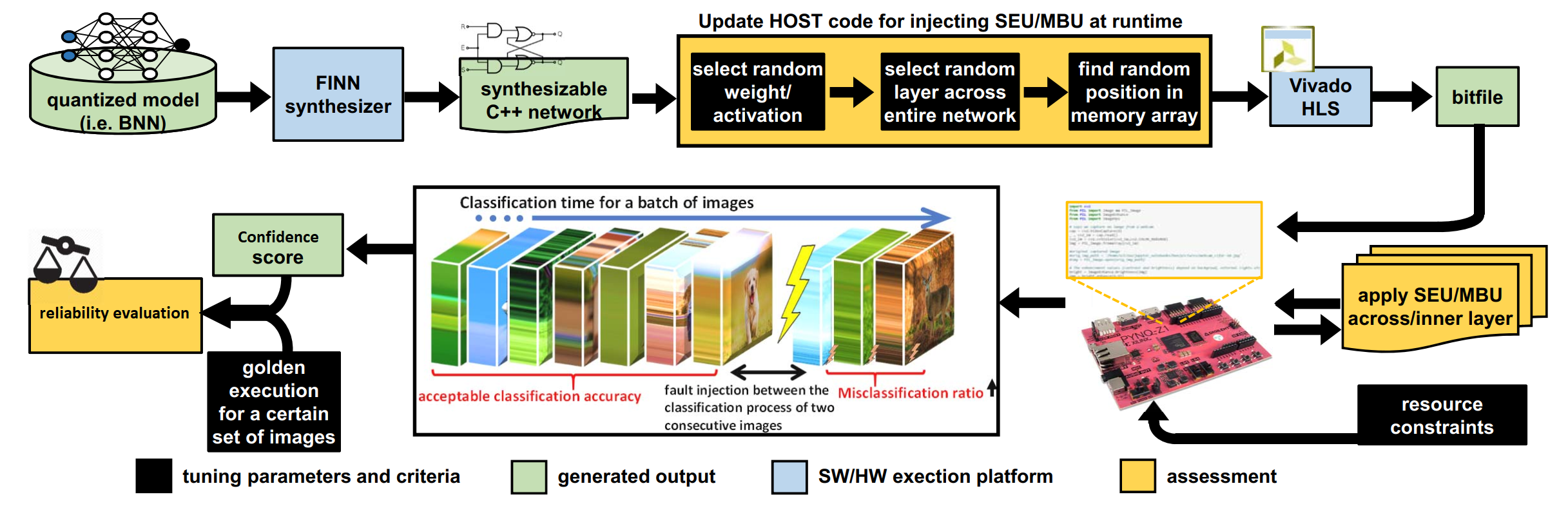}
\centering
\caption {Fiji-FIN framework for fault injection into FPGAs \cite{khoshavi2020fiji}}
\label{fig:fiji}
\end{figure*}

A framework named Fiji-FIN is proposed in \cite{khoshavi2020fiji} and the underlying method is also used in \cite{khoshavi2020shieldenn}\cite{khoshavi2020compression}. This framework is capable of injecting transient faults into both, configuration bits of FPGA and on-chip memories. In this method, FINN framework \cite{umuroglu2017finn} is used to develop and train the BNN, and the proposed framework manipulates the FINN's output to prepare it for the fault injection. The bit stream file of the FPGA is obtained by an HLS tool and imported to the FPGA. While the system is running, the faults are generated and injected by the embedded processor and the reliability is evaluated in comparison with the golden model. Fig.~\ref{fig:fiji} depicts in detail the steps of this FI framework.

In the third class, references \cite{gambardella2019efficient} and \cite{salami2018resilience} inject permanent faults and the work in \cite{benevenuti2021neutron} injects transient faults into the hardware implementation of the network. Authors in \cite{gambardella2019efficient} use the FINN framework to implement the QNN with 2-bit weights and activations, and a block has been added into the hardware design that is deployed for injecting stuck-at faults into the output of PEs. Reference \cite{salami2018resilience} injects permanent faults into the registers of the RTL model of the network. Authors in \cite{benevenuti2021neutron} explore the effect of transient faults to the configuration bits of FPGAs in which different accelerator architectures (Softcore FGPU and ZynqNet HLS) are implemented. 

\textbf{\textit{Evaluation}:} For evaluating the reliability of DNNs on the FPGA platform, accuracy loss is exploited in \cite{khoshavi2020shieldenn}\cite{gao2022reliability}\cite{khoshavi2020compression}\cite{khoshavi2020fiji}\cite{matanaluza2021emulating}\cite{wang2021impact}\cite{gambardella2019efficient}\cite{salami2018resilience}\cite{xu2020persistent}\cite{xu2021reliability}. Moreover, fault classification is also performed in \cite{khoshavi2020shieldenn}\cite{de2020emulation}\cite{de2022firenn}\cite{du2019reliability}\cite{khoshavi2020compression}\cite{libano2018selective}\cite{libano2020understanding}\cite{xu2021reliability}. References \cite{libano2018selective}\cite{libano2020understanding} classify SEUs in configuration bits of the FPGA as critical if a fault caused misclassification with respect to the golden model; otherwise, the fault is tolerable. In addition, Benign Errors are considered in \cite{libano2020understanding} which are the faults that caused true classification of the inputs that were misclassified in the golden model. Another fault classification is presented in \cite{de2020emulation}\cite{de2022firenn} that does not only consider critical and tolerable faults, but also categorizes the faults that prevent the accelerator to generate the classification output. In this regard, the effect of faults on the system performance degradation is the criterion for classifying faults in \cite{du2019reliability}.

Reliability is evaluated by different metrics considering accuracy loss regarding the application of the target networks in \cite{xu2020persistent}\cite{xu2021reliability}. These works consider top-5 and top-1 accuracy loss for image and audio classification tasks, respectively. For object detection, mean Average Precision (mAP), and for image generation, Structural Similarity Index (SSIM) is adopted. Regarding the adopted metrics for accuracy loss in each network, the faults are classified into three classes with different ranges of accuracy loss ($\leq$1\%, 1\%$\sim$5\%, $\geq$5\%) caused by FI. In addition, they categorize the faults which are caused by a system exception that may delay or terminate processes.

To characterize the status of DNN layers' vulnerability, authors in \cite{khoshavi2020shieldenn} classify the parameters of layers (i.e., weights and activations) separately by performing FI. In this work, parameters of layers are labeled as Low-risk, Medium-risk, and High-risk if FI process into the target layers' parameters results in less than 1\%, 1\%$\sim$5\%, and more than 5\% accuracy loss, respectively.

The metric AVF (defined in \ref{pre-definitions}) is adopted in \cite{libano2018selective}\cite{libano2020understanding} and expresses the probability of fault propagating to the output. These works obtain the AVF through the FI, by dividing the number of faults propagated to the output by the total number of injected faults. Furthermore, authors in \cite{libano2020understanding} provide a formula to estimate the cross-section (defined in \ref{pre-definitions}) of the configuration memory in \eqref{eq:cross_section_FPGA} where the obtained \textit{AVF} by FI is multiplied by the number of bits utilized by the design times the cross-section of bits of the configuration memory. This calculation can lead to further reliability metrics that authors present in \cite{libano2020understanding}.
\begin{equation}
\sigma = AVF \times (\#UtilizedBits) \times (\frac{\sigma_{static}}{\#MemBits})
\label{eq:cross_section_FPGA}
\end{equation}

In this regard, \cite{luza2021model} estimates the SER of HyperRam saving the weights similar to \eqref{eq:cross_section_FPGA} based on the extracted information from radiation experiment reports. By providing the rate of faults likely to occur in the memory, they inject faults into the weights of CNN on an FPGA accelerator.

Moreover, reference \cite{benevenuti2018comparative} expressed the reliability of the neural network with \textit{n} layers (\(L_1\), \(L_2\), ..., \(L_n\)) that are implemented serially as different modules on the FPGA, as an exponential distribution in \eqref{eq:reliability_FPGA}.
\begin{equation}
R_{NN}(t) = e^{-(\lambda_{L_1} + \lambda_{L_2} + ... + \lambda_{L_n})t}
\label{eq:reliability_FPGA}
\end{equation}
Where \(\lambda = \frac{1}{MTTF}\) (MTTF = Mean Time to Failure).

\paragraph{\textbf{5.1.2.2 \hspace{1mm} GPU Platform}}

In this subsection, we explore FI in DNNs in which faults are emulated and injected into the GPU. Nearly all works on this platform have studied the effect of transient faults on GPUs. Permanent faults are studied in \cite{lotfi2019resiliency}\cite{condia2022multi}\cite{guerrero2022neural}\cite{guerrero2022effective}\cite{guerrero2022evaluating}\cite{guerrero2022reliability}. To perform FI on GPUs, researchers adopt an FI framework on GPUs; except in \cite{hari2021making}\cite{lotfi2019resiliency} which implemented their own FI process on CUDA and TensorRT \cite{tensorrt}, respectively. FI frameworks in GPUs including FlexGripPlus \cite{condia2020flexgripplus}, NVBitFI \cite{tsai2021nvbitfi}, and CAROL-FI \cite{oliveira2017experimental} are used in \cite{condia2021combining,condia2022multi}, \cite{cavagnero2022transient,dos2022characterizing,garrett2021improving,rech2020impact}, and \cite{dos2019impact}, respectively. Nonetheless, an FI framework is proposed in \cite{guerrero2022reliability} adapting and customizing NVBitFI for studying permanent faults in GPUs and is leveraged in \cite{guerrero2022neural}\cite{guerrero2022effective}\cite{guerrero2022evaluating}. Moreover, a cross-layer fault injector framework CLASSES is presented in \cite{bolchini2022fast} to inject SEUs at the architecture level enabling study of the corresponding fault effects in \cite{bolchini2022selective}. In all works, the rate of injected faults and the number of experiments in the target locations varies and depends on the confidence level and error margin as mentioned in \cite{dos2018analyzing}\cite{ibrahim2020softerror}\cite{adam2021selective}\cite{dos2017evaluation}\cite{dos2019impact}. 

SASSIFI \cite{hari2017sassifi} is the most frequently used framework for FI into GPUs running DNNs that is used in \cite{dos2018analyzing}\cite{ibrahim2020softerror}\cite{adam2021selective}\cite{adam2021analyzing}\cite{adam2021impact}\cite{ibrahim2020analyzing}\cite{ibrahim2020analyzinginstruction}\cite{dos2017evaluation}. This framework is developed by NVIDIA to conduct fault injections and is a powerful framework with different fault models covering various locations of GPUs and provides extensive reliability evaluation metrics. The studies which use SASSIFI for fault injection investigate the effect of transient faults with SASSIFI's bit-flip model into the ISA (Instruction Set Architecture) visible states, including general-purpose registers, memory values' predicate registers, and condition registers in single or multiple threads.

\textbf{\textit{Evaluation}:} Reliability evaluation of DNNs in GPUs is carried out more extensively than in other platforms. Nearly all works have classified injected faults \cite{dos2018analyzing}\cite{ibrahim2020softerror}\cite{adam2021selective}\cite{adam2021analyzing}\cite{adam2021impact}\cite{condia2021combining}\cite{dos2022characterizing}\cite{garrett2021improving}\cite{ibrahim2020analyzing}\cite{ibrahim2020analyzinginstruction} \cite{dos2017evaluation}\cite{dos2019impact}\cite{lotfi2019resiliency}\cite{condia2022multi}\cite{guerrero2022effective}\cite{guerrero2022evaluating}. The general model for classifying faults in the mentioned works is as follows:

\begin{itemize}
    \item \textbf{Masked:} Fault does not affect the output,
    \item \textbf{SDC:} Output confidence score differs from that of the golden model,
    \item \textbf{DUE:} The program hangs or the system reboots (also called \textit{Crash} in \cite{dos2018analyzing}\cite{dos2017evaluation})
\end{itemize}

Furthermore, SDC is also categorized regarding the effect of faults on the accuracy of the DNN for the object recognition task in \cite{adam2021selective}\cite{ibrahim2020softerror}. They define three categories of SDCs based on the effect of faults on the output confidence score and ranking of objects:

\begin{itemize}
    \item \textbf{Non-critical:} Output confidence score changed, and no misclassification occurred and no objects ranking modified,
    \item \textbf{Light-critical:} Objects ranking modified, and no misclassification occurred,
    \item \textbf{Critical:} Impacted the output confidence score and caused misclassification.
\end{itemize}

On the other hand, the fault classification of SDCs proposed in \cite{dos2019impact} is beyond the classic SDCs and is based on the impact of faults on the precision and recall for object detection tasks in a self-driving car, as follows:

\begin{itemize}
    \item \textbf{Non-critical}: Precision maintains larger than 90\% (a new object is detected that is not in the original classification) and recall remains 100\% (all previous objects are detected).
    \item \textbf{Critical:} Precision is lower than 90\% (many wrong objects detected) and recall is not 100\% (real objects are not detected).
\end{itemize}

\begin{figure}[h]
    \includegraphics[width=0.5\textwidth]{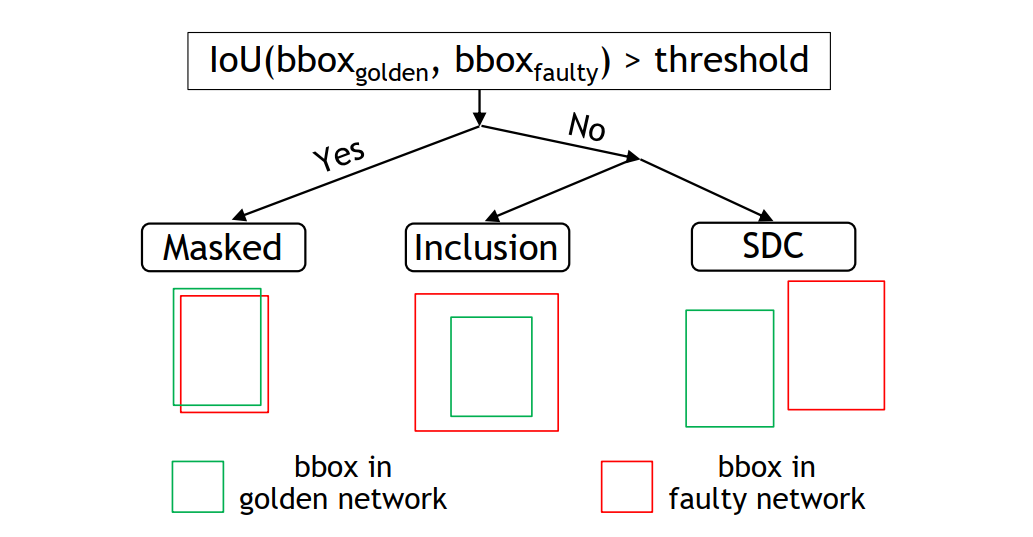}
    \centering
    \caption {Fault classification in the object detection task based on bounding boxes \cite{lotfi2019resiliency}}
    \label{fig:bounding}
\end{figure}

Furthermore, new classes of faults are presented in \cite{lotfi2019resiliency} which considers the margins of the bounding box in the DNN for object detection. Authors compare the overlaps of the bounding box of the detected objects in each image for golden and faulty models and categorize the SDCs based on a threshold. Their fault classification method is depicted in Fig.~\ref{fig:bounding}. 

Vulnerability factors are also adopted to analyze the reliability of DNNs on GPU platform \cite{dos2018analyzing}\cite{ibrahim2020softerror}\cite{adam2021selective}\cite{adam2021analyzing}\cite{condia2021combining}\cite{ibrahim2020analyzing} \cite{ibrahim2020analyzinginstruction}\cite{dos2017evaluation}\cite{dos2019impact}. Vulnerability factors express the probability of propagating faults from a particular component to the output. Since faults may be injected to different locations, so that vulnerability factor of the location (in different abstraction levels from architecture to program) can be measured. In this regard, Kernel Vulnerability Factor (KVF) \cite{adam2021selective}\cite{ibrahim2020analyzing}, Layer Vulnerability Factor (LVF) \cite{adam2021selective}\cite{bolchini2022selective}\cite{ibrahim2020analyzing}, Instruction Vulnerability Factor (IVF) \cite{adam2021selective}\cite{adam2021analyzing}\cite{ibrahim2020analyzinginstruction}, Program Vulnerability Factor (PVF) \cite{dos2018analyzing}\cite{ibrahim2020softerror}\cite{adam2021selective}\cite{dos2017evaluation}, Operation Vulnerability Factor \cite{garrett2021improving}, and Architecture Vulnerability Factor (AVF) \cite{dos2018analyzing}\cite{ibrahim2020softerror}\cite{adam2021selective}\cite{cavagnero2022transient}\cite{condia2021combining}\cite{dos2017evaluation}\cite{dos2019impact} have been presented. These metrics provide a thorough understanding of the vulnerability of each location either in DNN or in GPU.

\paragraph{\textbf{5.1.2.3 \hspace{1mm} Processors Platform}}

DNNs exploit processors mostly for IoT and edge applications. The research works in which faults are emulated on multi-core processors running DNNs are reviewed in this subsection.
Soft errors in the register file of ARM processors running DNNs have been studied extensively in \cite{abich2020soft}\cite{bandeira2019non}\cite{abich2021impactprecision}\cite{abich2021applying}\cite{abich2022impactthread}\cite{abich2022impact}\cite{gava2022soft}\cite{liu2022using}\cite{liu2022efficient} \cite{liu2021analyzing}. The vulnerability of instructions is studied in \cite{liu2021analyzing}. To emulate faults modeling soft errors in target processors, ARM-FI is developed and adopted in \cite{liu2022using}\cite{liu2022efficient}\cite{liu2021analyzing} and SOFIA \cite{bandeira2019non} is exploited in \cite{abich2020soft}\cite{bandeira2019non}\cite{abich2021impactprecision}\cite{abich2021applying}\cite{abich2022impactthread}\cite{abich2022impact}\cite{gava2022soft} as fault injection frameworks. Each of the aforementioned fault injectors enables fault emulation in different components of processors.

\textbf{\textit{Evaluation}:} All works in this class have evaluated the reliability by fault classification. The classification is performed similarly to the general scheme of classifying faults in the previous platforms (Masked, tolerable SDC, critical SDC, and DUE). 

Furthermore, references \cite{abich2020soft}\cite{bandeira2019non} classify the faults in an object detection task for autonomous vehicles as:

\begin{itemize}
    \item \textbf{Incorrect probability:} All objects detected correctly with different output confidence score,
    \item \textbf{Wrong detection:} Misclassification or missing an object,
    \item \textbf{No prediction:} No object detection.
\end{itemize}

Mean Work To Failure (MWTF) is also exploited as a reliability metric to show the amount of work a neural network can perform until meeting a failure, as:

\begin{equation}
    MWTF = \frac{1}{execution time \times AVF_{critical-faults}}
\end{equation}
where $AVF_{critical-faults}$ is the probability of an erroneous classification due to faults. MWTF is adopted as a relationship between performance and reliability in \cite{liu2022efficient}\cite{liu2021analyzing}. 
AVF is obtained as the reliability metric for the register file in \cite{abich2021applying}\cite{liu2022efficient}\cite{liu2021analyzing}.
Program Vulnerability Factor (PVF) is leveraged to express the vulnerability of operations and instructions in \cite{liu2021analyzing}.

\subsubsection{\large{Irradiation}} \label{char-fi-rad}

The most realistic way of fault injection is to irradiate the devices under the beam of particles, e.g., neutron or ion. In this subsection, the research works which study the reliability of DNN accelerators i.e., FPGA and GPU under radiation, are described.

\paragraph{\textbf{5.1.3.1 \hspace{1mm} FPGA Platform}}

\begin{figure}[ht]
    \includegraphics[width=0.5\textwidth]{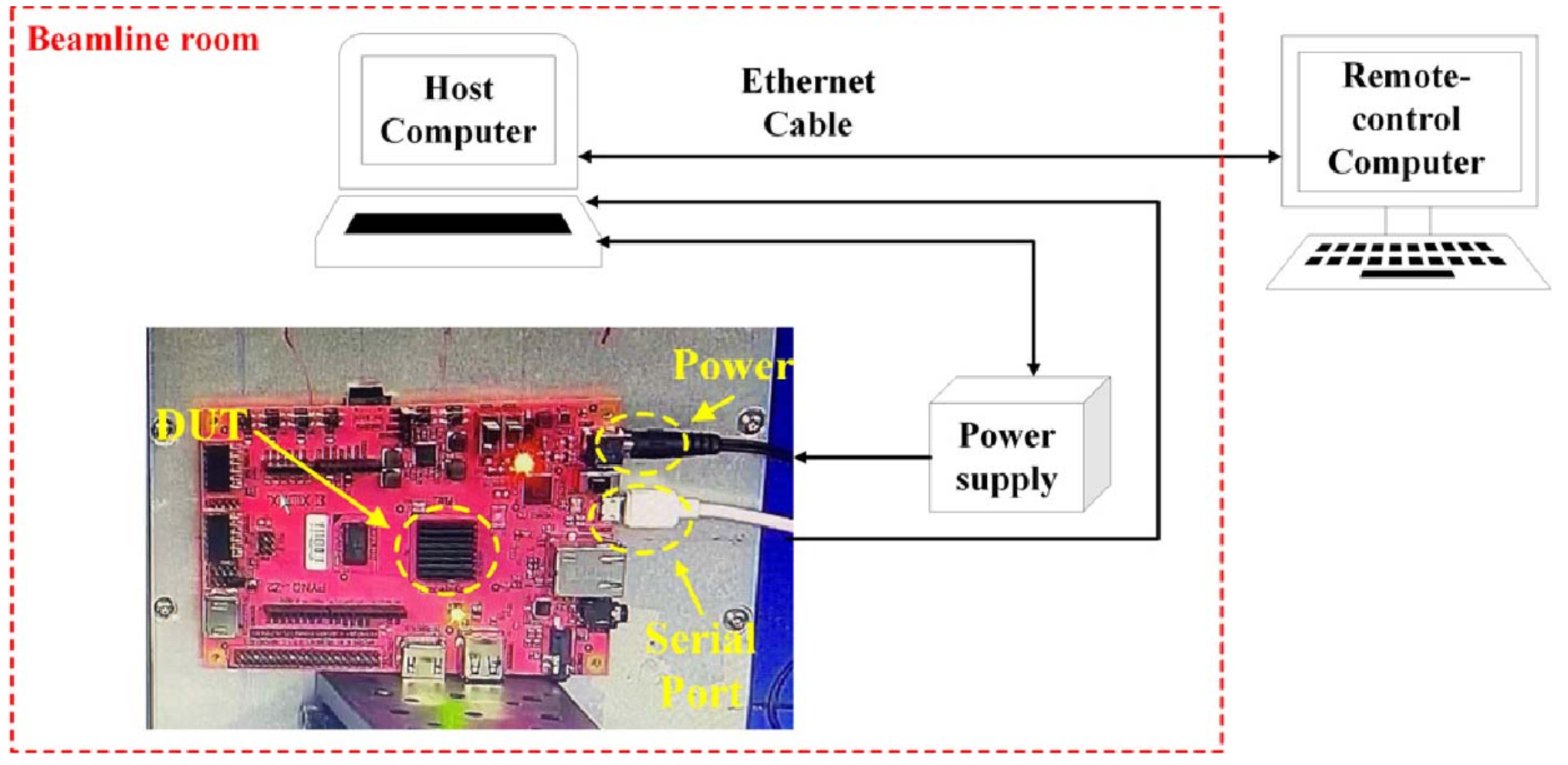}
    \centering
    \caption {Block diagram of the setup of beam experiment in \cite{wang2021impact}}
    \label{fig:FPGA_rad}
\end{figure}

Zynq SoCs have been examined under radiation tests to assess the reliability of DNNs in \cite{benevenuti2018comparative}\cite{benevenuti2021neutron}\cite{libano2018selective}\cite{matanaluza2021emulating}\cite{wang2021impact}\cite{libano2021reduced}\cite{luza2020investigating}. FPGAs are irradiated with neutrons in \cite{benevenuti2018comparative}\cite{benevenuti2021neutron}\cite{libano2018selective}\cite{wang2021impact}\cite{agiakatsikas2021evaluation}\cite{gambardella2022accelerated}\cite{libano2021reduced} and with protons in \cite{maillard2022radiation}. References \cite{gambardella2022accelerated} and \cite{maillard2022radiation} have applied fault-aware training to DNNs and studied its impact under radiation. HyperRAM which includes constant and dynamic variables (e.g., weights and biases) is bombarded with ionizing particles in \cite{matanaluza2021emulating}\cite{luza2020investigating}. The research works set up the configuration of the system before the experiment mostly based on HW/SW co-design and save the results for further analysis. Fig.~\ref{fig:FPGA_rad} shows an example of the setup of the FPGA irradiation.

\textbf{\textit{Evaluation}:} Radiation experiments enable reliability evaluation by SER or FIT metrics \cite{libano2018selective}\cite{matanaluza2021emulating} \cite{wang2021impact}\cite{luza2020investigating}. To formulate the SER, cross-section is defined as the proportion of observed faults ($errors$) over all particles collided to the surface ($Flux$), as expressed in \eqref{eq:cross_section} \cite{wang2021impact}. Cross-section \(\sigma\) is expressed as a unit of \(cm^2\) and is the probability that a particle may cause an observable error \cite{libano2018selective}. The cross-section is exclusively adopted in \cite{agiakatsikas2021evaluation}\cite{gambardella2022accelerated}.

\begin{equation}
\sigma = errors / Flux \label{eq:cross_section}
\end{equation}

The cross-section can lead to SER or FIT calculation by getting multiplied by the particle flux that the device will experience in the environment ($\phi$). SER represents the number of failures of the device in \(10^9\) hours as shown in \eqref{eq:SER}.
\begin{equation}
SER = \sigma \times \phi \label{eq:SER}
\end{equation}

Most research works that study irradiation on FPGAs evaluate the reliability of devices under test by the above metrics. In addition, some works classify the faults radiated into FPGA by observing the outputs \cite{libano2018selective}\cite{libano2021reduced}\cite{maillard2022radiation}. Here, both works provide fault classification based on output confidence scores of the neural network. \cite{libano2018selective} sets up a HW/SW co-design implementation on a target board and identifies the faults causing no misclassification (tolerable) and misclassification (critical). Thereafter,   the FIT of different classes of faults is obtained. \cite{libano2021reduced}\cite{maillard2022radiation} also present the cross-sections of the device for different classes of faults (including tolerable errors, critical errors, and crashes). Moreover, the reliability is estimated by the aforementioned metrics in \cite{benevenuti2018comparative} as expressed in \eqref{eq:reliability_FPGA}.

\paragraph{\textbf{5.1.3.2 \hspace{1mm} GPU Platform}}

\begin{figure}[ht]
    \includegraphics[width=0.4\textwidth]{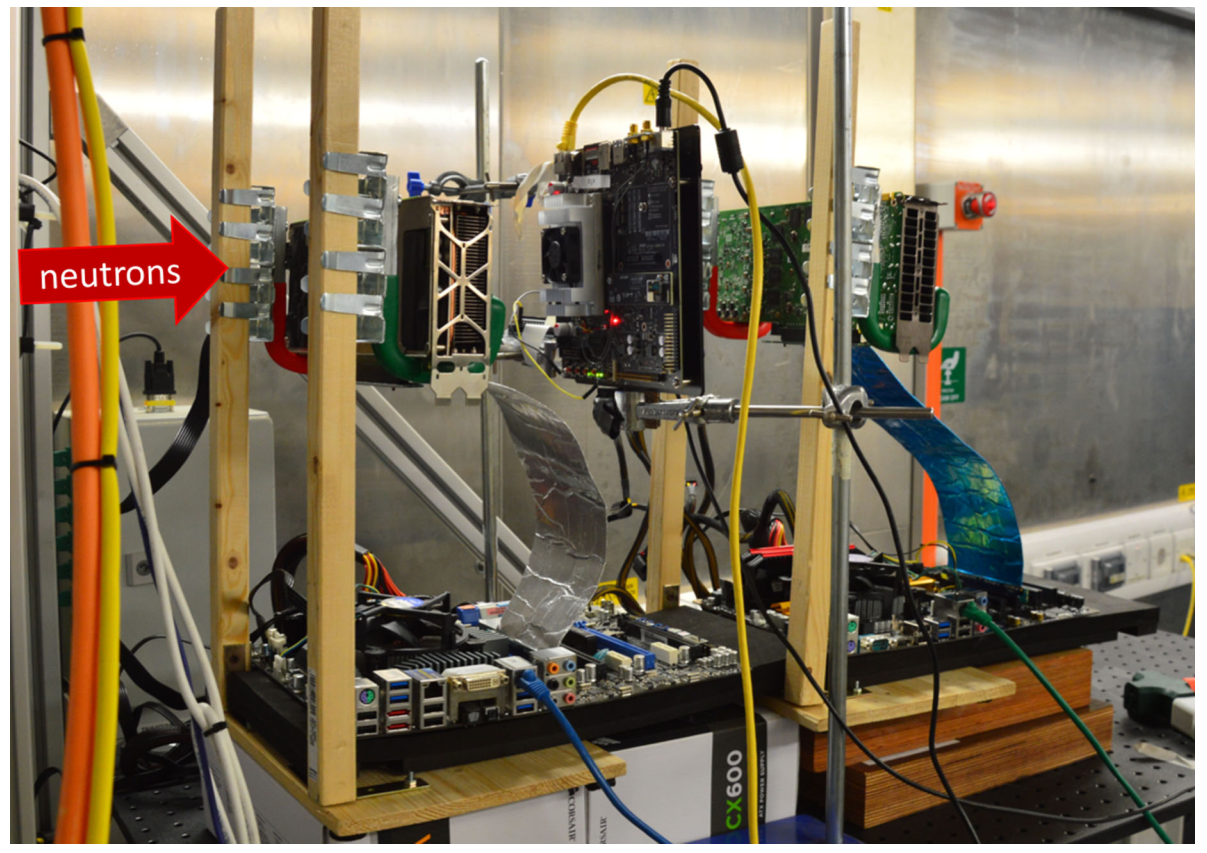}
    \centering
    \caption {Setup of neutron irradiation to GPU \cite{dos2018analyzing}\cite{dos2017evaluation}\cite{basso2020impact}}
    \label{fig:GPU_rad}
\end{figure}

Reliability of DNNs on GPUs are assessed under neutron beam radiation in \cite{dos2018analyzing}\cite{dos2022characterizing}\cite{hari2021making} \cite{dos2017evaluation}\cite{dos2019impact}\cite{basso2020impact}\cite{lotfi2019resiliency}. All GPUs under test are manufactured by NVIDIA and have different architectures. They also provide tests by enabling and disabling ECC configurations, and different data representations. Each work has specified flux of neutrons and radiation time, e.g., \cite{lotfi2019resiliency} tests the GPU equivalent to 2,000 years of exposure to terrestrial neutron, or \cite{dos2018analyzing} reports data that cover more than 110,000 years of GPU operation. Fig.~\ref{fig:GPU_rad} illustrates the radiation test setup in \cite{dos2018analyzing}\cite{dos2017evaluation}\cite{basso2020impact}.

\textbf{\textit{Evaluation}:} Research works of this group present reliability evaluation of DNNs on GPUs by FIT as well as fault classification similar to the works on FPGAs radiation. Authors in \cite{dos2018analyzing}\cite{dos2017evaluation} identify faults that caused SDC and Crash and report their FIT, separately. \cite{dos2022characterizing} and \cite{dos2019impact} report FIT of faults caused SDC and DUE separately in different data representations of the DNN, and in \cite{lotfi2019resiliency} irradiated faults are classified based on Fig.~\ref{fig:bounding}. SDC rate is also the adopted evaluation metric in \cite{hari2021making}.

\paragraph{\textbf{5.1.3.3 \hspace{1mm} TPU Platform}}

The reliability of Google's Tensor Processing Unit (TPU) is studied under neutron beam radiation in \cite{rech2022reliability} and \cite{junior2022high}. These works experimented Coral TPU chip, a low-power accelerator for DNNs, with several neural networks for image classification and object detection tasks. 

\textbf{\textit{Evaluation}}: The research works performing radiation experiments on Coral TPU have evaluated the reliability by FIT and cross section as well as by fault classification. In this regard, SDC and DUE fault effects are reported based on FIT and cross section.

\subsection{\large{Analytical Methods}}

Analytical methods in reliability assessment model the reliability mathematically and do not inject faults into the platform to be simulated to evaluate the reliability. These methods rely on the function and algorithm of DNNs, and if needed, also consider the structure of the accelerator. Nevertheless, they carry out fault injection to assess the efficacy of the methods. For the sake of generalization, all works in this group analyze the relations of neurons and layers to find their effect and contribution to the output. In this regard, they estimate the vulnerability of neurons and analyze how a faulty neuron may impact the output to find critical neurons. Therefore, they link the reliability of the network with the vulnerability of its neurons and provide an analytical model of calculating the reliability for DNNs.

We have identified four approaches in analytical methods: 
\begin{itemize}
    \item Layerwise Relevance Propagation (LRP) based analysis \cite{abdullah2020salvagednn}\cite{ruospo2021reliability}\cite{ruospo2022selective}\cite{schorn2018accurate}\cite{schorn2019efficient}, 
    \item Gradient-based analysis \cite{choi2019sensitivity}\cite{mahmoud2020hardnn}\cite{mahmoud2021optimizing}\cite{sabih2021fault}, 
    \item Estimation-based analysis \cite{mahmoud2020hardnn}\cite{mahmoud2021optimizing}\cite{ping2020sern}, 
    \item ML-base analysis \cite{gavarini2022open}. 
\end{itemize}

In the first approach, DNNs are analyzed based on an algorithm called Layerwise Relevance Propagation (LRP) that leads to obtaining critical scores for neurons/fmaps. The second approach is based on the gradients of weights/fmaps with respect to the output leading to their sensitivity. Research works in the third approach estimate the vulnerability of DNNs by finding correlations between some information from DNNs and the vulnerability of layers/fmaps. In the last approach, ML-based techniques are adopted in the context of fault analysis in DNNs.

In the LRP-based analysis, a hypothesis is raised in \cite{schorn2018accurate} proposing that the higher the contribution of neurons to the DNN's output, the more impact they have on the classification accuracy. Accuracy loss is one of the most important metrics in the reliability evaluation. Therefore, the more impact a neuron has on the accuracy, the more vulnerable it is which means it has more influence on the reliability of the network, consequently. Hence, the authors adopted the Layerwise Relevance Propagation (LRP) algorithm to obtain the value of the contribution of each neuron to the output. LRP indicates 
the proportion of each connected neuron in constructing the value of the target neuron and calculates this ratio for all neurons from the last layers to the first. LRP specifies \(R_{i,j}(y_{0}, t)\) for each neuron \textit{j} in layer \textit{i} which is its output contribution score between 0 and 1 with the input \(y_0\) and output class \textit{t}. Then, the average score of each neuron over the entire training set of M inputs is obtained representing the resilience of the corresponding neuron as \eqref{eq:LRP}. 
\begin{equation}
r_{i, j} = \frac{M}{\sum_{m=0}^{M-1} R_{i, j}(y_{0, m}, t_m)}
\label{eq:LRP}
\end{equation}

Thereafter, the sorted list of neurons regarding their \(r_{i, j}\) represents the most to least vulnerable neurons that can lead to protecting the most vulnerable neurons to improve reliability. Furthermore, by this analytical method, another reliability improvement method is presented in \cite{schorn2019efficient} based on balancing the resilience distribution inside the DNN. Similarly, \cite{abdullah2020salvagednn} proposes an approach to extract the saliency or importance of each neuron and proposes a mapping scheme for neurons on PEs of a systolic array to minimize the score of corrupted weights. 

Authors in \cite{ruospo2021reliability} extend the LRP algorithm based on different output classes of input images and provide the list of neurons' resilience scores (score maps) for individual classes separately, as well as the score map of the whole network regardless of the output classes. Then, all sorted score maps are combined in descending order to set the maximum score to each corresponding neuron. Subsequently, a scheduling algorithm is applied to map neurons to PEs of an MPSoC based on the score maps.

In gradient-based analysis, three papers are identified. Explainable AI that explains how the network computes the output by the input is exploited in \cite{sabih2021fault} to obtain the sensitivity of layers and importance of weights. This work defines the sensitivity of layers in compliance to the difference of the two highest output confidence scores of the last layer. Therefore, they obtain the average sensitivity of all layers and relate it to the importance of weights. They provide the most important weights and their critical bits consequently to be protected.

Sensitivity of filters and weights are analyzed in \cite{choi2019sensitivity} that refers to the amount of accuracy drop with bit-flip occurrence in weights. In the proposed method in this paper, the gradient of weights with respect to the output is calculated over a dataset considering a cost function. Also, the expectation for the probability of weights to be faulty is obtained as a noise measurement ($\varepsilon_w$). The sensitivity of a weight $w$ is measured as \eqref{eq:weight-sensitivity}.
\begin{equation}
Sensitivity_w = gradient_w \times \varepsilon_w
\label{eq:weight-sensitivity}
\end{equation}
Sensitivity analysis in this work leads to allocation of robust hardware to the more sensitive weights.

\cite{mahmoud2020hardnn}\cite{mahmoud2021optimizing} have presented three gradient-based approaches for vulnerability estimation of fmaps in a DNN. \textit{Gradient} approach considers the absolute values of fmaps' gradients with respect to the cross-entropy loss at the output in a backpropagation as the vulnerability of fmaps. \textit{Gain} approach measures the noise gain by obtaining the expectation for a set of corrupted neurons affecting the DNN's accuracy, based on the derivatives of outputs with respect to the neurons over a set of data and the variance of noise source. \textit{Modified Gain} is also proposed based on the \textit{Gain} approach to violate the independence between neurons and noise. The three mentioned approaches evaluate the vulnerability of fmaps in a DNN.

Authors in \cite{mahmoud2020hardnn}\cite{mahmoud2021optimizing} also presented three estimation-based approaches for the vulnerability of fmaps. They estimate the relative fmaps' vulnerability by calculating the \textit{max neuron value}, \textit{fmap range}, and \textit{average L2} over the input samples. They have provided approximate yet scalable and fast approaches to estimate the vulnerability of fmaps. 

\cite{ping2020sern} presents an equation to estimate the misclassification rate of CNNs in case of soft error occurrence in a specific layer. The authors consider any operation resulting in a non-zero value as a 
critical computation, since soft errors may corrupt their results. The estimation is based on the proportion of critical operations (\textit{Crit\_OPs}) in the target layer \textit{i} and subsequent layers relative to all operations in those layers, to model the misclassification rate (\textit{SERN}) in a CNN with \textit{n} layers. Equation \eqref{eq:sern} provides a representation of this estimation.
\begin{equation}
SERN = \frac{Crit\_OPs_i + \sum_{i+1}^{n}OPs}{\sum_{i}^{n}OPs}
\label{eq:sern}
\end{equation}

An ML-based approach for analytical reliability analysis is presented in \cite{gavarini2022open} where Open-Set Recognition (OSR) methods are explored to analyze the criticality of faults in DNNs' parameters. The concept of OSR is to identify  whether the output classification corresponds to the trained classes of the DNN. This concept is adapted to analyze the output logits (output of \textit{softmax} in the last layer) of DNNs to identify the critical fault in the parameters. Four different OSR-based methods have been leveraged for this task and their efficacies are reported. In each method, a threshold for the output logits is obtained for identifying critical fault occurrence.

All the works in this group evaluate their analytical methods on the reliability by FI. The FI methods that are used in these works are similar to the FI methods presented and characterized in section \ref{char-FI}. It is shown that analytical methods can evaluate/estimate the vulnerability/sensitivity of different components of DNNs including neurons, fmaps, and weights. Analytical methods are more lightweight than FI by far and are accelerator-agnostic. However, their analysis results can be utilized for designing robust DNN accelerators. Among the existing approaches, estimation-based analyses are faster than others while less accurate when the results are compared with FI experiments. LRP-based and gradient analyses provide more accurate results close to FI experiments yet they are faster and incurring less complexity.

\subsection{\large{Hybrid Methods}}

In hybrid methods, both FI and analytical methods are carried out to assess the reliability of DNNs. To that end,  \cite{he2020fidelity} proposes a reliability assessment framework called Fidelity based on a hybrid method. This framework studies the transient faults in both, data and control path of accelerators. Fidelity contains fault injection in software framework TensorFlow to obtain the probability of masking faults in the DNN. In addition, the framework is capable of analyzing the architectural model of the accelerator, and map Flip Flops (FFs) of datapath and control logic to the parameters of a high-level implementation of the DNN. By the fault injection and elaborate analysis, it models the probability of activeness/inactiveness of FFs during the execution time as well as the probability of masking faults. Subsequently, the framework provides the \textit{FIT rate} of the accelerator. Furthermore, the framework is validated by analyzing the NVDLA \cite{NVDLA}, i.e., an open-source NVIDIA's DNN accelerator. To further improve this method, a software model for NVDLA is proposed in \cite{veronesi2022exploring} to enable reliability study of accelerators at the software level and provide a more accurate, more hardware-aware, and faster method to obtain \textit{FIT rate} of the accelerator.

Zhang et al. \cite{zhang2022estimating} propose a hybrid of ML-based analysis and FI to estimate the vulnerability of all parameters in DNNs by a low number of fault injections. The proposed method involves selecting a set of random parameters of the DNN and evaluating their vulnerabilities by injecting bitflip faults and measuring the accuracy loss. Thereafter, some features for the selected parameters (absolute value, gradient, calculation times, and layer location) are extracted. A random forest as a machine learning approach is trained and tested using the features and vulnerability of the corresponding parameters so that when it reaches a high accuracy, it can be used for vulnerability estimation of the entire set of parameters. 

\begin{table*}[h!]
\caption {Pros and cons of reliability assessment methods for DNNs.}
\small
\begin{adjustbox}{width=\textwidth}
\begin{tabular}{|p{1.55cm}||p{6.25cm}|p{6.15cm} | } 

\hline
\textbf{Method}  & \textbf{Pros} & \textbf{Cons}\\ 
\hhline{|=||=|=|}

\textbf{Fault Simulation} & 
\begin{tabular}[c]{@{}l@{}}
- Low \hfill design \hfill time \hfill and \hfill fast \hfill execution \hfill in \\ high-level software implementations \\ 
- Adoptable \hfill for \hfill various \hfill DNNs, \hfill DHA \hfill models, \\ and fault models\\
- Enabling \hfill reliability \hfill study \hfill of \hfill variations \hfill of \hfill DNNs \\  under \hfill approximation, \hfill quantization, \hfill encryption \\ etc.\\ 
- The \hfill availability \hfill of \hfill open-source \hfill frameworks \\ for high-level software simulation \\
- No \hfill need \hfill for \hfill special \hfill facilities \hfill and \hfill capable \hfill of \\ being run on regular PCs\\ 
- Enabling \hfill a \hfill fast \hfill evaluation \hfill of \hfill reliability \\ enhancement \hfill methods \hfill at \hfill high-level \hfill software \\ implementations \\
- Providing various reliability evaluation metrics \\[1ex]
\end{tabular} & 

\begin{tabular}[c]{@{}l@{}}
- High \hfill time \hfill complexity \hfill to \hfill achieve \hfill a \hfill sufficient\\ confidence level \\
- Not \hfill realistic \hfill model \hfill of \hfill fault \hfill effects \hfill in \\ high-level software implementations \\ 
- Inaccurate \hfill results \hfill at \hfill high-level \hfill software \\ implementations \\ 
- Time-consuming \hfill design \hfill and \hfill development for \\ HDL implementations
\end{tabular}\\  
\hline

\textbf{Fault Emulation}  & 
\begin{tabular}[c]{@{}l@{}}
- Providing realistic reliability analysis of DHA\\ 
- Enabling \hfill experiments \hfill for \hfill real \hfill conditions \hfill of \\DHA operation\\ 
- Providing \hfill full \hfill access \hfill to \hfill possible \hfill locations \hfill of \\the DHA for FI\\ 
- Enabling realistic studying of faults in datapath\\ 
- Providing fault-tolerant designs and evaluating \\them directly\\ 
- Providing \hfill several \hfill evaluation \hfill metrics \hfill and \hfill fault \\classifications \\[1ex]
\end{tabular} & 

\begin{tabular}[c]{@{}l@{}}
- Time consuming design and development\\ 
- Need for the physical DHA \\
- Different \hfill platforms \hfill need \hfill their \hfill own \hfill specific \\ design and development to perform FI\\ 
- Need for platform-specific frameworks for FI \\
\end{tabular} \\ 
\hline

\textbf{Irradiation} & 
\begin{tabular}[c]{@{}l@{}}
- Performing \hfill realistic \hfill experiments  \hfill as \hfill real \\ physical faults are injected into the chip\\ 
- Suitable for developing fault models\\ 
- Enabling \hfill the \hfill study \hfill for \hfill validating \hfill simulation \\ and emulation approaches\\ 
- Providing the real behavior of the DHA when \\ faced with a physical effect \\[1ex]
\end{tabular} & 
\begin{tabular}[c]{@{}l@{}}
- Need \hfill for \hfill specific \hfill facilities \hfill for \hfill performing \\radiation \\  
- Low control over accuracy of fault injection  \\ in terms of number and locations of occurred \\ faults \\ 
- Lack of the visibility of fault propagation
\end{tabular}\\ 
\hhline{|=||=|=|}

\textbf{Analytical} & 
\begin{tabular}[c]{@{}l@{}}
- Implementable at software-level \\ 
- Scalable and less complex than FI \\
- Leading to fault tolerant hardware designs \\
- Providing \hfill information \hfill for \hfill algorithm-level \\resiliency for DNNs \\
- DHA-agnostic \\[1ex]
\end{tabular} & 

\begin{tabular}[c]{@{}l@{}}
- Not providing quantitative evaluation metrics \\
- Not considering DHA models\\
- Inaccurate in estimating the vulnerabilities of \\ DNN components (neurons, fmaps, etc.) \\
\end{tabular}\\ 
\hhline{|=||=|=|}

\textbf{Hybrid} & 
\begin{tabular}[c]{@{}l@{}}
- Combining fast FI with an analytical approach \\
- Capability of reliability study for DHAs\\
- Possibility of evaluation by either vulnerability \\ estimation or quantitative metrics \\ [1ex]
\end{tabular} & 

\begin{tabular}[c]{@{}l@{}}
- Need \hfill for \hfill detailed \hfill information \hfill of \hfill the \hfill DHA \\ (depending on the method) \\
- Accuracy \hfill of \hfill the \hfill results \hfill could \hfill be \hfill low \\ (depending on the method) \\
\end{tabular}\\ 
\hline

\end{tabular}
\end{adjustbox}
\centering
\label{tab:FI_pros_cons}
\end{table*}

\section{Discussion} \label{discussion}

In this section, we will first discuss the reliability assessment methods for DNNs based on the works reviewed and presented in Section \ref{charaterization}. Then, we will summarize the current status in the three main categories of reliability assessment: FI, analytical, and hybrid methods, respectively and address their pros and cons in the research domain of this literature review. Thereafter, we will present a qualitative comparison of different reliability assessment methods for DNNs. Lastly, we will list the open challenges as well as major potential research directions for the future. 

Table \ref{tab:FI_pros_cons} lists the pros and cons of all the methods categorized in this work and described in Section \ref{charaterization}. 

Of the reviewed papers, FI as a conventional method for reliability assessment, is frequently used for evaluating the DNNs' reliability. FI provides realistic results about how faults impact the system's execution. FI methods can be conducted for modeling various faults which can be injected at the different locations in the platform for reliability evaluation. Moreover, they are applicable to any platform at any system abstraction level and provide various reliability evaluations based on metrics and fault classifications. Therefore, many research works choose FI as their primary method of DNNs' reliability assessment. Nevertheless, FI methods are accompanied by a prohibitively high complexity due to the need to consider several cases for fault occurrence and to iteratively repeat the executions.

Analytical methods have been proposed as a way to cope with the high complexity of FI methods. These methods study the function of DNNs and assess the model's reliability using mathematical equations, leading to less complex approaches. Since analytical methods are developed mathematically, they have the potential to be generalized and adapted to various DNNs. Notably, analytical methods have the potential to be exploited in the reliability assessment of the training phase. However, current analytical methods do not consider the accelerator models, and there is a gap in the use of reliability evaluation metrics. While this survey identifies a relatively small number of works relying on analytical methods for DNNs' reliability assessment, the future of research in this area should pay greater attention to the potential of analytical methods. 

Finally, hybrid methods combine the strength of both, FI and analytical methods. By applying analysis of the network or the accelerator in addition to conducting fault injection, hybrid methods are capable of obtaining a comprehensive and realistic evaluation of reliability. Although a limited number of research works have been identified in this category in the present survey, there is a huge room to explore these methods for DNNs' reliability assessment in the future.
Table~\ref{tab:methods-comparison} presents a qualitative comparison between the categorized methods of reliability assessment for DNNs regarding the papers included to this survey.

The analysis of statistics presented in Fig.~\ref{fig:methods-dist} highlights that the majority of the identified research works employ FI to assess the DNNs' reliability. This can be attributed to the fact that, while DNNs are an emerging topic in computer science, the problem of reliability has been a classic issue for a long time. In addition, the investigation of reliability over DNNs has started gaining traction since 2017, as indicated in Fig.~\ref{fig:published-years}. As a result, it is not surprising that the early research in this area has primarily focused on conventional methods such as FI. This could be the main reason for the significant imbalance in the number of published papers across different method categories. However, in the future, the emergence of analytical and hybrid methods is expected to bridge this gap and increase their application in the field of DNN reliability assessment.

\begin{table*}[h!]
\caption {Qualitative analysis comparing different reliability assessment methods for DNNs.}
\small
\begin{adjustbox}{width=\textwidth}
\begin{tabular}{|l|l|l|l|}
\hline
                      & Fault injection  & Analytical      & Hybrid              \\ \hline
Time Complexity        & High             & Low to Moderate  & Moderate \\ \hline
HDA-aware         & Yes              & No              & Yes                 \\ \hline
Leading to fault-tolerant design & Yes              & Yes             & Yes                 \\ \hline
Fault models variety  & All fault models       & Few fault models      & Few fault models          \\ \hline
Implementation system level          & Software and hardware       & Software  & Software          \\ \hline
Evaluation accuracy   & Moderate to high & Low to moderate & Moderate            \\ \hline
Development time      & Low to Moderate & Moderate        & High                \\ \hline
Evaluation metrics &
  \begin{tabular}[c]{@{}l@{}}Accuracy loss\\ Fault classification\\ Vulnerability factors\\ SDC rate\\ Reliability equations\end{tabular} &
  \begin{tabular}[c]{@{}l@{}}Criticality scores\\ Sensitivity\\ Vulnerability estimation\end{tabular} &
  \begin{tabular}[c]{@{}l@{}}FIT Rate\\ Vulnerability estimation\end{tabular} \\ \hline
\end{tabular}%
\end{adjustbox}

\centering
\label{tab:methods-comparison}
\end{table*}

To address open challenges in reliability assessment methods for DNNs, this survey has identified the following main observations:

\begin{itemize}
    \item Although some research works, such as \cite{chan2022fault}, have studied the impact of faulty data during training, no work on the reliability assessment of the training phase has been identified that considers faulty parameters or computational units. This issue should be studied in future research;
    \item Nearly all included works focus on CNNs, with image classification and object detection tasks excluding other types of DNNs, such as RNNs and LSTMs as well as different applications that should also be evaluated in terms of reliability;
    \item The survey has identified no software FI framework in hardware-aware platforms. Hence, DNN accelerator simulators could be exploited or developed for reliability assessment of DNNs in this platform;
    \item Fault emulation on FPGAs can take advantage of HLS designs. Therefore, a general FI framework for these platforms could be presented using HLS to minimize design time;
    \item Based on this survey, very few works study the reliability of the control part of DHAs, especially in FPGAs and ASICs. The control part may play a significant role in the reliability of DNN accelerators and this should be explored in future studies;
    \item There is a limited number of analytical methods for DNNs reliability assessment in this survey, all of which rely on finding critical neurons for fault-tolerant designs. Also, only one work tries to predict the accuracy loss caused by soft errors, and ML-based approaches are proposed in one work. Nevertheless, none of them can estimate the reliability of DNNs on their own or evaluate the reliability using specific metrics. ML-based algorithms can significantly assist in efficient reliability assessment, and therefore, there is a huge potential for developing new analytical methods of reliability assessment for DNNs;
    \item Analytical methods could be generalized for other DNNs and applications rather than considering only CNNs and image processing;
    \item Hybrid methods appear to be powerful and capable of being exploited for developing reliability assessment frameworks. They can be one of the major methods for reliability assessment of DNNs in future works;
    \item Several FI research works carry out accuracy loss and fault classification as an evaluation of reliability. Also, some works considered FIT. However, there is still an urgent need to present DNN-specific metrics for reliability evaluation.
\end{itemize}

As an outcome of this survey, in addition to the listed open challenges, the major possible research directions for future studies in this domain are addressed below:

\begin{itemize}
    \item Although analytical and hybrid methods have potential in the literature, they are not evolved to the extent that their effectiveness can be fully realized. Existing methods have shown that analytical and hybrid methods are capable of assessing the DNNs' reliability as realistically as FI, and lead to effective fault-tolerant designs. Moreover, ML-based approaches in conjunction with analytical and hybrid methods are emerging. Therefore, researchers can be directed to develop novel analytical and hybrid methods, especially those that adopt ML-based algorithms, for reliability assessment of DNNs that are faster, less complex, more scalable, and more specific to DNNs than the conventional FI approaches.
    \item Bringing reliability as a classical issue into an emerging topic such as DNNs requires new tools to respond to the requirements of the new domain. Therefore, the new research not only needs to adopt commonly used metrics in the reliability domain, but also requires the introduction and proposal of novel DNNs-specific reliability evaluation metrics.
    \item There are several IoT and edge applications for DNNs emerging day by day, and reliability is not only a concern for safety-critical applications. New research can focus on the unstudied applications of DNNs while taking reliability into consideration.
\end{itemize}

\section{Conclusion} \label{conclusion}

DNNs are being utilized in an increasingly diverse range of applications in our daily life. Consequently, their deployment in safety-critical applications has emerged to be expanding incessantly. However, threats to reliability are one of the major issues that they experience in the real world. To address this, several studies have been published in recent years to assess the reliability of DNNs, with or without the use of accelerators, resulting in the development of various assessment methods. In this work, we conduct a systematic literature review to present a categorization of the reliability assessment methods for DNNs.

Out of the 139 papers related to the subject of the review, three major approaches to reliability assessment of DNNs were identified, i.e., Fault Injection, Analytical, and Hybrid methods. Since the majority of works assess the reliability using conventional fault injection methods, the related works relying on FI methods are characterized based on different approaches and platforms. In addition, we have addressed the advantages and disadvantages of the different methods and highlighted the open challenges that may become the focus of future studies in this domain. Based on the analysis of this survey, future research could focus on developing lightweight, DNN-specific analytical and hybrid methods for assessing reliability, as well as providing new quantitative evaluation metrics that take into account emerging applications for DNNs.

\section*{}

\printbibliography

@inproceedings{bosio2021emerging,
  title={Emerging Computing Devices: Challenges and Opportunities for Test and Reliability},
  author={Bosio, Alberto and O’Connor, Ian and Traiola, Marcello and Echavarria, Jorge and Teich, J{\"u}rgen and Hanif, Muhammad Abdullah and Shafique, Muhammad and Hamdioui, Said and Deveautour, Bastien and Girard, Patrick and others},
  booktitle={2021 IEEE European Test Symposium (ETS)},
  pages={1--10},
  year={2021},
  organization={IEEE}
}

@inproceedings{forsberg2020challenges,
  title={Challenges in Using Neural Networks in Safety-Critical Applications},
  author={Forsberg, H{\aa}kan and Lind{\'e}n, Joakim and Hjorth, Johan and M{\aa}nefjord, Torbj{\"o}rn and Daneshtalab, Masoud},
  booktitle={2020 AIAA/IEEE 39th Digital Avionics Systems Conference (DASC)},
  pages={1--7},
  year={2020},
  organization={IEEE}
}

@inproceedings{nardi2017functional,
  title={Functional safety methodologies for automotive applications},
  author={Nardi, Alessandra and Armato, Antonino},
  booktitle={2017 IEEE/ACM International Conference on Computer-Aided Design (ICCAD)},
  pages={970--975},
  year={2017},
  organization={IEEE}
}

@article{torres2017fault,
  title={Fault and error tolerance in neural networks: A review},
  author={Torres-Huitzil, Cesar and Girau, Bernard},
  journal={IEEE Access},
  volume={5},
  pages={17322--17341},
  year={2017},
  publisher={IEEE}
}

@article{mittal2020survey,
  title={A survey on modeling and improving reliability of DNN algorithms and accelerators},
  author={Mittal, Sparsh},
  journal={Journal of Systems Architecture},
  volume={104},
  pages={101689},
  year={2020},
  publisher={Elsevier}
}

@article{shafique2020robust,
  title={Robust machine learning systems: Challenges, current trends, perspectives, and the road ahead},
  author={Shafique, Muhammad and Naseer, Mahum and Theocharides, Theocharis and Kyrkou, Christos and Mutlu, Onur and Orosa, Lois and Choi, Jungwook},
  journal={IEEE Design \& Test},
  volume={37},
  number={2},
  pages={30--57},
  year={2020},
  publisher={IEEE}
}

@article{ibrahim2020soft,
  title={Soft errors in DNN accelerators: A comprehensive review},
  author={Ibrahim, Younis and Wang, Haibin and Liu, Junyang and Wei, Jinghe and Chen, Li and Rech, Paolo and Adam, Khalid and Guo, Gang},
  journal={Microelectronics Reliability},
  volume={115},
  pages={113969},
  year={2020},
  publisher={Elsevier}
}

@article{talib2021systematic,
  title={A systematic literature review on hardware implementation of artificial intelligence algorithms},
  author={Talib, Manar Abu and Majzoub, Sohaib and Nasir, Qassim and Jamal, Dina},
  journal={The Journal of Supercomputing},
  volume={77},
  pages={1897--1938},
  year={2021},
  publisher={Springer}
}

@article{ozen2020low,
  title={Low-Cost Error Detection in Deep Neural Network Accelerators with Linear Algorithmic Checksums},
  author={Ozen, Elbruz and Orailoglu, Alex},
  journal={Journal of Electronic Testing},
  volume={36},
  number={6},
  pages={703--718},
  year={2020},
  publisher={Springer}
}

@inproceedings{burel2021mozart,
  title={MOZART: Masking Outputs with Zeros for Architectural Robustness and Testing of DNN Accelerators},
  author={Burel, St{\'e}phane and Evans, Adrian and Anghel, Lorena},
  booktitle={2021 IEEE 27th International Symposium on On-Line Testing and Robust System Design (IOLTS)},
  pages={1--6},
  year={2021},
  organization={IEEE}
}

@inproceedings{khoshavi2020shieldenn,
  title={Shieldenn: Online accelerated framework for fault-tolerant deep neural network architectures},
  author={Khoshavi, Navid and Roohi, Arman and Broyles, Connor and Sargolzaei, Saman and Bi, Yu and Pan, David Z},
  booktitle={2020 57th ACM/IEEE Design Automation Conference (DAC)},
  pages={1--6},
  year={2020},
  organization={IEEE}
}

@inproceedings{chitty2020model,
  title={Model Compression on Faulty Array-based Neural Network Accelerator},
  author={Chitty-Venkata, Krishna Teja and Somani, Arun K},
  booktitle={2020 IEEE 25th Pacific Rim International Symposium on Dependable Computing (PRDC)},
  pages={90--99},
  year={2020},
  organization={IEEE}
}

@article{dos2018analyzing,
  title={Analyzing and increasing the reliability of convolutional neural networks on GPUs},
  author={dos Santos, Fernando Fernandes and Pimenta, Pedro Foletto and Lunardi, Caio and Draghetti, Lucas and Carro, Luigi and Kaeli, David and Rech, Paolo},
  journal={IEEE Transactions on Reliability},
  volume={68},
  number={2},
  pages={663--677},
  year={2018},
  publisher={IEEE}
}

@article{lavallee2013performing,
  title={Performing systematic literature reviews with novices: An iterative approach},
  author={Lavall{\'e}e, Mathieu and Robillard, Pierre-N and Mirsalari, Reza},
  journal={IEEE Transactions on Education},
  volume={57},
  number={3},
  pages={175--181},
  year={2013},
  publisher={IEEE}
}

@article{cicchetti2019multi,
  title={Multi-view approaches for software and system modelling: a systematic literature review},
  author={Cicchetti, Antonio and Ciccozzi, Federico and Pierantonio, Alfonso},
  journal={Software and Systems Modeling},
  volume={18},
  number={6},
  pages={3207--3233},
  year={2019},
  publisher={Springer}
}

@article{sze2017efficient,
  title={Efficient processing of deep neural networks: A tutorial and survey},
  author={Sze, Vivienne and Chen, Yu-Hsin and Yang, Tien-Ju and Emer, Joel S},
  journal={Proceedings of the IEEE},
  volume={105},
  number={12},
  pages={2295--2329},
  year={2017},
  publisher={Ieee}
}

@article{lecun1998gradient,
  title={Gradient-based learning applied to document recognition},
  author={LeCun, Yann and Bottou, L{\'e}on and Bengio, Yoshua and Haffner, Patrick},
  journal={Proceedings of the IEEE},
  volume={86},
  number={11},
  pages={2278--2324},
  year={1998},
  publisher={Ieee}
}

@article{krizhevsky2012imagenet,
  title={Imagenet classification with deep convolutional neural networks},
  author={Krizhevsky, Alex and Sutskever, Ilya and Hinton, Geoffrey E},
  journal={Advances in neural information processing systems},
  volume={25},
  pages={1097--1105},
  year={2012}
}

@inproceedings{szegedy2015going,
  title={Going deeper with convolutions},
  author={Szegedy, Christian and Liu, Wei and Jia, Yangqing and Sermanet, Pierre and Reed, Scott and Anguelov, Dragomir and Erhan, Dumitru and Vanhoucke, Vincent and Rabinovich, Andrew},
  booktitle={Proceedings of the IEEE conference on computer vision and pattern recognition},
  pages={1--9},
  year={2015}
}

@article{simonyan2014very,
  title={Very deep convolutional networks for large-scale image recognition},
  author={Simonyan, Karen and Zisserman, Andrew},
  journal={arXiv preprint arXiv:1409.1556},
  year={2014}
}

@inproceedings{he2016deep,
  title={Deep residual learning for image recognition},
  author={He, Kaiming and Zhang, Xiangyu and Ren, Shaoqing and Sun, Jian},
  booktitle={Proceedings of the IEEE conference on computer vision and pattern recognition},
  pages={770--778},
  year={2016}
}

@inproceedings{redmon2016you,
  title={You only look once: Unified, real-time object detection},
  author={Redmon, Joseph and Divvala, Santosh and Girshick, Ross and Farhadi, Ali},
  booktitle={Proceedings of the IEEE conference on computer vision and pattern recognition},
  pages={779--788},
  year={2016}
}

@inproceedings{deng2009imagenet,
  title={Imagenet: A large-scale hierarchical image database},
  author={Deng, Jia and Dong, Wei and Socher, Richard and Li, Li-Jia and Li, Kai and Fei-Fei, Li},
  booktitle={2009 IEEE conference on computer vision and pattern recognition},
  pages={248--255},
  year={2009},
  organization={Ieee}
}

@misc{MNIST_DS,
  author = {Yann, C. J. B. and LeCun, Y. and Cortes, C.},
  title = {{"The MNIST DATABASE of Handwritten Digits"}},
  howpublished = "\url{http://yann.lecun.com/exdb/mnist/}",
  year = {}, 
  note = "[Online]"
}

@misc{CIFAR_DS,
  author = {Krizhevsky, A. and Nair, v. and Hinton, G.},
  title = {{"The CIFAR-10 Dataset"}},
  howpublished = "\url{https://www.cs.toronto.edu/~kriz/cifar.html}",
  year = {}, 
  note = "[Online]"
}

@article{hubara2017quantized,
  title={Quantized neural networks: Training neural networks with low precision weights and activations},
  author={Hubara, Itay and Courbariaux, Matthieu and Soudry, Daniel and El-Yaniv, Ran and Bengio, Yoshua},
  journal={The Journal of Machine Learning Research},
  volume={18},
  number={1},
  pages={6869--6898},
  year={2017},
  publisher={JMLR. org}
}

@article{courbariaux2016binarized,
  title={Binarized neural networks: Training deep neural networks with weights and activations constrained to+ 1 or-1},
  author={Courbariaux, Matthieu and Hubara, Itay and Soudry, Daniel and El-Yaniv, Ran and Bengio, Yoshua},
  journal={arXiv preprint arXiv:1602.02830},
  year={2016}
}

@inproceedings{abadi2016tensorflow,
  title={Tensorflow: A system for large-scale machine learning},
  author={Abadi, Mart{\'\i}n and Barham, Paul and Chen, Jianmin and Chen, Zhifeng and Davis, Andy and Dean, Jeffrey and Devin, Matthieu and Ghemawat, Sanjay and Irving, Geoffrey and Isard, Michael and others},
  booktitle={12th $\{$USENIX$\}$ symposium on operating systems design and implementation ($\{$OSDI$\}$ 16)},
  pages={265--283},
  year={2016}
}

@article{paszke2019pytorch,
  title={Pytorch: An imperative style, high-performance deep learning library},
  author={Paszke, Adam and Gross, Sam and Massa, Francisco and Lerer, Adam and Bradbury, James and Chanan, Gregory and Killeen, Trevor and Lin, Zeming and Gimelshein, Natalia and Antiga, Luca and others},
  journal={Advances in neural information processing systems},
  volume={32},
  pages={8026--8037},
  year={2019}
}

@misc{keras_tool,
  author = {},
  title = {{"Keras: The python deep learning API"}},
  howpublished = "\url{https://keras.io/}",
  year = {}, 
  note = "[Online]"
}

@misc{darknet13,
  author = {Joseph Redmon},
  title = {{"Darknet: Open Source Neural Networks in C"}},
  howpublished = {\url{http://pjreddie.com/darknet/}},
  year = {},
  note = "[Online]"
}

@misc{tinydnn,
  author = {},
  title = {{"Tiny-CNN Framework"}},
  howpublished = {\url{https://github.com/tiny-dnn/tiny-dnn}},
  year = {},
  note = "[Online]"
}

@misc{tensorrt,
  author = {NVIDIA Corporation},
  title = {{"NVIDIA TensorRT"}},
  howpublished = "\url{https://developer.nvidia.com/tensorrt}",
  year = {}, 
  note = "[Online]"
}

@article{guo2019dl,
  title={[DL] A survey of FPGA-based neural network inference accelerators},
  author={Guo, Kaiyuan and Zeng, Shulin and Yu, Jincheng and Wang, Yu and Yang, Huazhong},
  journal={ACM Transactions on Reconfigurable Technology and Systems (TRETS)},
  volume={12},
  number={1},
  pages={1--26},
  year={2019},
  publisher={ACM New York, NY, USA}
}

@article{dhouibi2021accelerating,
  title={Accelerating Deep Neural Networks implementation: A survey},
  author={Dhouibi, Meriam and Ben Salem, Ahmed Karim and Saidi, Afef and Ben Saoud, Slim},
  journal={IET Computers \& Digital Techniques},
  volume={15},
  number={2},
  pages={79--96},
  year={2021}
}

@inproceedings{jouppi2017datacenter,
  title={In-datacenter performance analysis of a tensor processing unit},
  author={Jouppi, Norman P and Young, Cliff and Patil, Nishant and Patterson, David and Agrawal, Gaurav and Bajwa, Raminder and Bates, Sarah and Bhatia, Suresh and Boden, Nan and Borchers, Al and others},
  booktitle={Proceedings of the 44th annual international symposium on computer architecture},
  pages={1--12},
  year={2017}
}

@article{moolchandani2021accelerating,
  title={Accelerating CNN inference on ASICs: A survey},
  author={Moolchandani, Diksha and Kumar, Anshul and Sarangi, Smruti R},
  journal={Journal of Systems Architecture},
  volume={113},
  pages={101887},
  year={2021},
  publisher={Elsevier}
}

@article{johnson1984fault,
  title={Fault-tolerant microprocessor-based systems},
  author={Johnson, Barry},
  journal={IEEE Micro},
  volume={4},
  number={06},
  pages={6--21},
  year={1984},
  publisher={IEEE Computer Society}
}

@misc{koren2007fault,
  title={Fault-Tolerant Systems},
  author={Koren, Israel and Krishna, C Mani},
  year={2007},
  publisher={Morgan Kaufmann Publishers Inc.}
}

@article{ibrahim2020softerror,
  title={Soft error resilience of deep residual networks for object recognition},
  author={Ibrahim, Younis and Wang, Haibin and Bai, Man and Liu, Zhi and Wang, Jianan and Yang, Zhiming and Chen, Zhengming},
  journal={IEEE Access},
  volume={8},
  pages={19490--19503},
  year={2020},
  publisher={IEEE}
}

@inproceedings{biswas2005computing,
  title={Computing architectural vulnerability factors for address-based structures},
  author={Biswas, Arijit and Racunas, Paul and Cheveresan, Razvan and Emer, Joel and Mukherjee, Shubhendu S and Rangan, Ram},
  booktitle={32nd International Symposium on Computer Architecture (ISCA'05)},
  pages={532--543},
  year={2005},
  organization={IEEE}
}

@article{eslami2020survey,
  title={A survey on fault injection methods of digital integrated circuits},
  author={Eslami, Mohammad and Ghavami, Behnam and Raji, Mohsen and Mahani, Ali},
  journal={Integration},
  volume={71},
  pages={154--163},
  year={2020},
  publisher={Elsevier}
}

@article{benso2011art,
  title={The art of fault injection},
  author={Benso, Alfredo and DiCarlo, Stefano},
  journal={Journal of Control Engineering and Applied Informatics},
  volume={13},
  number={4},
  pages={9--18},
  year={2011}
}

@inproceedings{ruospo2021pros,
  title={Pros and Cons of Fault Injection Approaches for the Reliability Assessment of Deep Neural Networks},
  author={Ruospo, Annachiara and Luza, Lucas Matana and Bosio, Alberto and Traiola, Marcello and Dilillo, Luigi and Sanchez, Ernesto},
  booktitle={2021 IEEE 22nd Latin American Test Symposium (LATS)},
  pages={1--5},
  year={2021},
  organization={IEEE}
}

@inproceedings{ruospo2020pipelined,
  title={A Pipelined Multi-Level Fault Injector for Deep Neural Networks},
  author={Ruospo, Annachiara and Balaara, Angelo and Bosio, Alberto and Sanchez, Ernesto},
  booktitle={2020 IEEE International Symposium on Defect and Fault Tolerance in VLSI and Nanotechnology Systems (DFT)},
  pages={1--6},
  year={2020},
  organization={IEEE}
}

@inproceedings{leveugle2009statistical,
  title={Statistical fault injection: Quantified error and confidence},
  author={Leveugle, R{\'e}gis and Calvez, A and Maistri, Paolo and Vanhauwaert, Pierre},
  booktitle={2009 Design, Automation \& Test in Europe Conference \& Exhibition},
  pages={502--506},
  year={2009},
  organization={IEEE}
}

@article{ali2020erdnn,
  title={ERDNN: Error-Resilient Deep Neural Networks With a New Error Correction Layer and Piece-Wise Rectified Linear Unit},
  author={Ali, Muhammad Salman and Iqbal, Tauhid Bin and Lee, Kang-Ho and Muqeet, Abdul and Lee, Seunghyun and Kim, Lokwon and Bae, Sung-Ho},
  journal={IEEE Access},
  volume={8},
  pages={158702--158711},
  year={2020},
  publisher={IEEE}
}

@inproceedings{hoang2020ft,
  title={Ft-clipact: Resilience analysis of deep neural networks and improving their fault tolerance using clipped activation},
  author={Hoang, Le-Ha and Hanif, Muhammad Abdullah and Shafique, Muhammad},
  booktitle={2020 Design, Automation \& Test in Europe Conference \& Exhibition (DATE)},
  pages={1241--1246},
  year={2020},
  organization={IEEE}
}

@inproceedings{goldstein2020reliability,
  title={Reliability evaluation of compressed deep learning models},
  author={Goldstein, Brunno F and Srinivasan, Sudarshan and Das, Dipankar and Banerjee, Kunal and Santiago, Leandro and Ferreira, Victor C and Nery, Alexandre S and Kundu, Sandip and Fran{\c{c}}a, Felipe MG},
  booktitle={2020 IEEE 11th Latin American Symposium on Circuits \& Systems (LASCAS)},
  pages={1--5},
  year={2020},
  organization={IEEE}
}

@inproceedings{neggaz2018reliability,
  title={A reliability study on CNNs for critical embedded systems},
  author={Neggaz, Mohamed A and Alouani, Ihsen and Lorenzo, Pablo R and Niar, Smail},
  booktitle={2018 IEEE 36th International Conference on Computer Design (ICCD)},
  pages={476--479},
  year={2018},
  organization={IEEE}
}

@article{neggaz2019cnns,
  title={Are cnns reliable enough for critical applications? an exploratory study},
  author={Neggaz, Mohamed A and Alouani, Ihsen and Niar, Smail and Kurdahi, Fadi},
  journal={IEEE Design \& Test},
  volume={37},
  number={2},
  pages={76--83},
  year={2019},
  publisher={IEEE}
}

@inproceedings{gao2020reliability,
  title={Reliability Evaluation of Pruned Neural Networks against Errors on Parameters},
  author={Gao, Zhen and Wei, Xiaohui and Zhang, Han and Li, Wenshuo and Ge, Guangjun and Wang, Yu and Reviriego, Pedro},
  booktitle={2020 IEEE International Symposium on Defect and Fault Tolerance in VLSI and Nanotechnology Systems (DFT)},
  pages={1--6},
  year={2020},
  organization={IEEE}
}

@inproceedings{guan2019place,
  title={In-place zero-space memory protection for CNN},
  author={Guan, Hui and Ning, Lin and Lin, Zhen and Shen, Xipeng and Zhou, Huiyang and Lim, Seung-Hwan},
  booktitle={Proceedings of the 33rd International Conference on Neural Information Processing Systems},
  pages={5734--5743},
  year={2019}
}

@inproceedings{sabbagh2019evaluating,
  title={Evaluating fault resiliency of compressed deep neural networks},
  author={Sabbagh, Majid and Gongye, Cheng and Fei, Yunsi and Wang, Yanzhi},
  booktitle={2019 IEEE International Conference on Embedded Software and Systems (ICESS)},
  pages={1--7},
  year={2019},
  organization={IEEE}
}

@inproceedings{arechiga2018effect,
  title={The effect of weight errors on neural networks},
  author={Arechiga, Austin P and Michaels, Alan J},
  booktitle={2018 IEEE 8th Annual Computing and Communication Workshop and Conference (CCWC)},
  pages={190--196},
  year={2018},
  organization={IEEE}
}

@inproceedings{arechiga2018robustness,
  title={The robustness of modern deep learning architectures against single event upset errors},
  author={Arechiga, Austin P and Michaels, Alan J},
  booktitle={2018 IEEE High Performance extreme Computing Conference (HPEC)},
  pages={1--6},
  year={2018},
  organization={IEEE}
}

@inproceedings{schorn2019efficient,
  title={An efficient bit-flip resilience optimization method for deep neural networks},
  author={Schorn, Christoph and Guntoro, Andre and Ascheid, Gerd},
  booktitle={2019 Design, Automation \& Test in Europe Conference \& Exhibition (DATE)},
  pages={1507--1512},
  year={2019},
  organization={IEEE}
}

@inproceedings{chen2021low,
  title={A low-cost fault corrector for deep neural networks through range restriction},
  author={Chen, Zitao and Li, Guanpeng and Pattabiraman, Karthik},
  booktitle={2021 51st Annual IEEE/IFIP International Conference on Dependable Systems and Networks (DSN)},
  pages={1--13},
  year={2021},
  organization={IEEE}
}

@inproceedings{ponader2021milr,
  title={MILR: Mathematically Induced Layer Recovery for Plaintext Space Error Correction of CNNs},
  author={Ponader, Jonathan and Thomas, Kyle and Kundu, Sandip and Solihin, Yan},
  booktitle={2021 51st Annual IEEE/IFIP International Conference on Dependable Systems and Networks (DSN)},
  pages={75--87},
  year={2021},
  organization={IEEE}
}

@inproceedings{ruospo2020evaluating,
  title={Evaluating convolutional neural networks reliability depending on their data representation},
  author={Ruospo, Annachiara and Bosio, Alberto and Ianne, Alessandro and Sanchez, Ernesto},
  booktitle={2020 23rd Euromicro Conference on Digital System Design (DSD)},
  pages={672--679},
  year={2020},
  organization={IEEE}
}

@inproceedings{bosio2019reliability,
  title={A reliability analysis of a deep neural network},
  author={Bosio, Alberto and Bernardi, Paolo and Ruospo, Annachiara and Sanchez, Ernesto},
  booktitle={2019 IEEE Latin American Test Symposium (LATS)},
  pages={1--6},
  year={2019},
  organization={IEEE}
}

@inproceedings{yan2020single,
  title={When single event upset meets deep neural networks: Observations, explorations, and remedies},
  author={Yan, Zheyu and Shi, Yiyu and Liao, Wang and Hashimoto, Masanori and Zhou, Xichuan and Zhuo, Cheng},
  booktitle={2020 25th Asia and South Pacific Design Automation Conference (ASP-DAC)},
  pages={163--168},
  year={2020},
  organization={IEEE}
}

@inproceedings{cantoro2020evaluating,
  title={Evaluating Data Encryption Effects on the Resilience of an Artificial Neural Network},
  author={Cantoro, Riccardo and Deligiannis, Nikolaos I and Reorda, Matteo Sonza and Traiola, Marcello and Valea, Emanuele},
  booktitle={2020 IEEE International Symposium on Defect and Fault Tolerance in VLSI and Nanotechnology Systems (DFT)},
  pages={1--4},
  year={2020},
  organization={IEEE}
}

@inproceedings{burel2021zero,
  title={Zero-Overhead Protection for CNN Weights},
  author={Burel, St{\'e}phane and Evans, Adrian and Anghel, Lorena},
  booktitle={2021 IEEE International Symposium on Defect and Fault Tolerance in VLSI and Nanotechnology Systems (DFT)},
  pages={1--6},
  year={2021},
  organization={IEEE}
}

@inproceedings{mahmoud2020pytorchfi,
  title={Pytorchfi: A runtime perturbation tool for dnns},
  author={Mahmoud, Abdulrahman and Aggarwal, Neeraj and Nobbe, Alex and Vicarte, Jose Rodrigo Sanchez and Adve, Sarita V and Fletcher, Christopher W and Frosio, Iuri and Hari, Siva Kumar Sastry},
  booktitle={2020 50th Annual IEEE/IFIP International Conference on Dependable Systems and Networks Workshops (DSN-W)},
  pages={25--31},
  year={2020},
  organization={IEEE}
}

@inproceedings{li2018tensorfi,
  title={Tensorfi: A configurable fault injector for tensorflow applications},
  author={Li, Guanpeng and Pattabiraman, Karthik and DeBardeleben, Nathan},
  booktitle={2018 IEEE International symposium on software reliability engineering workshops (ISSREW)},
  pages={313--320},
  year={2018},
  organization={IEEE}
}

@inproceedings{chen2020tensorfi,
  title={TensorFI: A flexible fault injection framework for TensorFlow applications},
  author={Chen, Zitao and Narayanan, Niranjhana and Fang, Bo and Li, Guanpeng and Pattabiraman, Karthik and DeBardeleben, Nathan},
  booktitle={2020 IEEE 31st International Symposium on Software Reliability Engineering (ISSRE)},
  pages={426--435},
  year={2020},
  organization={IEEE}
}

@inproceedings{reagen2018ares,
  title={Ares: A framework for quantifying the resilience of deep neural networks},
  author={Reagen, Brandon and Gupta, Udit and Pentecost, Lillian and Whatmough, Paul and Lee, Sae Kyu and Mulholland, Niamh and Brooks, David and Wei, Gu-Yeon},
  booktitle={2018 55th ACM/ESDA/IEEE Design Automation Conference (DAC)},
  pages={1--6},
  year={2018},
  organization={IEEE}
}

@inproceedings{chen2019binfi,
  title={BinFI: an efficient fault injector for safety-critical machine learning systems},
  author={Chen, Zitao and Li, Guanpeng and Pattabiraman, Karthik and DeBardeleben, Nathan},
  booktitle={Proceedings of the International Conference for High Performance Computing, Networking, Storage and Analysis},
  pages={1--23},
  year={2019}
}

@inproceedings{li2017understanding,
  title={Understanding error propagation in deep learning neural network (DNN) accelerators and applications},
  author={Li, Guanpeng and Hari, Siva Kumar Sastry and Sullivan, Michael and Tsai, Timothy and Pattabiraman, Karthik and Emer, Joel and Keckler, Stephen W},
  booktitle={Proceedings of the International Conference for High Performance Computing, Networking, Storage and Analysis},
  pages={1--12},
  year={2017}
}

@inproceedings{azizimazreah2018tolerating,
  title={Tolerating soft errors in deep learning accelerators with reliable on-chip memory designs},
  author={Azizimazreah, Arash and Gu, Yongbin and Gu, Xiang and Chen, Lizhong},
  booktitle={2018 IEEE International Conference on Networking, Architecture and Storage (NAS)},
  pages={1--10},
  year={2018},
  organization={IEEE}
}

@inproceedings{li2020soft,
  title={Soft Error Mitigation for Deep Convolution Neural Network on FPGA Accelerators},
  author={Li, Wenshuo and Ge, Guangjun and Guo, Kaiyuan and Chen, Xiaoming and Wei, Qi and Gao, Zhen and Wang, Yu and Yang, Huazhong},
  booktitle={2020 2nd IEEE International Conference on Artificial Intelligence Circuits and Systems (AICAS)},
  pages={1--5},
  year={2020},
  organization={IEEE}
}

@inproceedings{ozen2019sanity,
  author={Ozen, Elbruz and Orailoglu, Alex},
  title={Sanity-Check: Boosting the Reliability of Safety-Critical Deep Neural Network Applications}, 
  booktitle={2019 IEEE 28th Asian Test Symposium (ATS)}, 
  pages={7-75},
  year={2019},
  organization={IEEE}
}

@inproceedings{kim2019dris,
  title={DRIS-3: Deep neural network reliability improvement scheme in 3D die-stacked memory based on fault analysis},
  author={Kim, Jae-San and Yang, Joon-Sung},
  booktitle={2019 56th ACM/IEEE Design Automation Conference (DAC)},
  pages={1--6},
  year={2019},
  organization={IEEE}
}

@inproceedings{ozen2020just,
  title={Just say zero: containing critical bit-error propagation in deep neural networks with anomalous feature suppression},
  author={Ozen, Elbruz and Orailoglu, Alex},
  booktitle={2020 IEEE/ACM International Conference On Computer Aided Design (ICCAD)},
  pages={1--9},
  year={2020},
  organization={IEEE}
}

@article{ozen2020boosting,
  title={Boosting bit-error resilience of DNN accelerators through median feature selection},
  author={Ozen, Elbruz and Orailoglu, Alex},
  journal={IEEE Transactions on Computer-Aided Design of Integrated Circuits and Systems},
  volume={39},
  number={11},
  pages={3250--3262},
  year={2020},
  publisher={IEEE}
}

@inproceedings{goldstein2021lightweight,
  title={A Lightweight Error-Resiliency Mechanism for Deep Neural Networks},
  author={Goldstein, Brunno F and Ferreira, Victor C and Srinivasan, Sudarshan and Das, Dipankar and Nery, Alexandre S and Kundu, Sandip and Fran{\c{c}}a, Felipe MG},
  booktitle={2021 22nd International Symposium on Quality Electronic Design (ISQED)},
  pages={311--316},
  year={2021},
  organization={IEEE}
}

@misc{N2D2_fr,
  author = {},
  title = {{"N2D2 CAD framework for DNNs"}},
  howpublished = "\url{https://github.com/cea-list/N2D2}",
  year = {}, 
  note = "[Online]"
}

@inproceedings{zahid2020fat,
  title={FAT: Training Neural Networks for Reliable Inference Under Hardware Faults},
  author={Zahid, Ussama and Gambardella, Giulio and Fraser, Nicholas J and Blott, Michaela and Vissers, Kees},
  booktitle={2020 IEEE International Test Conference (ITC)},
  pages={1--10},
  year={2020},
  organization={IEEE}
}

@inproceedings{hoang2021tre,
  title={TRe-Map: Towards Reducing the Overheads of Fault-Aware Retraining of Deep Neural Networks by Merging Fault Maps},
  author={Hoang, Le-Ha and Hanif, Muhammad Abdullah and Shafique, Muhammad},
  booktitle={2021 24th Euromicro Conference on Digital System Design (DSD)},
  pages={434--441},
  year={2021},
  organization={IEEE}
}

@inproceedings{de2020emulation,
  title={An Emulation Platform for Evaluating the Reliability of Deep Neural Networks},
  author={De Sio, Corrado and Azimi, Sarah and Sterpone, Luca},
  booktitle={2020 IEEE International Symposium on Defect and Fault Tolerance in VLSI and Nanotechnology Systems (DFT)},
  pages={1--4},
  year={2020},
  organization={IEEE}
}

@article{wang2021impact,
  title={Impact of Single-Event Upsets on Convolutional Neural Networks in Xilinx Zynq FPGAs},
  author={Wang, H-B and Wang, Y-S and Xiao, J-H and Wang, S-L and Liang, T-J},
  journal={IEEE Transactions on Nuclear Science},
  volume={68},
  number={4},
  pages={394--401},
  year={2021},
  publisher={IEEE}
}

@inproceedings{du2019reliability,
  title={On the reliability of convolutional neural network implementation on SRAM-based FPGA},
  author={Du, Boyang and Azimi, Sarah and De Sio, Corrado and Bozzoli, Ludovica and Sterpone, Luca},
  booktitle={2019 IEEE International Symposium on Defect and Fault Tolerance in VLSI and Nanotechnology Systems (DFT)},
  pages={1--6},
  year={2019},
  organization={IEEE}
}

@inproceedings{souvatzoglou2021analyzing,
  title={Analyzing the Single Event Upset Vulnerability of Binarized Neural Networks on SRAM FPGAs},
  author={Souvatzoglou, Ioanna and Papadimitriou, Athanasios and Sari, Aitzan and Vlagkoulis, Vasileios and Psarakis, Mihalis},
  booktitle={2021 IEEE International Symposium on Defect and Fault Tolerance in VLSI and Nanotechnology Systems (DFT)},
  pages={1--6},
  year={2021},
  organization={IEEE}
}

@article{xu2021reliability,
  title={Reliability Evaluation and Analysis of FPGA-Based Neural Network Acceleration System},
  author={Xu, Dawen and Zhu, Ziyang and Liu, Cheng and Wang, Ying and Zhao, Shuang and Zhang, Lei and Liang, Huaguo and Li, Huawei and Cheng, Kwang-Ting},
  journal={IEEE Transactions on Very Large Scale Integration (VLSI) Systems},
  volume={29},
  number={3},
  pages={472--484},
  year={2021},
  publisher={IEEE}
}

@inproceedings{xu2020persistent,
  title={Persistent fault analysis of neural networks on FPGA-based acceleration system},
  author={Xu, Dawen and Zhu, Ziyang and Liu, Cheng and Wang, Ying and Li, Huawei and Zhang, Lei and Cheng, Kwang-Ting},
  booktitle={2020 IEEE 31st International Conference on Application-specific Systems, Architectures and Processors (ASAP)},
  pages={85--92},
  year={2020},
  organization={IEEE}
}

@inproceedings{khoshavi2020compression,
  title={Compression or corruption? a study on the effects of transient faults on bnn inference accelerators},
  author={Khoshavi, Navid and Broyles, Connor and Bi, Yu},
  booktitle={2020 21st International Symposium on Quality Electronic Design (ISQED)},
  pages={99--104},
  year={2020},
  organization={IEEE}
}

@inproceedings{khoshavi2020fiji,
  title={Fiji-FIN: A Fault Injection Framework on Quantized Neural Network Inference Accelerator},
  author={Khoshavi, Navid and Broyles, Connor and Bi, Yu and Roohi, Arman},
  booktitle={2020 19th IEEE International Conference on Machine Learning and Applications (ICMLA)},
  pages={1139--1144},
  year={2020},
  organization={IEEE}
}

@article{libano2018selective,
  title={Selective hardening for neural networks in FPGAs},
  author={Libano, Fabiano and Wilson, Brittany and Anderson, J and Wirthlin, Michael J and Cazzaniga, Carlo and Frost, Christopher and Rech, Paolo},
  journal={IEEE Transactions on Nuclear Science},
  volume={66},
  number={1},
  pages={216--222},
  year={2018},
  publisher={IEEE}
}

@article{libano2020understanding,
  title={Understanding the impact of quantization, accuracy, and radiation on the reliability of convolutional neural networks on FPGAs},
  author={Libano, Fabiano and Wilson, Brittany and Wirthlin, Michael and Rech, Paolo and Brunhaver, John},
  journal={IEEE Transactions on Nuclear Science},
  volume={67},
  number={7},
  pages={1478--1484},
  year={2020},
  publisher={IEEE}
}

@inproceedings{benevenuti2018comparative,
  title={Comparative analysis of inference errors in a neural network implemented in SRAM-based FPGA induced by neutron irradiation and fault injection methods},
  author={Benevenuti, Fabio and Libano, Fabiano and Pouget, Vincent and Kastensmidt, Fernanda Lima and Rech, Paolo},
  booktitle={2018 31st Symposium on Integrated Circuits and Systems Design (SBCCI)},
  pages={1--6},
  year={2018},
  organization={IEEE}
}

@article{matanaluza2021emulating,
  title={Emulating the Effects of Radiation-Induced Soft-Errors for the Reliability Assessment of Neural Networks},
  author={Matanaluza, Lucas and Ruospo, Annachiara and Soderstrom, Daniel and Cazzaniga, Carlo and Kastriotou, Maria and Sanchez, Ernesto and Bosio, Alberto and Dilillo, Luigi},
  journal={IEEE Transactions on Emerging Topics in Computing},
  year={2021},
  publisher={IEEE}
}

@inproceedings{salami2018resilience,
  title={On the resilience of rtl nn accelerators: Fault characterization and mitigation},
  author={Salami, Behzad and Unsal, Osman S and Kestelman, Adrian Cristal},
  booktitle={2018 30th International Symposium on Computer Architecture and High Performance Computing (SBAC-PAD)},
  pages={322--329},
  year={2018},
  organization={IEEE}
}

@inproceedings{gambardella2019efficient,
  title={Efficient error-tolerant quantized neural network accelerators},
  author={Gambardella, Giulio and Kappauf, Johannes and Blott, Michaela and Doehring, Christoph and Kumm, Martin and Zipf, Peter and Vissers, Kees},
  booktitle={2019 IEEE International Symposium on Defect and Fault Tolerance in VLSI and Nanotechnology Systems (DFT)},
  pages={1--6},
  year={2019},
  organization={IEEE}
}

@inproceedings{zhang2018analyzing,
  title={Analyzing and mitigating the impact of permanent faults on a systolic array based neural network accelerator},
  author={Zhang, Jeff Jun and Gu, Tianyu and Basu, Kanad and Garg, Siddharth},
  booktitle={2018 IEEE 36th VLSI Test Symposium (VTS)},
  pages={1--6},
  year={2018},
  organization={IEEE}
}

@article{zhang2019fault,
  title={Fault-tolerant systolic array based accelerators for deep neural network execution},
  author={Zhang, Jeff Jun and Basu, Kanad and Garg, Siddharth},
  journal={IEEE Design \& Test},
  volume={36},
  number={5},
  pages={44--53},
  year={2019},
  publisher={IEEE}
}

@inproceedings{corneliou2021fine,
  title={Fine-Grained Vulnerability Analysis of Resource Constrained Neural Inference Accelerators},
  author={Corneliou, Panayiotis and Nikolaou, Panagiota and Michael, Maria K and Theocharides, Theocharis},
  booktitle={2021 IEEE International Symposium on Defect and Fault Tolerance in VLSI and Nanotechnology Systems (DFT)},
  pages={1--6},
  year={2021},
  organization={IEEE}
}

@inproceedings{xu2020hybrid,
  title={A hybrid computing architecture for fault-tolerant deep learning accelerators},
  author={Xu, Dawen and Chu, Cheng and Wang, Qianlong and Liu, Cheng and Wang, Ying and Zhang, Lei and Liang, Huaguo and Cheng, Kwang-Ting},
  booktitle={2020 IEEE 38th International Conference on Computer Design (ICCD)},
  pages={478--485},
  year={2020},
  organization={IEEE}
}

@inproceedings{abich2020soft,
  title={Soft Error Reliability Assessment of Neural Networks on Resource-constrained IoT Devices},
  author={Abich, Geancarlo and Gava, Jonas and Reis, Ricardo and Ost, Luciano},
  booktitle={2020 27th IEEE International Conference on Electronics, Circuits and Systems (ICECS)},
  pages={1--4},
  year={2020},
  organization={IEEE}
}

@inproceedings{bandeira2019non,
  title={Non-intrusive fault injection techniques for efficient soft error vulnerability analysis},
  author={Bandeira, Vitor and Rosa, Felipe and Reis, Ricardo and Ost, Luciano},
  booktitle={2019 IFIP/IEEE 27th International Conference on Very Large Scale Integration (VLSI-SoC)},
  pages={123--128},
  year={2019},
  organization={IEEE}
}

@article{chen2016eyeriss,
  title={Eyeriss: A spatial architecture for energy-efficient dataflow for convolutional neural networks},
  author={Chen, Yu-Hsin and Emer, Joel and Sze, Vivienne},
  journal={ACM SIGARCH Computer Architecture News},
  volume={44},
  number={3},
  pages={367--379},
  year={2016},
  publisher={ACM New York, NY, USA}
}

@inproceedings{lotfi2019resiliency,
  title={Resiliency of automotive object detection networks on GPU architectures},
  author={Lotfi, Atieh and Hukerikar, Saurabh and Balasubramanian, Keshav and Racunas, Paul and Saxena, Nirmal and Bramley, Richard and Huang, Yanxiang},
  booktitle={2019 IEEE International Test Conference (ITC)},
  pages={1--9},
  year={2019},
  organization={IEEE}
}

@article{hari2021making,
  title={Making convolutions resilient via algorithm-based error detection techniques},
  author={Hari, Siva Kumar Sastry and Sullivan, Michael and Tsai, Timothy and Keckler, Stephen W},
  journal={IEEE Transactions on Dependable and Secure Computing},
  year={2021},
  publisher={IEEE}
}

@article{condia2020flexgripplus,
  title={FlexGripPlus: An improved GPGPU model to support reliability analysis},
  author={Condia, Josie E Rodriguez and Du, Boyang and Reorda, Matteo Sonza and Sterpone, Luca},
  journal={Microelectronics Reliability},
  volume={109},
  pages={113660},
  year={2020},
  publisher={Elsevier}
}

@inproceedings{tsai2021nvbitfi,
  title={Nvbitfi: dynamic fault injection for gpus},
  author={Tsai, Timothy and Hari, Siva Kumar Sastry and Sullivan, Michael and Villa, Oreste and Keckler, Stephen W},
  booktitle={2021 51st Annual IEEE/IFIP International Conference on Dependable Systems and Networks (DSN)},
  pages={284--291},
  year={2021},
  organization={IEEE}
}

@inproceedings{oliveira2017experimental,
  title={Experimental and analytical study of xeon phi reliability},
  author={Oliveira, Daniel and Pilla, La{\'e}rcio and DeBardeleben, Nathan and Blanchard, Sean and Quinn, Heather and Koren, Israel and Navaux, Philippe and Rech, Paolo},
  booktitle={Proceedings of the International Conference for High Performance Computing, Networking, Storage and Analysis},
  pages={1--12},
  year={2017}
}

@inproceedings{condia2021combining,
  title={Combining Architectural Simulation and Software Fault Injection for a Fast and Accurate CNNs Reliability Evaluation on GPUs},
  author={Condia, Josie E Rodriguez and dos Santos, Fernando Fernandes and Reorda, Matteo Sonza and Rech, Paolo},
  booktitle={2021 IEEE 39th VLSI Test Symposium (VTS)},
  pages={1--7},
  year={2021},
  organization={IEEE}
}

@inproceedings{rech2020impact,
  title={Impact of Layers Selective Approximation on CNNs Reliability and Performance},
  author={Rech, Rubens Luiz and Rech, Paolo},
  booktitle={2020 IEEE International Symposium on Defect and Fault Tolerance in VLSI and Nanotechnology Systems (DFT)},
  pages={1--4},
  year={2020},
  organization={IEEE}
}

@inproceedings{dos2019impact,
  title={Impact of reduced precision in the reliability of deep neural networks for object detection},
  author={dos Santos, Fernando Fernandes and Navaux, Philippe and Carro, Luigi and Rech, Paolo},
  booktitle={2019 IEEE European Test Symposium (ETS)},
  pages={1--6},
  year={2019},
  organization={IEEE}
}

@article{adam2021selective,
  title={A Selective Mitigation Technique of Soft Errors for DNN Models Used in Healthcare Applications: DenseNet201 Case Study},
  author={Adam, Khalid and Mohamed, Izzeldin Ibrahim and Ibrahim, Younis},
  journal={IEEE Access},
  volume={9},
  pages={65803--65823},
  year={2021},
  publisher={IEEE}
}

@inproceedings{dos2017evaluation,
  title={Evaluation and mitigation of soft-errors in neural network-based object detection in three GPU architectures},
  author={dos Santos, Fernando Fernandes and Draghetti, Lucas and Weigel, Lucas and Carro, Luigi and Navaux, Philippe and Rech, Paolo},
  booktitle={2017 47th Annual IEEE/IFIP International Conference on Dependable Systems and Networks Workshops (DSN-W)},
  pages={169--176},
  year={2017},
  organization={IEEE}
}

@inproceedings{luza2020investigating,
  title={Investigating the impact of radiation-induced soft errors on the reliability of approximate computing systems},
  author={Luza, Lucas Matana and S{\"o}derstr{\"o}m, Daniel and Tsiligiannis, Georgios and Puchner, Helmut and Cazzaniga, Carlo and Sanchez, Ernesto and Bosio, Alberto and Dilillo, Luigi},
  booktitle={2020 IEEE International Symposium on Defect and Fault Tolerance in VLSI and Nanotechnology Systems (DFT)},
  pages={1--6},
  year={2020},
  organization={IEEE}
}

@article{libano2021reduced,
  title={How Reduced Data Precision and Degree of Parallelism Impact the Reliability of Convolutional Neural Networks on FPGAs},
  author={Libano, F and Rech, P and Neuman, B and Leavitt, J and Wirthlin, M and Brunhaver, J},
  journal={IEEE Transactions on Nuclear Science},
  volume={68},
  number={5},
  pages={865--872},
  year={2021},
  publisher={IEEE}
}

@inproceedings{ibrahim2020analyzing,
  title={Analyzing the Reliability of Convolutional Neural Networks on GPUs: GoogLeNet as a Case Study},
  author={Ibrahim, Younis and Wang, Haibin and Adam, Khalid},
  booktitle={2020 International Conference on Computing and Information Technology (ICCIT-1441)},
  pages={1--6},
  year={2020},
  organization={IEEE}
}

@article{ibrahim2020analyzinginstruction,
  title={Analyzing the Impact of Soft Errors in Deep Neural Networks on GPUs from Instruction Level},
  author={Ibrahin, Younis and Liu, Junyang and Yang, Xuanxuan and Sha, Hongwei and Wang, Haibin},
  journal={WSEAS Transaction on Systems and Control},
  volume={15},
  pages={699--708},
  year={2020},
  publisher={IEEE}
}

@inproceedings{hari2017sassifi,
  title={Sassifi: An architecture-level fault injection tool for gpu application resilience evaluation},
  author={Hari, Siva Kumar Sastry and Tsai, Timothy and Stephenson, Mark and Keckler, Stephen W and Emer, Joel},
  booktitle={2017 IEEE International Symposium on Performance Analysis of Systems and Software (ISPASS)},
  pages={249--258},
  year={2017},
  organization={IEEE}
}

@article{basso2020impact,
  title={Impact of tensor cores and mixed precision on the reliability of matrix multiplication in GPUs},
  author={Basso, Pedro Martins and dos Santos, Fernando Fernandes and Rech, Paolo},
  journal={IEEE Transactions on Nuclear Science},
  volume={67},
  number={7},
  pages={1560--1565},
  year={2020},
  publisher={IEEE}
}

@inproceedings{schorn2018accurate,
  title={Accurate neuron resilience prediction for a flexible reliability management in neural network accelerators},
  author={Schorn, Christoph and Guntoro, Andre and Ascheid, Gerd},
  booktitle={2018 Design, Automation \& Test in Europe Conference \& Exhibition (DATE)},
  pages={979--984},
  year={2018},
  organization={IEEE}
}

@article{abdullah2020salvagednn,
  title={Salvagednn: salvaging deep neural network accelerators with permanent faults through saliency-driven fault-aware mapping},
  author={Abdullah Hanif, Muhammad and Shafique, Muhammad},
  journal={Philosophical Transactions of the Royal Society A},
  volume={378},
  number={2164},
  pages={20190164},
  year={2020},
  publisher={The Royal Society Publishing}
}

@article{ruospo2021reliability,
  title={On the Reliability Assessment of Artificial Neural Networks Running on AI-Oriented MPSoCs},
  author={Ruospo, Annachiara and Sanchez, Ernesto},
  journal={Applied Sciences},
  volume={11},
  number={14},
  pages={6455},
  year={2021},
  publisher={Multidisciplinary Digital Publishing Institute}
}

@inproceedings{sabih2021fault,
  title={Fault-tolerant low-precision dnns using explainable ai},
  author={Sabih, Muhammad and Hannig, Frank and Teich, J{\"u}rgen},
  booktitle={2021 51st Annual IEEE/IFIP International Conference on Dependable Systems and Networks Workshops (DSN-W)},
  pages={166--174},
  year={2021},
  organization={IEEE}
}

@inproceedings{ping2020sern,
  title={SERN: Modeling and Analyzing the Soft Error Reliability of Convolutional Neural Networks},
  author={Ping, Liqi and Tan, Jingweijia and Yan, Kaige},
  booktitle={Proceedings of the 2020 on Great Lakes Symposium on VLSI},
  pages={445--450},
  year={2020}
}

@article{mahmoud2020hardnn,
  title={HarDNN: Feature Map Vulnerability Evaluation in CNNs},
  author={Mahmoud, Abdulrahman and Hari, Siva Kumar Sastry and Fletcher, Christopher W and Adve, Sarita V and Sakr, Charbel and Shanbhag, Naresh and Molchanov, Pavlo and Sullivan, Michael B and Tsai, Timothy and Keckler, Stephen W},
  journal={arXiv preprint arXiv:2002.09786},
  year={2020}
}

@inproceedings{he2020fidelity,
  title={Fidelity: Efficient resilience analysis framework for deep learning accelerators},
  author={He, Yi and Balaprakash, Prasanna and Li, Yanjing},
  booktitle={2020 53rd Annual IEEE/ACM International Symposium on Microarchitecture (MICRO)},
  pages={270--281},
  year={2020},
  organization={IEEE}
}

@misc{NVDLA,
  author = {NVIDIA Corporation},
  title = {{"NVDLA Open Source Project"}},
  howpublished = "\url{http://nvdla.org}",
  year = {}, 
  note = "[Online]"
}

@inproceedings{umuroglu2017finn,
  title={Finn: A framework for fast, scalable binarized neural network inference},
  author={Umuroglu, Yaman and Fraser, Nicholas J and Gambardella, Giulio and Blott, Michaela and Leong, Philip and Jahre, Magnus and Vissers, Kees},
  booktitle={Proceedings of the 2017 ACM/SIGDA International Symposium on Field-Programmable Gate Arrays},
  pages={65--74},
  year={2017}
}

@inproceedings{menze2015object,
  title={Object scene flow for autonomous vehicles},
  author={Menze, Moritz and Geiger, Andreas},
  booktitle={Proceedings of the IEEE conference on computer vision and pattern recognition},
  pages={3061--3070},
  year={2015}
}

@article{everingham2010pascal,
  title={The pascal visual object classes (voc) challenge},
  author={Everingham, Mark and Van Gool, Luc and Williams, Christopher KI and Winn, John and Zisserman, Andrew},
  journal={International journal of computer vision},
  volume={88},
  number={2},
  pages={303--338},
  year={2010},
  publisher={Springer}
}

@article{hou2019survey,
  title={A survey on partitioning models, solution algorithms and algorithm parallelization for hardware/software co-design},
  author={Hou, Neng and Yan, Xiaohu and He, Fazhi},
  journal={Design Automation for Embedded Systems},
  volume={23},
  number={1},
  pages={57--77},
  year={2019},
  publisher={Springer}
}

@misc{zynq_xilinx,
  author = {XILINX},
  title = {{"SoCs with Hardware and Software Programmability"}},
  howpublished = "\url{https://www.xilinx.com/products/silicon-devices/soc/zynq-7000.html}",
  year = {}, 
  note = "[Online]"
}

@inproceedings{luza2021model,
  title={A Model-Based Framework to Assess the Reliability of Safety-Critical Applications},
  author={Luza, Lucas Matana and Ruospo, Annachiara and Bosio, Alberto and Sanchez, Ernesto and Dilillo, Luigi},
  booktitle={2021 24th International Symposium on Design and Diagnostics of Electronic Circuits \& Systems (DDECS)},
  pages={41--44},
  year={2021},
  organization={IEEE}
}

@article{baumann2005radiation,
  title={Radiation-induced soft errors in advanced semiconductor technologies},
  author={Baumann, Robert C},
  journal={IEEE Transactions on Device and materials reliability},
  volume={5},
  number={3},
  pages={305--316},
  year={2005},
  publisher={IEEE}
}

@inproceedings{chen2003dynamic,
  title={Dynamic NBTI of PMOS transistors and its impact on device lifetime},
  author={Chen, GCKY and Chuah, KY and Li, MF and Chan, Daniel SH and Ang, CH and Zheng, JZ and Jin, Y and Kwong, DL},
  booktitle={2003 IEEE International Reliability Physics Symposium Proceedings, 2003. 41st Annual.},
  pages={196--202},
  year={2003},
  organization={IEEE}
}

@article{borkar2005designing,
  title={Designing reliable systems from unreliable components: the challenges of transistor variability and degradation},
  author={Borkar, Shekhar},
  journal={Ieee Micro},
  volume={25},
  number={6},
  pages={10--16},
  year={2005},
  publisher={IEEE}
}

@inproceedings{choi2019sensitivity,
  title={Sensitivity based error resilient techniques for energy efficient deep neural network accelerators},
  author={Choi, Wonseok and Shin, Dongyeob and Park, Jongsun and Ghosh, Swaroop},
  booktitle={Proceedings of the 56th Annual Design Automation Conference 2019},
  pages={1--6},
  year={2019}
}

@article{deligiannis2021towards,
  title={Towards the integration of reliability and security mechanisms to enhance the fault resilience of neural networks},
  author={Deligiannis, Nikolaos Ioannis and Cantoro, Riccardo and Reorda, Matteo Sonza and Traiola, Marcello and Valea, Emanuele},
  journal={IEEE Access},
  volume={9},
  pages={155998--156012},
  year={2021},
  publisher={IEEE}
}

@inproceedings{burelt2022improving,
  title={Improving DNN Fault Tolerance in Semantic Segmentation Applications},
  author={BurelT, St{\'e}phane and EvansT, Adrian and Anghel, Lorena},
  booktitle={2022 IEEE International Symposium on Defect and Fault Tolerance in VLSI and Nanotechnology Systems (DFT)},
  pages={1--6},
  year={2022},
  organization={IEEE}
}

@inproceedings{chan2022fault,
  title={The fault in our data stars: studying mitigation techniques against faulty training data in machine learning applications},
  author={Chan, Abraham and Gujarati, Arpan and Pattabiraman, Karthik and Gopalakrishnan, Sathish},
  booktitle={2022 52nd Annual IEEE/IFIP International Conference on Dependable Systems and Networks (DSN)},
  pages={163--171},
  year={2022},
  organization={IEEE}
}

@inproceedings{laskar2022characterizing,
  title={Characterizing Deep Learning Neural Network Failures Between Algorithmic Inaccuracy and Transient Hardware Faults},
  author={Laskar, Sabuj and Rahman, Md Hasanur and Zhang, Bohan and Li, Guanpeng},
  booktitle={2022 IEEE 27th Pacific Rim International Symposium on Dependable Computing (PRDC)},
  pages={54--67},
  year={2022},
  organization={IEEE}
}

@inproceedings{laskar2022tensorfi+,
  title={TensorFI+: A Scalable Fault Injection Framework for Modern Deep Learning Neural Networks},
  author={Laskar, Sabuj and Rahman, Md Hasanur and Li, Guanpeng},
  booktitle={2022 IEEE International Symposium on Software Reliability Engineering Workshops (ISSREW)},
  pages={246--251},
  year={2022},
  organization={IEEE}
}

@inproceedings{agarwal2022lltfi,
  title={LLTFI: Framework Agnostic Fault Injection for Machine Learning Applications (Tools and Artifact Track)},
  author={Agarwal, Udit Kumar and Chan, Abraham and Pattabiraman, Karthik},
  booktitle={2022 IEEE 33rd International Symposium on Software Reliability Engineering (ISSRE)},
  pages={286--296},
  year={2022},
  organization={IEEE}
}

@article{narayanan2022fault,
  title={Fault Injection for TensorFlow Applications},
  author={Narayanan, Niranjhana and Chen, Zitao and Fang, Bo and Li, Guanpeng and Pattabiraman, Karthik and Debardeleben, Nathan},
  journal={IEEE Transactions on Dependable and Secure Computing},
  year={2022},
  publisher={IEEE}
}

@inproceedings{rojas2021understanding,
  title={Understanding soft error sensitivity of deep learning models and frameworks through checkpoint alteration},
  author={Rojas, Elvis and P{\'e}rez, Diego and Calhoun, Jon C and Gomez, Leonardo Bautista and Jones, Terry and Meneses, Esteban},
  booktitle={2021 IEEE International Conference on Cluster Computing (CLUSTER)},
  pages={492--503},
  year={2021},
  organization={IEEE}
}

@inproceedings{syed2021fault,
  title={Fault Resilience Analysis of Quantized Deep Neural Networks},
  author={Syed, Rizwan Tariq and Ulbricht, Markus and Piotrowski, Krzysztof and Krstic, Milos},
  booktitle={2021 IEEE 32nd International Conference on Microelectronics (MIEL)},
  pages={275--279},
  year={2021},
  organization={IEEE}
}

@article{zhan2021improving,
  title={Improving fault tolerance for reliable DNN using boundary-aware activation},
  author={Zhan, Jinyu and Sun, Ruoxu and Jiang, Wei and Jiang, Yucheng and Yin, Xunzhao and Zhuo, Cheng},
  journal={IEEE Transactions on Computer-Aided Design of Integrated Circuits and Systems},
  volume={41},
  number={10},
  pages={3414--3425},
  year={2021},
  publisher={IEEE}
}

@article{ruospo2021investigating,
  title={Investigating data representation for efficient and reliable convolutional neural networks},
  author={Ruospo, Annachiara and Sanchez, Ernesto and Traiola, Marcello and O’connor, Ian and Bosio, Alberto},
  journal={Microprocessors and Microsystems},
  volume={86},
  pages={104318},
  year={2021},
  publisher={Elsevier}
}

@article{jang2021mate,
  title={MATE: Memory-and Retraining-Free Error Correction for Convolutional Neural Network Weights},
  author={Jang, Myeungjae and Hong, Jeongkyu},
  journal={Journal of information and communication convergence engineering},
  volume={19},
  number={1},
  pages={22--28},
  year={2021},
  publisher={The Korea Institute of Information and Commucation Engineering}
}

@article{ozen2021snr,
  title={Snr: S queezing n umerical r ange defuses bit error vulnerability surface in deep neural networks},
  author={Ozen, Elbruz and Orailoglu, Alex},
  journal={ACM Transactions on Embedded Computing Systems (TECS)},
  volume={20},
  number={5s},
  pages={1--25},
  year={2021},
  publisher={ACM New York, NY}
}

@inproceedings{malekzadeh2021impact,
  title={The Impact of Faults on DNNs: A Case Study},
  author={Malekzadeh, Elaheh and Rohbani, Nezam and Lu, Zhonghai and Ebrahimi, Masoumeh},
  booktitle={2021 IEEE International Symposium on Defect and Fault Tolerance in VLSI and Nanotechnology Systems (DFT)},
  pages={1--6},
  year={2021},
  organization={IEEE}
}

@inproceedings{lee2022bipolar,
  title={Bipolar vector classifier for fault-tolerant deep neural networks},
  author={Lee, Suyong and Choi, Insu and Yang, Joon-Sung},
  booktitle={Proceedings of the 59th ACM/IEEE Design Automation Conference},
  pages={673--678},
  year={2022}
}

@inproceedings{ghavami2022fitact,
  title={FitAct: Error Resilient Deep Neural Networks via Fine-Grained Post-Trainable Activation Functions},
  author={Ghavami, Behnam and Sadati, Mani and Fang, Zhenman and Shannon, Lesley},
  booktitle={2022 Design, Automation \& Test in Europe Conference \& Exhibition (DATE)},
  pages={1239--1244},
  year={2022},
  organization={IEEE}
}

@inproceedings{amarnath2022soft,
  title={Soft Error Resilient Deep Learning Systems Using Neuron Gradient Statistics},
  author={Amarnath, Chandramouli and Mejri, Mohamed and Ma, Kwondo and Chatterjee, Abhijit},
  booktitle={2022 IEEE 28th International Symposium on On-Line Testing and Robust System Design (IOLTS)},
  pages={1--7},
  year={2022},
  organization={IEEE}
}

@inproceedings{lee2022value,
  title={Value-aware parity insertion ECC for fault-tolerant deep neural network},
  author={Lee, Seo-Seok and Yang, Joon-Sung},
  booktitle={2022 Design, Automation \& Test in Europe Conference \& Exhibition (DATE)},
  pages={724--729},
  year={2022},
  organization={IEEE}
}

@article{adam2021analyzing,
  title={Analyzing the instructions vulnerability of dense convolutional network on GPUS},
  author={Adam, Khalid and Mohd, Izzeldin I and Ibrahim, Younis},
  journal={International Journal of Electrical and Computer Engineering (IJECE)},
  volume={11},
  number={5},
  pages={4481--4488},
  year={2021}
}

@article{adam2021impact,
  title={The impact of the soft errors in convolutional neural network on GPUs: Alexnet as case study},
  author={Adam, Khalid and Mohd, Izzeldin I and Younis, Younis M},
  journal={Procedia Computer Science},
  volume={182},
  pages={89--94},
  year={2021},
  publisher={Elsevier}
}

@inproceedings{garrett2021improving,
  title={Improving dependability of onboard deep learning with resilient tensorflow},
  author={Garrett, Tyler and George, Alan D},
  booktitle={2021 IEEE Space Computing Conference (SCC)},
  pages={134--142},
  year={2021},
  organization={IEEE}
}

@inproceedings{condia2022multi,
  title={A multi-level approach to evaluate the impact of gpu permanent faults on cnn's reliability},
  author={Condia, Josie E Rodriguez and Guerrero-Balaguera, Juan-David and Dos Santos, Fernando F and Reorda, Matteo Sonza and Rech, Paolo},
  booktitle={2022 IEEE International Test Conference (ITC)},
  pages={278--287},
  year={2022},
  organization={IEEE}
}

@article{dos2022characterizing,
  title={Characterizing a Neutron-Induced Fault Model for Deep Neural Networks},
  author={Dos Santos, Fernando Fernandes and Kritikakou, Angeliki and Condia, Josie E Rodriguez and Guerrero-Balaguera, Juan-David and Reorda, Matteo Sonza and Sentieys, Olivier and Rech, Paolo},
  journal={IEEE Transactions on Nuclear Science},
  year={2022},
  publisher={IEEE}
}

@inproceedings{guerrero2022effective,
  title={Effective fault simulation of GPU’s permanent faults for reliability estimation of CNNs},
  author={Guerrero-Balaguera, Juan-David and Sierra, Robert Limas and Reorda, Matteo Sonza},
  booktitle={2022 IEEE 28th International Symposium on On-Line Testing and Robust System Design (IOLTS)},
  pages={1--6},
  year={2022},
  organization={IEEE}
}

@inproceedings{guerrero2022evaluating,
  title={Evaluating the impact of Permanent Faults in a GPU running a Deep Neural Network},
  author={Guerrero-Balaguera, Juan-David and Galasso, Luigi and Sierra, Robert Limas and Sanchez, Ernesto and Reorda, Matteo Sonza},
  booktitle={2022 IEEE International Test Conference in Asia (ITC-Asia)},
  pages={96--101},
  year={2022},
  organization={IEEE}
}

@inproceedings{guerrero2022reliability,
  title={Reliability assessment of neural networks in gpus: A framework for permanent faults injections},
  author={Guerrero-Balaguera, Juan-David and Galasso, Luigi and Sierra, Robert Limas and Reorda, Matteo Sonza},
  booktitle={2022 IEEE 31st International Symposium on Industrial Electronics (ISIE)},
  pages={959--962},
  year={2022},
  organization={IEEE}
}

@inproceedings{guerrero2022neural,
  title={Neural Network's Reliability to Permanent Faults: Analyzing the Impact of Performance Optimizations in GPUs},
  author={Guerrero-Balaguera, Juan-David and Condia, Josie E Rodriguez and Reorda, Matteo Sonza},
  booktitle={2022 29th IEEE International Conference on Electronics, Circuits and Systems (ICECS)},
  pages={1--4},
  year={2022},
  organization={IEEE}
}

@article{bolchini2022fast,
  title={Fast and accurate error simulation for cnns against soft errors},
  author={Bolchini, Cristiana and Cassano, Luca and Miele, Antonio and Toschi, Alessandro},
  journal={IEEE Transactions on Computers},
  year={2022},
  publisher={IEEE}
}

@inproceedings{bolchini2022selective,
  title={Selective Hardening of CNNs based on Layer Vulnerability Estimation},
  author={Bolchini, Cristiana and Cassano, Luca and Miele, Antonio and Nazzari, Alessandro},
  booktitle={2022 IEEE International Symposium on Defect and Fault Tolerance in VLSI and Nanotechnology Systems (DFT)},
  pages={1--6},
  year={2022},
  organization={IEEE}
}

@inproceedings{cavagnero2022transient,
  title={Transient-Fault-Aware Design and Training to Enhance DNNs Reliability with Zero-Overhead},
  author={Cavagnero, Niccol{\`o} and Dos Santos, Fernando and Ciccone, Marco and Averta, Giuseppe and Tommasi, Tatiana and Rech, Paolo},
  booktitle={2022 IEEE 28th International Symposium on On-Line Testing and Robust System Design (IOLTS)},
  pages={1--7},
  year={2022},
  organization={IEEE}
}

@article{liu2021analyzing,
  title={Analyzing and increasing soft error resilience of Deep Neural Networks on ARM processors},
  author={Liu, Zhi and Liu, Yuhong and Chen, Zhengming and Guo, Gang and Wang, Haibin},
  journal={Microelectronics Reliability},
  volume={124},
  pages={114331},
  year={2021},
  publisher={Elsevier}
}

@article{liu2022efficient,
  title={An efficient structure to improve the reliability of deep neural networks on ARMs},
  author={Liu, Zhi and Yang, Xinni},
  journal={Microelectronics Reliability},
  volume={136},
  pages={114729},
  year={2022},
  publisher={Elsevier}
}

@article{liu2022using,
  title={Using checksum to improve the reliability of embedded convolutional neural networks},
  author={Liu, Zhi and Deng, Zhen and Yang, Xinni},
  journal={Microelectronics Reliability},
  volume={136},
  pages={114666},
  year={2022},
  publisher={Elsevier}
}

@article{abich2021applying,
  title={Applying lightweight soft error mitigation techniques to embedded mixed precision deep neural networks},
  author={Abich, Geancarlo and Gava, Jonas and Garibotti, Rafael and Reis, Ricardo and Ost, Luciano},
  journal={IEEE Transactions on Circuits and Systems I: Regular Papers},
  volume={68},
  number={11},
  pages={4772--4782},
  year={2021},
  publisher={IEEE}
}

@article{abich2022impact,
  title={The impact of soft errors in memory units of edge devices executing convolutional neural networks},
  author={Abich, Geancarlo and Garibotti, Rafael and Reis, Ricardo and Ost, Luciano},
  journal={IEEE Transactions on Circuits and Systems II: Express Briefs},
  volume={69},
  number={3},
  pages={679--683},
  year={2022},
  publisher={IEEE}
}

@inproceedings{abich2022impactthread,
  title={Impact of thread parallelism on the soft error reliability of convolution neural networks},
  author={Abich, Geancarlo and Garibotti, Rafael and Gava, Jonas and Reis, Ricardo and Ost, Luciano},
  booktitle={2022 IEEE 13th Latin America Symposium on Circuits and System (LASCAS)},
  pages={1--4},
  year={2022},
  organization={IEEE}
}

@inproceedings{gava2022soft,
  title={Soft Error Assessment of CNN Inference Models Running on a RISC-V Processor},
  author={Gava, Jonas and Dorneles, Guilherme and Reis, Ricardo and Garibotti, Rafael and Ost, Luciano},
  booktitle={2022 29th IEEE International Conference on Electronics, Circuits and Systems (ICECS)},
  pages={1--4},
  year={2022},
  organization={IEEE}
}

@inproceedings{abich2021impactprecision,
  title={The impact of precision bitwidth on the soft error reliability of the MobileNet network},
  author={Abich, Geancarlo and Reis, Ricardo and Ost, Luciano},
  booktitle={2021 IEEE 12th Latin America Symposium on Circuits and System (LASCAS)},
  pages={1--4},
  year={2021},
  organization={IEEE}
}

@article{liu2021hyca,
  title={HyCA: A Hybrid Computing Architecture for Fault-Tolerant Deep Learning},
  author={Liu, Cheng and Chu, Cheng and Xu, Dawen and Wang, Ying and Wang, Qianlong and Li, Huawei and Li, Xiaowei and Cheng, Kwang-Ting},
  journal={IEEE Transactions on Computer-Aided Design of Integrated Circuits and Systems},
  volume={41},
  number={10},
  pages={3400--3413},
  year={2021},
  publisher={IEEE}
}

@inproceedings{mahmoud2021optimizing,
  title={Optimizing Selective Protection for CNN Resilience.},
  author={Mahmoud, Abdulrahman and Hari, Siva Kumar Sastry and Fletcher, Christopher W and Adve, Sarita V and Sakr, Charbel and Shanbhag, Naresh R and Molchanov, Pavlo and Sullivan, Michael B and Tsai, Timothy and Keckler, Stephen W},
  booktitle={ISSRE},
  pages={127--138},
  year={2021}
}

@inproceedings{ruospo2022selective,
  title={Selective hardening of critical neurons in deep neural networks},
  author={Ruospo, Annachiara and Gavarini, Gabriele and Bragaglia, Ilaria and Traiola, Marcello and Bosio, Alberto and Sanchez, Ernesto},
  booktitle={2022 25th International Symposium on Design and Diagnostics of Electronic Circuits and Systems (DDECS)},
  pages={136--141},
  year={2022},
  organization={IEEE}
}

@inproceedings{zhang2022estimating,
  title={Estimating vulnerability of all model parameters in dnn with a small number of fault injections},
  author={Zhang, Yangchao and Itsuji, Hiroaki and Uezono, Takumi and Toba, Tadanobu and Hashimoto, Masanori},
  booktitle={2022 Design, Automation \& Test in Europe Conference \& Exhibition (DATE)},
  pages={60--63},
  year={2022},
  organization={IEEE}
}

@inproceedings{gavarini2022open,
  title={Open-Set Recognition: an Inexpensive Strategy to Increase DNN Reliability},
  author={Gavarini, G and Stucchi, D and Ruospo, A and Boracchi, G and Sanchez, E},
  booktitle={2022 IEEE 28th International Symposium on On-Line Testing and Robust System Design (IOLTS)},
  pages={1--7},
  year={2022},
  organization={IEEE}
}

@inproceedings{veronesi2022exploring,
  title={Exploring Software Models for the Resilience Analysis of Deep Learning Accelerators: the NVDLA Case Study},
  author={Veronesi, Alessandro and Dall’Occo, Francesco and Bertozzi, Davide and Favalli, Michele and Krstic, Milos},
  booktitle={2022 25th International Symposium on Design and Diagnostics of Electronic Circuits and Systems (DDECS)},
  pages={142--147},
  year={2022},
  organization={IEEE}
}

@article{burel2022mozart+,
  title={Mozart+: Masking outputs with zeros for improved architectural robustness and testing of dnn accelerators},
  author={Burel, Stephane and Evans, Adrian and Anghel, Lorena},
  journal={IEEE Transactions on Device and Materials Reliability},
  volume={22},
  number={2},
  pages={120--128},
  year={2022},
  publisher={IEEE}
}

@article{jasemi2020enhancing,
  title={Enhancing reliability of emerging memory technology for machine learning accelerators},
  author={Jasemi, Masoomeh and Hessabi, Shaahin and Bagherzadeh, Nader},
  journal={IEEE Transactions on Emerging Topics in Computing},
  volume={9},
  number={4},
  pages={2234--2240},
  year={2020},
  publisher={IEEE}
}

@inproceedings{tsai2021evaluating,
  title={Evaluating the Impact of Fault-Tolerance Capability of Deep Neural Networks Caused by Faults},
  author={Tsai, Yung-Yu and Li, Jin-Fu},
  booktitle={2021 IEEE 34th International System-on-Chip Conference (SOCC)},
  pages={272--277},
  year={2021},
  organization={IEEE}
}

@inproceedings{nguyen2021low,
  title={Low-cost and effective fault-tolerance enhancement techniques for emerging memories-based deep neural networks},
  author={Nguyen, Thai-Hoang and Imran, Muhammad and Choi, Jaehyuk and Yang, Joon-Sung},
  booktitle={2021 58th ACM/IEEE Design Automation Conference (DAC)},
  pages={1075--1080},
  year={2021},
  organization={IEEE}
}

@article{samajdar2018scale,
  title={Scale-sim: Systolic cnn accelerator simulator},
  author={Samajdar, Ananda and Zhu, Yuhao and Whatmough, Paul and Mattina, Matthew and Krishna, Tushar},
  journal={arXiv preprint arXiv:1811.02883},
  year={2018}
}

@inproceedings{zhao2022fsa,
  title={FSA: An Efficient Fault-tolerant Systolic Array-based DNN Accelerator Architecture},
  author={Zhao, Yingnan and Wang, Ke and Louri, Ahmed},
  booktitle={2022 IEEE 40th International Conference on Computer Design (ICCD)},
  pages={545--552},
  year={2022},
  organization={IEEE}
}

@article{de2022firenn,
  title={Firenn: Neural networks reliability evaluation on hybrid platforms},
  author={De Sio, Corrado and Azimi, Sarah and Sterpone, Luca},
  journal={IEEE Transactions on Emerging Topics in Computing},
  volume={10},
  number={2},
  pages={549--563},
  year={2022},
  publisher={IEEE}
}

@inproceedings{siddique2021exploring,
  title={Exploring fault-energy trade-offs in approximate DNN hardware accelerators},
  author={Siddique, Ayesha and Basu, Kanad and Hoque, Khaza Anuarul},
  booktitle={2021 22nd International Symposium on Quality Electronic Design (ISQED)},
  pages={343--348},
  year={2021},
  organization={IEEE}
}

@inproceedings{benevenuti2021neutron,
  title={Neutron-induced Faults on CNN for Aerial Image Classification on SRAM-based FPGA Using Softcore GPU and HLS},
  author={Benevenuti, Fabio and Gon{\c{c}}alves, M{\'a}rcio and Junior, Evaldo Carlos Fonseca Pereira and Vaz, Rafael Galhardo and Gon{\c{c}}alez, Odair Lelis and Azambuja, Jos{\'e} Rodrigo and Kastensmidt, Fernanda Lima},
  booktitle={2021 21th European Conference on Radiation and Its Effects on Components and Systems (RADECS)},
  pages={1--4},
  year={2021},
  organization={IEEE}
}

@inproceedings{hosseinkhani2021improving,
  title={Improving Soft Error Reliability of FPGA-based Deep Neural Networks with Reduced Approximate TMR},
  author={Hosseinkhani, Anahita and Ghavami, Behnam},
  booktitle={2021 11th International Conference on Computer Engineering and Knowledge (ICCKE)},
  pages={459--464},
  year={2021},
  organization={IEEE}
}

@article{gao2022reliability,
  title={Reliability evaluation of FPGA based pruned neural networks},
  author={Gao, Zhen and Yao, Yi and Wei, Xiaohui and Yan, Tong and Zeng, Shulin and Ge, Guangjun and Wang, Yu and Ullah, Anees and Reviriego, Pedro},
  journal={Microelectronics Reliability},
  volume={130},
  pages={114498},
  year={2022},
  publisher={Elsevier}
}

@inproceedings{agiakatsikas2021evaluation,
  title={Evaluation of the Xilinx Deep Learning Processing Unit under Neutron Irradiation},
  author={Agiakatsikas, Dimitris and Foutris, Nikos and Sari, Aitzan and Vlagkoulis, Vasileios and Souvatzoglou, Ioanna and Psarakis, Mihalis and Luj{\'a}n, Mikel and Kastriotou, Maria and Cazzaniga, Carlo},
  booktitle={2021 21th European Conference on Radiation and Its Effects on Components and Systems (RADECS)},
  pages={1--4},
  year={2021},
  organization={IEEE}
}

@inproceedings{gambardella2022accelerated,
  title={Accelerated Radiation Test on Quantized Neural Networks trained with Fault Aware Training},
  author={Gambardella, Giulio and Fraser, Nicholas J and Zahid, Ussama and Furano, Gianluca and Blott, Michaela},
  booktitle={2022 IEEE Aerospace Conference (AERO)},
  pages={1--7},
  year={2022},
  organization={IEEE}
}

@article{maillard2022radiation,
  title={Radiation Tolerant Deep Learning Processor Unit (DPU) based platform using Xilinx 20nm Kintex UltraScale™ FPGA},
  author={Maillard, Pierre and Chen, Yanran P and Vidmar, Jason and Fraser, Nicholas and Gambardella, Giulio and Sawant, Minal and Voogel, Martin L},
  journal={IEEE Transactions on Nuclear Science},
  year={2022},
  publisher={IEEE}
}

@inproceedings{rech2022reliability,
  title={Reliability of google's tensor processing units for embedded applications},
  author={Rech, Rubens Luiz and Rech, Paolo},
  booktitle={2022 Design, Automation \& Test in Europe Conference \& Exhibition (DATE)},
  pages={376--381},
  year={2022},
  organization={IEEE}
}

@article{junior2022high,
  title={High energy and thermal neutron sensitivity of google tensor processing units},
  author={Junior, Rubens Luiz Rech and Malde, Sujit and Cazzaniga, Carlo and Kastriotou, Maria and Letiche, Manon and Frost, Christopher and Rech, Paolo},
  journal={IEEE Transactions on Nuclear Science},
  volume={69},
  number={3},
  pages={567--575},
  year={2022},
  publisher={IEEE}
}

@article{perez2022gpu,
  title={GPU Devices for Safety-Critical Systems: A Survey},
  author={Perez-Cerrolaza, Jon and Abella, Jaume and Kosmidis, Leonidas and Calderon, Alejandro J and Cazorla, Francisco and Flores, Jose Luis},
  journal={ACM Computing Surveys},
  volume={55},
  number={7},
  pages={1--37},
  year={2022},
  publisher={ACM New York, NY}
}

@article{su2023testability,
  title={Testability and dependability of AI hardware: Survey, trends, challenges, and perspectives},
  author={Su, Fei and Liu, Chunsheng and Stratigopoulos, Haralampos-G},
  journal={IEEE Design \& Test},
  year={2023},
  publisher={IEEE}
}

@article{ruospo2023survey,
  title={A Survey on Deep Learning Resilience Assessment Methodologies},
  author={Ruospo, Annachiara and Sanchez, Ernesto and Luza, Lucas Matana and Dilillo, Luigi and Traiola, Marcello and Bosio, Alberto},
  journal={Computer},
  volume={56},
  number={2},
  pages={57--66},
  year={2023},
  publisher={IEEE}
}

@article{mahdavinejad2018machine,
  title={Machine learning for Internet of Things data analysis: A survey},
  author={Mahdavinejad, Mohammad Saeid and Rezvan, Mohammadreza and Barekatain, Mohammadamin and Adibi, Peyman and Barnaghi, Payam and Sheth, Amit P},
  journal={Digital Communications and Networks},
  volume={4},
  number={3},
  pages={161--175},
  year={2018},
  publisher={Elsevier}
}

@article{lai2018cmsis,
  title={Cmsis-nn: Efficient neural network kernels for arm cortex-m cpus},
  author={Lai, Liangzhen and Suda, Naveen and Chandra, Vikas},
  journal={arXiv preprint arXiv:1801.06601},
  year={2018}
}

@article{sanchez2020tinyml,
  title={Tinyml-enabled frugal smart objects: Challenges and opportunities},
  author={Sanchez-Iborra, Ramon and Skarmeta, Antonio F},
  journal={IEEE Circuits and Systems Magazine},
  volume={20},
  number={3},
  pages={4--18},
  year={2020},
  publisher={IEEE}
}

\end{document}